%% file: main.tex
\documentclass[10pt,twocolumn,letterpaper]{article}
\usepackage[pagenumbers]{wacv}

\makeatletter
\@namedef{ver@everyshi.sty}{}
\makeatother

\usepackage[dvipsnames,svgnames,x11names,table,xcdraw]{xcolor}
\usepackage{tikz}
\usepackage{pgfplotstable}
\pgfplotsset{compat=1.9}
\usepackage{booktabs}
\usepackage{multirow}
\usepackage{makecell}
\usepackage[export]{adjustbox}
\usepackage{amsmath}
\usepackage{amssymb}
\usepackage{enumitem}
\usepackage{diagbox}
\usepackage{algorithm}
\usepackage{algpseudocode}
\input{tex/abbrev}
\input{tex/defn}
\usepackage[pagebackref,breaklinks,colorlinks]{hyperref}
\newcommand\blfootnote[1]{%
  \begingroup
  \renewcommand\thefootnote{}\footnote{#1}%
  \addtocounter{footnote}{-1}%
  \endgroup
}

\usepackage[capitalize]{cleveref}
\crefname{section}{Sec.}{Secs.}
\Crefname{section}{Section}{Sections}
\Crefname{table}{Table}{Tables}
\crefname{table}{Tab.}{Tabs.}

\renewcommand\paragraph[1]{\noindent \textbf{#1}}
\begin{document}

\title{Composed Image Retrieval for Training-\textsc{Free} \textsc{Dom}ain Conversion}
\author{Nikos Efthymiadis$^{1^*}$
\and
Bill Psomas$^{1,2}$
\and
Zakaria Laskar$^1$
\and
Konstantinos Karantzalos$^2$
\and
Yannis Avrithis$^3$
\and
Ondřej Chum$^1$
\and
Giorgos Tolias$^1$
\and
\\
$^1$VRG, FEE, Czech Technical University in Prague \hspace{15pt} $^2$National Technical University of Athens \\
$^3$
Institute of Advanced Research in Artificial Intelligence (IARAI), Austria
}

\input{tex/teaser}
\maketitle
\begin{abstract}
    \input{tex/abstract}
\end{abstract}

\input{tex/intro}
\input{tex/related}
\input{tex/method}
\input{tex/exp}

\input{tex/conclusions}

\vspace{3pt}
\textbf{Acknowledgments:} This work was supported by the Junior Star GACR GM 21-28830M, the Czech Technical University in Prague grant No. SGS23/173/OHK3/3T/13, the programme Johannes Amos Comenius CZ.02.01.01/00/22\_010/0003405, the CTU institutional support (Future fund RVO13000), the RAMONES H2020 project (grant: 101017808), and the HFRI under the BiCUBES project (grant: 03943). The access to the computational infrastructure of the OP VVV funded project CZ.02.1.01/0.0/0.0/16\_019/0000765 ``Research Center for Informatics'' is gratefully acknowledged.

\clearpage
{
\small
\bibliographystyle{ieee_fullname}
\bibliography{egbib}
}
\clearpage
\appendix
\input{tex/appendix}
\end{document}

%% file: tex/abbrev.tex
\newcommand{\fig}[2][1]{\includegraphics[width=#1\linewidth]{fig/#2}}

\newcommand{\figsup}[2][1]{\includegraphics[width=#1\linewidth]{fig_supp/#2}}

\newcommand{\bfigsupc}[2][1]{\includegraphics[width=#1\linewidth,height=#1\linewidth,cfbox=applegreen 1.5pt 1.5pt]{fig_supp/#2}}
\newcommand{\bfigsupw}[2][1]{\includegraphics[width=#1\linewidth,height=#1\linewidth,cfbox=appleorange 1.5pt 1.5pt]{fig_supp/#2}}
\newcommand{\bfigsupq}[2][1]{\includegraphics[width=#1\linewidth,height=#1\linewidth,cfbox=OrangeFrame 1.5pt 1.5pt]{fig_supp/#2}}

\newcommand{\mr}[2]{\multirow{#1}{*}{#2}}
\newcommand{\mc}[2]{\multicolumn{#1}{c}{#2}}
\newcommand{\Th}[1]{\textsc{#1}}
\newcommand{\tb}[1]{\textbf{#1}}

\newcommand{\comment} [1]{{\color{orange} \Comment     #1}} 

\def\nssp{\hspace{-3pt}}

\def\xssp{\hspace{1pt}}
\def\ssp{\hspace{3pt}}
\def\msp{\hspace{6pt}}
\def\lsp{\hspace{12pt}}
\def\xlsp{\hspace{20pt}}

\newcommand{\eq}[1]{(\ref{eq:#1})\xspace}

\def\l1{\ensuremath{\ell_1}\xspace}
\def\l2{\ensuremath{\ell_2}\xspace}

\newcommand{\tran}{^\top}

\newcommand{\real}{\mathbb{R}}

\newcommand{\defn}{\mathrel{:=}}

\newcommand{\card}[1]{\left|{#1}\right|\xspace}

\newcommand{\cI}{\mathcal{I}}

\newcommand{\cT}{\mathcal{T}}

\newcommand{\cV}{\mathcal{V}}
\newcommand{\cW}{\mathcal{W}}
\newcommand{\cX}{\mathcal{X}}
\newcommand{\cY}{\mathcal{Y}}
\newcommand{\cZ}{\mathcal{Z}}

\newcommand{\va}{\mathbf{a}}

\newcommand{\vq}{\mathbf{q}}

\newcommand{\vt}{\mathbf{t}}

\newcommand{\vv}{\mathbf{v}}
\newcommand{\vw}{\mathbf{w}}
\newcommand{\vx}{\mathbf{x}}
\newcommand{\vy}{\mathbf{y}}
\newcommand{\vz}{\mathbf{z}}

\newcommand{\vone}{\mathbf{1}}

\definecolor{applegreen}{RGB}{52,199,89}
\definecolor{appleorange}{RGB}{255,149,0}
\definecolor{appleyellow}{RGB}{255,204,0}

\definecolor{lightred}{RGB}{231, 76, 60}   
\definecolor{lightblue}{RGB}{54, 69, 79}
\definecolor{lightgreen}{RGB}{50, 205, 50} 

\definecolor{appleblue}{RGB}{80,122,255}  
\definecolor{applepink}{RGB}{217,127,174}  
\definecolor{appleteal}{RGB}{85,163,152}  
\definecolor{applepurple}{RGB}{138,107,221}  
\definecolor{applemustard}{RGB}{234,179,77}  
\definecolor{appleorange}{RGB}{255,127,80}  

\makeatletter
\DeclareRobustCommand\onedot{\futurelet\@let@token\@onedot}
\def\@onedot{\ifx\@let@token.\else.\null\fi\xspace}
\def\eg{\emph{e.g}\onedot} 
\def\ie{\emph{i.e}\onedot} 
\def\vs{\emph{vs\onedot}}
 
 \def\vs{\emph{vs}\onedot}
\def\wrt{w.r.t\onedot} 

\makeatother


%% file: tex/defn.tex
\newcommand{\NN}{\operatorname{NN}}

\newcommand{\ours}{\textsc{FreeDom}\xspace}
\newcommand{\picword}{Pic2Word\xspace}
\newcommand{\searle}{SEARLE\xspace}
\newcommand{\compodiff}{CompoDiff\xspace}
\newcommand{\cirevl}{{CIReVL}\xspace}
\newcommand{\weicom}{WeiCom\xspace}
\newcommand{\magic}{MagicLens\xspace}

\newcommand{\imagenet}{{ImageNet-1k}\xspace}
\newcommand{\imagenetr}{{Image\-Net-R}\xspace}
\newcommand{\minidn}{{MiniDomainNet}\xspace}
\newcommand{\nico}{{NICO++}\xspace}
\newcommand{\ltll}{{LTLL}\xspace}
\newcommand{\fashioniq}{{FashionIQ}\xspace}
\newcommand{\cirr}{{CIRR}\xspace}
\newcommand{\circo}{{CIRCO}\xspace}

\newcommand{\txlabel}{text label\xspace}
\newcommand{\tlabels}{labels\xspace}
\newcommand{\txlabels}{text labels\xspace}

\newcommand{\ts}[1]{{\color{purple}#1}}

\definecolor{OrangeFrame}{HTML}{D79B00}
\definecolor{OliveGreen}{RGB}{85, 107, 47}
\definecolor{BrickRed}{RGB}{156, 32, 32}
\definecolor{LightSteelBlue1}{RGB}{202, 225, 255}

%% file: tex/teaser.tex
\makeatletter
\apptocmd\@maketitle{{\teaser{}}}{}{}
\makeatother
\newcommand{\teaser}{%
\vspace{-15pt}
\footnotesize
\centering
\setlength{\tabcolsep}{3.0pt}
\newcommand{\boxcolor}{\setlength{\fboxrule}{.8pt} \color{orange}}
\renewcommand{\cellalign}{bc}
\begin{tabular}{@{\xssp}l@{\ssp}cccc}
\makecell{(a) \\ image: category \\ domain: style \\[4pt] \phantom{}} &
{\boxcolor \fbox{\fig[.1500]{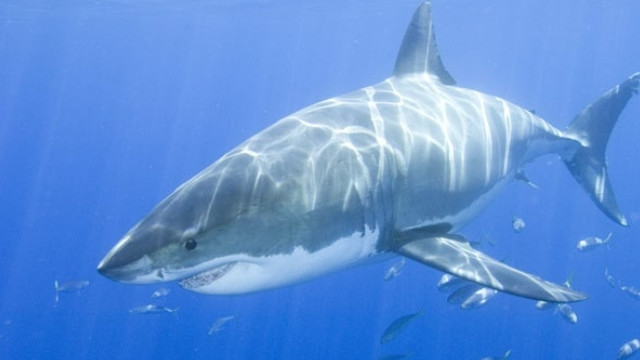}}} &
\fig[.1500]{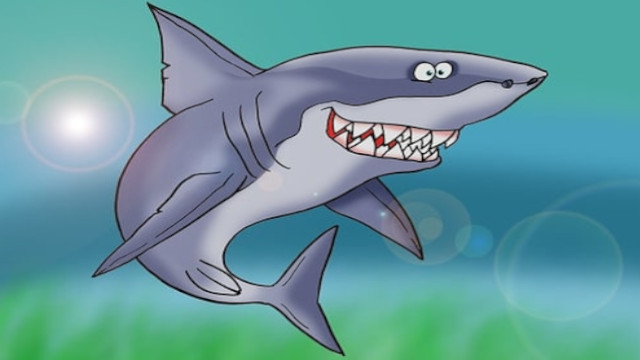} &
\fig[.1500]{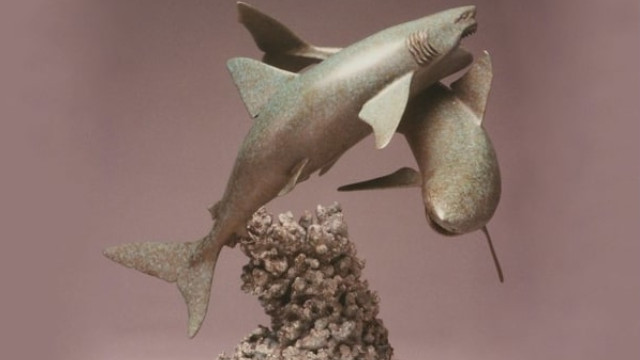} &
\fig[.1500]{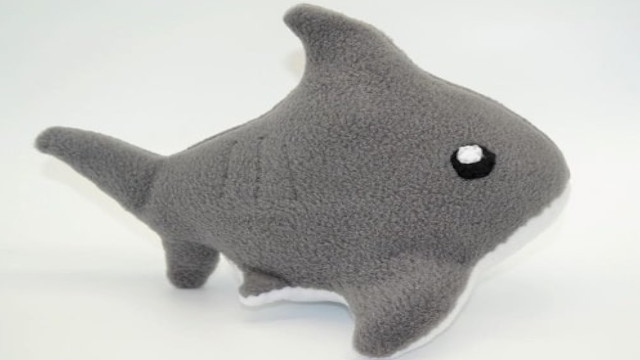} \\

&
image query &
text query: ``cartoon'' &
text query: ``sculpture''&
text query: ``toy''\\[1pt]

\makecell{(b) \\ image: category \\ domain: context \\[4pt] \phantom{}} &
{\boxcolor\fbox{\fig[.1500]{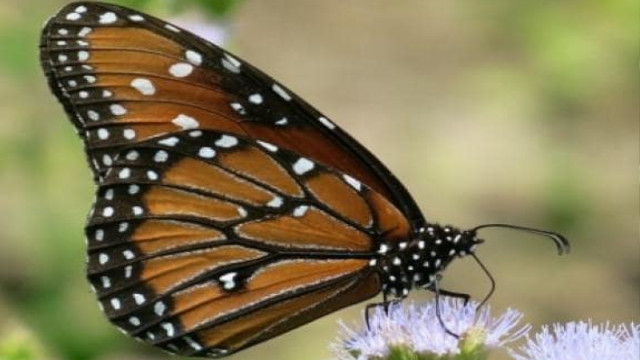}}} &
\fig[.1500]{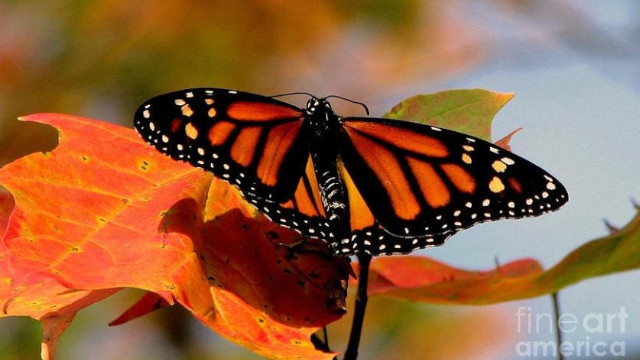} &
\fig[.1500]{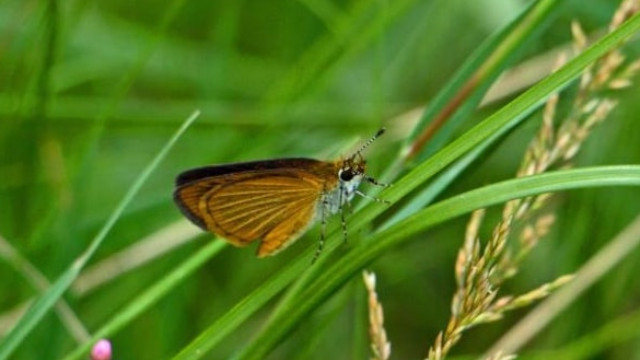} &
\fig[.1500]{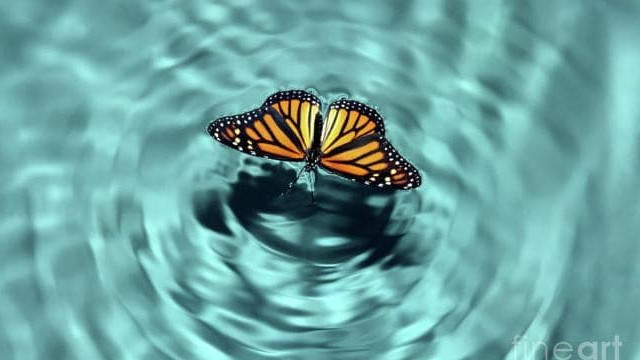} \\

&
image query &
text query: ``autumn'' &
text query: ``grass''&
text query: ``water'' \\[1pt]

\makecell{(c) \\ image: instance \\ domain: style \\[4pt] \phantom{}} &
{\boxcolor\fbox{\fig[.1500]{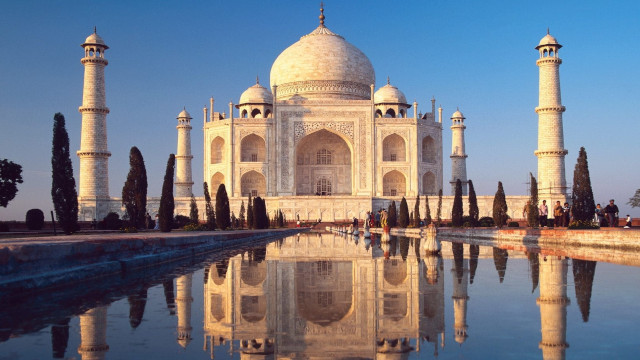}}} &
\fig[.1500]{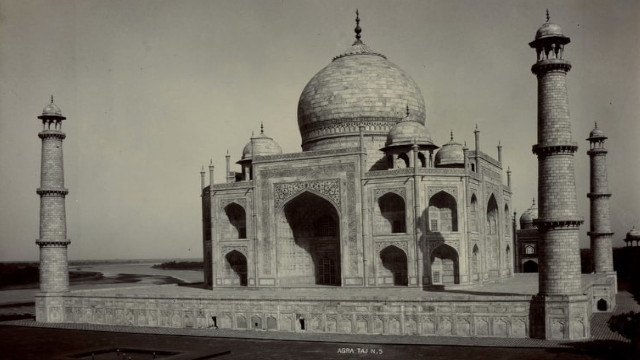} &
{\boxcolor\fbox{\fig[.1500]{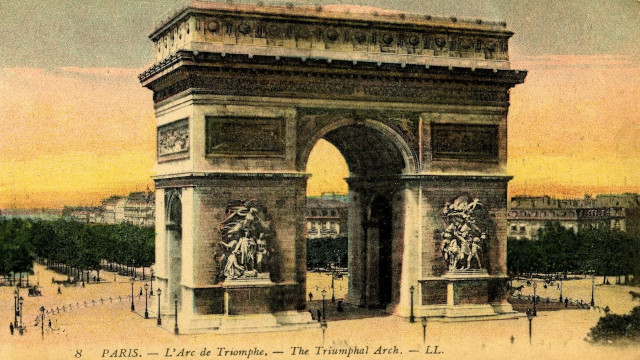}}} &
\fig[.1500]{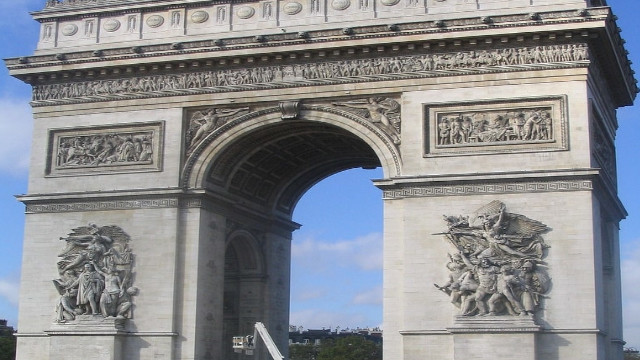} \\

&
image query &
text query: ``archive''&
image query &
text query: ``today''\\

\end{tabular}
\vspace{-5pt}
\captionsetup{type=figure}
\captionof{figure}{We introduce \ours, a training-free, composed image retrieval method for domain conversion based on CLIP~\cite{clip}. Given an \emph{image query} (\textcolor{orange}{framed}) and a \emph{text query} that names a \emph{domain}, images are retrieved having the class of the image query and the domain of the text query. A range of applications is targeted, where classes can be defined at \emph{category level} (a,b)~\cite{imagenet_r,nico++} or \emph{instance level} (c)~\cite{leuven}, and domains can be defined as \emph{styles} (a, c), or \emph{context} (b). For each image query, retrieved images are shown for different text queries.}
\label{fig:teaser}
\par\vspace{10pt}}

%% file: tex/abstract.tex
\blfootnote{$^*$Correspondence: efthynik@fel.cvut.cz} 
This work addresses composed image retrieval in the context of domain conversion, where the content of a query image is retrieved in the domain specified by the query text. We show that a strong vision-language model provides sufficient descriptive power without additional training. The query image is mapped to the text input space using textual inversion. Unlike common practice that invert in the continuous space of text tokens, we use the discrete word space via a nearest-neighbor search in a text vocabulary. With this inversion, the image is softly mapped across the vocabulary and is made more robust using retrieval-based augmentation. Database images are retrieved by a weighted ensemble of text queries combining mapped words with the domain text. Our method outperforms prior art by a large margin on standard and newly introduced benchmarks. Code: \url{https://github.com/NikosEfth/freedom}

%% file: tex/intro.tex
\section{Introduction}
\label{sec:intro}
Image-to-image retrieval is a task that involves applications to landmarks~\cite{rtc19}, fashion products~\cite{llq+16}, face recognition~\cite{dgx+19}, remote sensing~\cite{chaudhuri2019siamese}, and medical images~\cite{nair2020review}, among others. The retrieval is performed according to the visual content of the query~\cite{gar+16, nas+17}. 
On the other hand, if the object can be described with text, then text-to-image retrieval~\cite{sarafianos2019adversarial, zhang2020context, devise} applies. The most flexible way to express the user intent is a query comprising both an image and a text description. This is explored in \emph{composed image retrieval} (CIR)~\cite{tirg, val, lbf, combiner, cosmo, pic2word}, which aims to retrieve target images that are not only visually similar to the query image but also modified by the text query.

Traditionally, CIR methods are supervised by \emph{triplets}~\cite{tirg, jvsm, yin2020disentangled}. However, the labor-intensive process of labeling confined early research to specific applications in fashion~\cite{fashion200k, shoes, fashioniq}, physical states~\cite{mit_states}, object attributes, and object composition~\cite{tirg, lmb+14, mpc}. 
The emergence of vision-language models (VLM)~\cite{clip, align, blip} led to their integration into CIR. Initially, this has been achieved by fine-tuning using triplets~\cite{combiner}. More recently, 
\emph{zero-shot composed image retrieval} (ZS-CIR)~\cite{pic2word, searle} significantly increased the spectrum of applications. Most existing methods use \emph{textual inversion}, \ie, mapping the query image to text, thus allowing query composition purely by means of text. Also, most methods are trained using unlabeled images~\cite{searle, pic2word} or are not trained at all~\cite{cirevl}, but they require the use of large language models~\cite{gpt} (LLM).

In this paper, we focus on a specific variant of composed image retrieval, namely \emph{domain conversion}, where the text query defines the target domain~\cite{pic2word}. Unlike conventional cross-domain retrieval~\cite{huang2015cross}, where models are trained to use queries of a source domain and retrieve items from another target domain, we address a more practical, \emph{open-domain} setting, where the query and database may be from any unseen domain. We target different variants of this task, where the class of the query object is defined at \emph{category-level} or \emph{instance-level}. At the same time, the domain corresponds to descriptions of \emph{style} or \emph{context}, as shown in \autoref{fig:teaser}. Even though domain conversion is a subset of the tasks handled by existing CIR methods, the variants considered in our work reflect a more comprehensive set of applications than what was encountered in prior art~\cite{pic2word}.

Large pre-trained VLMs provide powerful representations of objects, domains, and their combinations. Our approach is \emph{training-free} by using a frozen VLM and performing textual inversion in a \emph{non-parametric} way, assuming access to an external large memory of words. Inversion maps images to the discrete input space of text instead of the continuous latent space of word tokens as in prior work~\cite{searle}. 
Compared to such an alternative, our memory-based inversion is more efficient and intuitive and comes with significant performance benefits. 
While our emphasis lies in domain conversion, the proposed approach is versatile and applicable to various composed image retrieval tasks, where its performance is competitive to the state-of-the-art approaches. We make the following contributions:

\begin{enumerate}[itemsep=2pt, parsep=0pt, topsep=3pt]
	\item We are the first to focus on composed image retrieval in the context of domain conversion, and we introduce three new benchmarks to the task while we extend an existing one to more source domains.
	\item We introduce \ours, a training-free CIR method for domain conversion that operates in an open world by inheriting the capabilities of a frozen CLIP model. 
	\item We demonstrate that textual inversion performs better in the discrete input space of known words than in the continuous latent space of pseudo-words.
	\item \ours outperforms all methods by a large margin on four benchmarks. 
 	\item Our experimental results form a testbed for future comparisons in this task.
\end{enumerate}

%% file: tex/related.tex
\section{Related work}
\label{sec:related}

\paragraph{Composed image retrieval (CIR).} Image-to-image~\cite{gar+16, nas+17} and text-to-image~\cite{sarafianos2019adversarial, zhang2020context, devise} retrieval provide useful ways to explore large image collections. Nevertheless, composed image retrieval offers more flexible ways to express the query and enables novel applications. TIRG~\cite{tirg} is the first to \emph{compose} image and text as a search query, where text serves as a modification of the image to refine the retrieval results. Training is supervised with cross-entropy loss, using triplets of the form \emph{reference image, query text, target image}. Following the same setting, JVSM~\cite{jvsm} learns image-text compositional embeddings in a unified space using multiple matching losses.

Other methods exploit attention in the form of multimodal disentangled non-local blocks~\cite{yin2020disentangled} to correlate text with image regions~\cite{lbf} extracted by an RPN~\cite{rhg+15}, to perform the composition at multiple depths~\cite{val}, to modulate content~\cite{cosmo}, and to discover the relation between composed query and target image~\cite{artemis}. DRA~\cite{dra} learns a dual relation, implicit/explicit, alignment network, while MPC~\cite{mpc} introduces a task variant with multiple queries. All these methods perform training from scratch and rely on triplets related to fashion~\cite{fashion200k, shoes, fashioniq}, physical states~\cite{mit_states}, object attributes, and object composition~\cite{tirg, lmb+14, mpc}. Labeling triplets is expensive and limits the broader use of CIR.

Inspired by vision-language foundation models~\cite{clip, align, bommasani2021opportunities}, recent work builds upon them in different ways. CIRPLANT~\cite{cirr} and FashionVLP~\cite{fashionvlp} extract features from a reference image as well as text features using a tokenizer~\cite{oscar, bert} and fine-tune the VLM~\cite{oscar, vinvl, clip} using triplets. CLIP4CIR~\cite{combiner} fine-tunes CLIP~\cite{clip} and trains a small network to combine image and text features using triplets. BLIP4CIR~\cite{liu2024bi} builds upon CLIP4CIR with BLIP~\cite{blip} and trains using reversed triplets, along with the original ones. SPRC~\cite{bai2023sentence} uses train triplets and trains a transformer model to combine the image and the text query.

To ensure scalability, datasets are often created through crowd-sourcing human text~\cite{cirr}, by exploiting LAION-5B~\cite{laion, liu2023zero} or VQA v2.0~\cite{goyal2017making, dataroaming}, or by automatically synthesizing millions of high-quality triplets~\cite{compodiff,jkm+24} using generative models~\cite{stablediffusion}. All these methods benefit from the compositional ability of vision-language models~\cite{oscar, vinvl, clip, blip} but still rely on triplets. Chen and Lai~\cite{cl24} train only with image-caption pairs and a masking technique that improves a simple inference-time baseline.

\picword~\cite{pic2word} relies on a VLM and is the first to avoid triplets and to evaluate on the domain conversion task. It follows self-supervised training to invert the query image to a text token. Thus, query composition takes place in the text domain by combining this token with the query text. KEDs \cite{smz+24} extends \picword by a retrieval augmentation process that uses an external image-caption database, while ISA \cite{du2023image2sentence} maps images to sentences in the form of multiple tokens. LinCIR \cite{gck+24} efficiently learns the inversion used by \picword but with language-only self-supervised training. In contrast to using an image-level representation in Pic2Word, Context-I2W~\cite{context_i2w} performs localization of the relevant image region to improve the inversion. 

\searle~\cite{searle} performs textual inversion with test-time optimization per query image. Then, a network is trained to imitate the result of such optimization to perform inference more efficiently. Our memory-based textual inversion avoids pre-training or test-time optimization.

\cirevl~\cite{cirevl} composes image and text queries solely in the language domain. It uses CLIP~\cite{clip} as an image and text encoder, BLIP-2~\cite{blip2} to caption the reference image, and GPT-3.5 turbo~\cite{gpt} to recompose the generated caption based on the query text. All backbones are pre-trained.

\magic~\cite{magiclens} is a co-current method that uses the implicit relations of images found on the same website to create triplets of query-image, instruction, and target-image. Then, they fine-tune a VLM that is expanded with four extra attention layers and one attention pooling layer.

\paragraph{Training-free use of VLMs.} The emergence of vision-language models (VLMs)~\cite{clip, align, blip, blip2} revolutionized the field of multimodal learning. Trained on massive datasets~\cite{laion}, these models have instrumental abilities to map images and text into a shared embedding space and are successful in training-free scenarios. MaskCLIP~\cite{maskclip} and CLIP-DIY~\cite{wrt+24} demonstrate the intrinsic potential of CLIP for semantic segmentation, while FLDM~\cite{fldm} highlights its effectiveness in text-guided video editing. The training-free paradigm extends to text-guided image editing~\cite{prompt2prompt} and layout control~\cite{chen2023training} using cross-attention. VLMs are also promising in specialized applications, such as deepfake detection~\cite{factor}, cross-domain image composition~\cite{tf_icon}, and phrase localization~\cite{li2022adapting}. Related to our training-free approach, CIReVL~\cite{cirevl} uses VLMs and LLMs to compose image and text queries in the language domain.

\paragraph{Cross-domain image retrieval (CDIR).} In this task, the query and database images come from different domains, and the challenge is bridging the domain gap~\cite{cross_domain_review}. As a visual domain, one might consider style~\cite{song2017deep}, color~\cite{fuentes2021sketch}, texture~\cite{song2017deep}, context~\cite{huang2015cross}, lighting conditions~\cite{huang2015cross}, or images captured using different sensors~\cite{he2018wasserstein}. One main line of research is in sketch-based image retrieval~\cite{hu2013performance, li2014fine, saavedra2014sketch, saavedra2015sketch, sangkloy2016sketchy, song2017deep, yu2016sketch}, and another on consumer scenarios~\cite{huang2015cross, ji2017cross, ibrahimi2019deep}.

Early methods do not generalize to new object classes or domains. This is the goal of zero-shot sketch-based retrieval~\cite{dey2019doodle, dutta2019semantically, yelamarthi2018zero}. More recent methods dispense with the need for labeled cross-modal pairs and are unsupervised~\cite{hu2021towards, kim2021cds, wang2023correspondence, ucdir}. Generalization to an unseen domain is only demonstrated by UCDR~\cite{paul2021universal}.
Nevertheless, no CDIR method includes the domain of the query image in the database, which becomes meaningful in our task, \ie, domain conversion with image-text queries.

%% file: tex/method.tex
\section{Method}
\label{sec:method}

\begin{figure*}
\begin{center}
\includegraphics[width=0.9\textwidth]{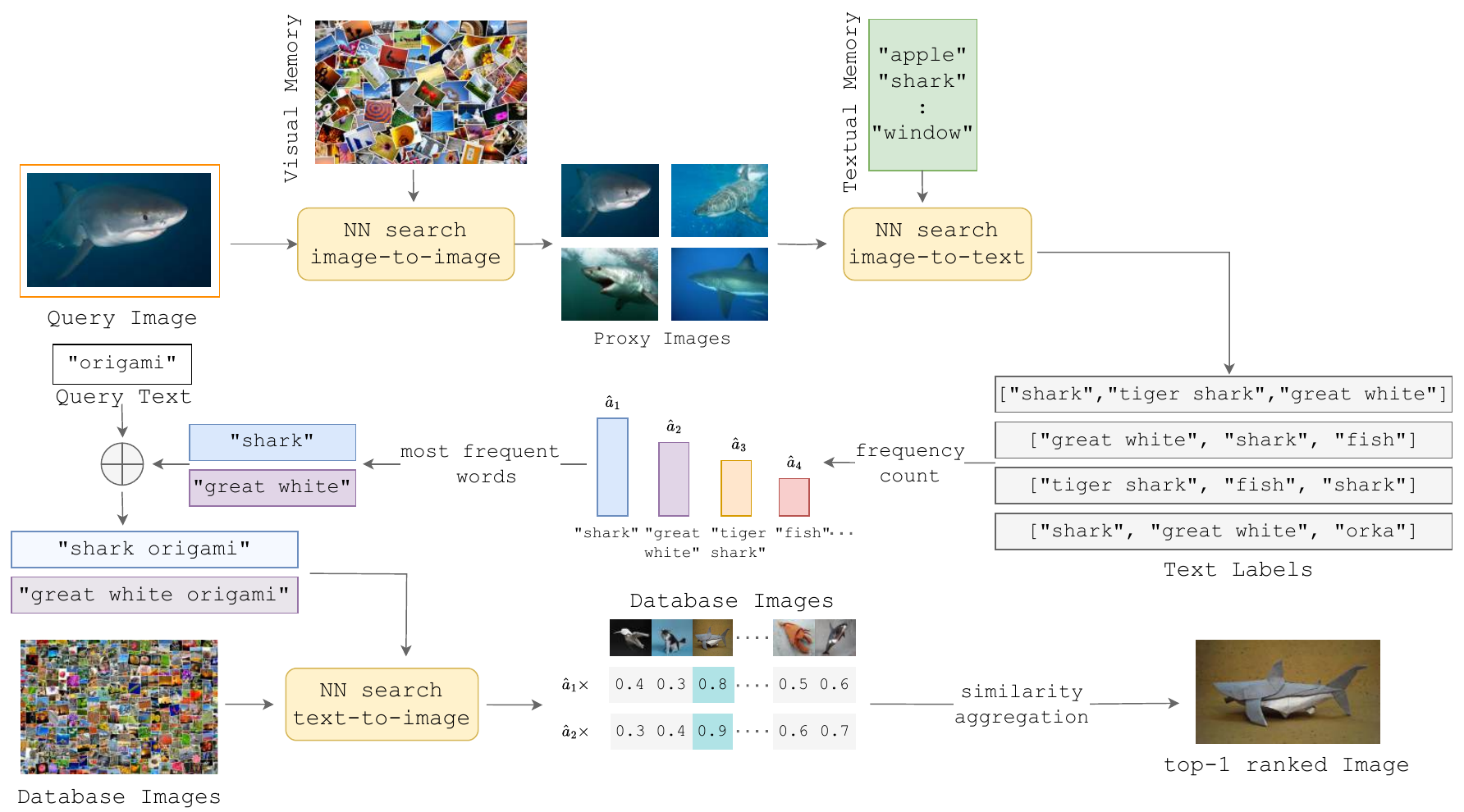}
\caption{
\emph{Overview} of \ours. Given a query image and a query text indicating the target domain, proxy images are first retrieved from the query through an image-to-image search over a visual memory. Then, a set of text labels is associated with each proxy image through an image-to-text search over a textual memory. Each of the most frequent text labels is combined with the query text in the text space, and images are retrieved from the database by text-to-image search. The resulting sets of similarities are linearly combined with the frequencies of occurrence as weights. Here: $k=4$ proxy images, $n=3$ text labels per proxy image, $m=2$ most frequent text labels.
\label{fig:method}
\vspace{-15pt}
}
\end{center}
\end{figure*}

\subsection{Preliminaries}
\label{sec:prelim}

Composed image retrieval is the task of retrieving images using a query that combines both a \emph{visual} component, represented by a query image, and a \emph{textual} component, represented by a query text. This work focuses on a specific variant of composed image retrieval that targets \emph{domain conversion}~\cite{pic2word}.

In particular, the query image $y$ depicts an object of \emph{class} $C(y)$ in the \emph{source domain} $D(y)$, while the query text $t$ represents the \emph{target domain}, $D(t)$. The two elements are jointly referred to as the \emph{composed query}, $q = (y, t)$. Given an image dataset $X$, the goal is to retrieve images from $X$ whose class is the same as that of the query image, $C(y)$, and whose domain is specified by the query text, $D(t)$. Retrieval amounts to ranking images $x \in X$ according to their \emph{composed similarity} $s(q, x) \in \real$ to the query $q$.

We rely on a pre-trained vision-language model that consists of a \emph{visual encoder} $f: \cI \to \real^d$ and a \emph{text encoder} $g: \cT \to \real^d$, which map respectively input images from the image space $\cI$ and words\footnote{With the term \emph{words}, we shall refer to both words and sentences.} from the text space $\cT$ to the same embedding space of dimension $d$. Using those encoders, a visual embedding $\vy = f(y) \in \real^d$ and a text embedding $\vt = g(t) \in \real^d$ are extracted for the query. Similarly, the embedding of an image $x \in X$ or a word $w$ are denoted by $\vx = f(x) \in \real^d$ and $\vw = g(w) \in \real^d$, respectively. All embeddings are $\ell_2$-normalized.

Given the visual-language model, this work aims to represent the composed query in the same embedding space as images and text. That is, define a \emph{composed encoder} $h: \cI \times \cT \to \real^d$ such that the composed query $q$ is mapped to $\vq = h(q) = h(y, t) \in \real^d$, again $\ell_2$-normalized. For any image $x \in X$, this allows us to express the composed similarity as the cosine similarity
\begin{equation}
	s(q, x) \defn h(q)\tran f(x) = h(y, t)\tran f(x).
\label{eq:sim}
\end{equation}
Thus, given the encoders $f, g$, the goal is to define $h$.

\subsection{Expanded textual inversion}
\label{sec:invert}

\paragraph{Textual inversion.} The ability of the text encoder, by design, to combine different concepts at its input and map them jointly to the embedding space motivates us to represent the composed query $q = (y, t)$ entirely in the text space $\cT$ and then map it to the embedding space, using $g$. The query text $t$ is already in $\cT$; but to map the query image $y$ to $w^* \in \cT$, it must first be embedded to $\vy = f(y)$ and then mapped from the embedding space back to $\cT$:
\begin{equation}
	w^* = g^{-1}(f(y)).
\label{eq:inv}
\end{equation}
The process of mapping the query image $y$ to text in $\cT$ is called \emph{textual inversion}, and the challenge is that the inverse mapping $g^{-1}$ is unknown.

Common approaches are \emph{pre-training}~\cite{pic2word} and \emph{test-time optimization}~\cite{searle}, both representing $w^*$ in the latent space of the vector tokens. The former defines a decoder and trains it on a dataset to \emph{learn} the inverse mapping $g^{-1}$ of the text encoder. Its challenge is the sheer scale of training required to reach anywhere close to the quality of the pre-trained text encoder $g$. The latter defines a variable at the input of $g$ and finds the optimal solution $w^*$ such that $g(w^*) = f(y)$. Its challenge is that there is a multitude of locally optimal solutions. Thus, the approach overly relies on the initialization of the variable, which remains unknown.

\paragraph{Memory-based inversion.} Contrary to existing approaches, the inversion is achieved by nearest neighbor search over an external vocabulary $V \subset \cT$ of words~\cite{open_images}, without training or optimization, and the $w^*$ is found in the discrete text space $\cT$ rather than the continuous latent space of vector tokens. 

In particular, if $V = \{ v_1, \dots, v_N \}$, vocabulary words $v_i$ are mapped to text embeddings $\vv_i = g(v_i)$ for $i = 1, \dots, N$. The \emph{text memory} of words $v_i$ and associated embeddings $\vv_i$ is the restriction $g|_V: V \to \cT$ of $g$ to $V$:

\begin{equation}
	g|_V : \{ v_1, \dots, v_N \} \rightarrow \{ \vv_1, \ldots, \vv_N \}.
\label{eq:voc}
\end{equation}

Given an embedding $\vv_i \in \cV = g(V) = \{ \vv_1, \dots, \vv_N \}$, the associated word $v_i$ can be instantly determined. This process essentially defines the restriction $g^{-1}|_\cV: \cV \to V$ of the otherwise unknown inverse $g^{-1}$ to $\cV$.
For brevity, we refer to $g^{-1}|_\cV$ as $g^{-1}$ in the following.

What remains is, given the query embedding $\vy \in \real^d$, to approximate it by one or more vectors in $\cV$. This is done by finding the $m$ 
nearest neighbors of $\vy$ in $\cV$,
\begin{equation}
	\cW = \{ \vw_1, \dots, \vw_m \} = \NN_m(\vy; \cV),
\label{eq:nn-v}
\end{equation}
given by descending order of (cosine) similarity. 
Since $\cW \subset \cV$, an embedding $\vw_i \in \cW$ can be mapped by $g^{-1}$ back to the associated word $w_i = g^{-1}(\vw_i)$ in $V$.
Thus, all neighbors are mapped to words
\begin{equation}
	W = \{ w_1, \dots, w_m \} = g^{-1}(\cW).
\label{eq:words-v}
\end{equation}
Putting everything together, this set of words is given by $W = \phi_V(y)$, where \emph{NN-inversion}
\begin{equation}
 \phi_V(y) \defn g^{-1}(\NN_m(f(y); g(V)))
\label{eq:inv-nn}
\end{equation}
is an approximation of~\eq{inv} by the vocabulary $V$.  The larger the vocabulary, the better the quality of approximation---but the more expensive the process. The function $\phi_V$~\eq{inv-nn} is used for defining different versions of composed encoder $h$ below.

Memory-based inversion is similar to zero-shot recognition, where the query image $y$ is represented by a set of words $W$ from the vocabulary $V$. We call the words found by $\phi_V(y)$ the \emph{\txlabels} or \emph{\tlabels} of $y$.

\paragraph{Single-word inversion.} The closest word $w_1$ to the query image is merged with the query text $t$ to form a composed query $w_1 \oplus t$ in the text space alone, where $\oplus$ denotes space-delimited string concatenation. Thus, $h$ becomes
\begin{equation}
	h_1(y, t) \defn \beta_1(\phi_V(y), t),
\label{eq:single}
\end{equation}
where $\phi_V$ is given by~\eq{inv-nn} and
\begin{equation}
	\beta_1(W, t) \defn g(w_1 \oplus t).
\label{eq:single-fun}
\end{equation}
A single word often works well, but using more words may help when the correct class is not top-ranked or when a collection of words represents better a particular image query.

\paragraph{Multi-word inversion: early fusion.}
We take advantage of the ability of the text encoder to combine several words in its input. Now, a composed query $w_1 \oplus \ldots \oplus w_m 
\oplus t$ is formed in the text space and $h$ becomes
\begin{equation}
	h_E(y, t) \defn \beta_E(\phi_V(y), t),
\label{eq:early}
\end{equation}
where $\phi_V$ is given by~\eq{inv-nn} and
\begin{equation}
\beta_E(W, t) \defn g(w_1 \oplus \cdots \oplus w_m \oplus t).
\label{eq:early-fun}
\end{equation}
We refer to this approach as \emph{early fusion} since the words $w_i$ are combined at the earliest possible stage.

\paragraph{Multi-word inversion: late fusion.}
Early fusion may be sensitive to words assigned incorrectly by NN-inversion $\phi_V$~\eq{inv-nn}. The other extreme is \emph{late fusion}, whereby words are composed at the latest possible stage. In particular, one composed query $w_i \oplus t$ is formed in the text space for each word $w_i$, is embedded separately, and a linear combination of these embeddings is formed. Thus, $h$ becomes
\begin{equation}
	h_L(y, t) \defn \beta_L(\phi_V(y), t),
\label{eq:late}
\end{equation}
where $\phi_V$ is given by~\eq{inv-nn},
\begin{equation}
	\beta_L(W, t, \va) \defn \sum_{i=1}^m a_i g(w_i \oplus t)
\label{eq:late-fun}
\end{equation}
and $a_i \in \real$ is a weight associated with word $w_i$, by default uniform $\va = \vone \in \real^m$. Because of the linearity of~\eq{sim}, this is equivalent to $m$ independent queries followed by a linear combination of the resulting similarities.

\paragraph{Memory-based expansion.}
Even if multiple nearest neighbors are used, the underlying cross-modal (image-to-text) similarity of~\eq{nn-v} remains challenging. 
A retrieval-based augmentation mechanism is employed to achieve more reliable zero-shot recognition. 
A \emph{visual memory} of images $z_i$ and the associated embeddings $\vz_i = f(z_i)$ from an external image set $Z$ are used. 
First, the query image is expanded through a set of $k$ 
\emph{proxy images} (including the query image), found as nearest neighbors of $\vy$ in the embeddings $\cZ = f(Z)$ based on unimodal (image-to-image) similarity:
\begin{equation}
	\cY = \{ \vy_1, \dots, \vy_k \} = \NN_k(\vy; \cZ).
\label{eq:nn-p}
\end{equation}
Then, following~\eq{nn-v}, for each proxy image $\vy_j$ in the embedding space, the $n$ nearest neighbors in $\cV$ and the associated words in the text space are found
\begin{equation}
	W_j = \{ w_{j1}, \dots, w_{jn} \} = g^{-1}(\NN_n(\vy_j; \cV)).
\label{eq:words-p}
\end{equation}
From the union $W^+ = \cup_{j=1}^k W_j$ with 
$\card{W^+} \le nk$ because of repeating words $w_{ji}$, the $m$ most frequent words $\hat{W} = \{\hat{w}_1, \dots, \hat{w}_m \}$ are selected and the $\hat{a}_i$ is defined as the frequency associated with word $\hat{w}_i$. We write this filtered set of words as $\hat{W} = \phi^+_{X,V}(y)$ as a function of the image query $y$, where $\phi^+_{X,V}$ is called \emph{expanded NN-inversion}. Finally, $h$ can be defined via either early fusion
\begin{equation}
	h_{E^{+}}(y, t) \defn \beta_E(\phi^+_{X,V}(y), t),
\label{eq:early-exp}
\end{equation}
where $\beta_E$ is given by \eq{early-fun}, or by late fusion
\begin{align}
	h_{L^{+}}(y, t)   & \defn \beta_L(\phi^+_{X,V}(y), t, \vone)  \mbox{\quad and} \label{eq:late-exp} \\
	h_{L^{+}_{\alpha}}(y, t) & \defn \beta_L(\phi^+_{X,V}(y), t, \hat{\va}),              \label{eq:late-exp-w}
\end{align}
with uniform and frequency weights $\vone, \hat{\va} \in \real^m$, respectively, and $\beta_L$ is given by \eq{late-fun}. The last expression~\eq{late-exp-w} is the complete \ours method, summarized in \autoref{fig:method}. 

%% file: tex/exp.tex
\section{Experiments}
\label{sec:exp}

\begin{table*}[t]
\centering
\label{tab:combined_sota}

\begin{minipage}{.5\textwidth}
\centering
\begin{subtable}{\textwidth}
\caption{\imagenetr}
\label{tab:imgnet_sota}
\input{tab/sota_imagenet_r}
\end{subtable}
\end{minipage}%
\begin{minipage}{.5\textwidth}
\centering
\begin{subtable}{\textwidth}
\caption{\minidn}
\label{tab:minidn_sota}
\input{tab/sota_minidn}
\end{subtable}
\end{minipage}

\vspace{0pt} 

\begin{minipage}{.5\textwidth}
\centering
\begin{subtable}{\textwidth}
\caption{\nico}
\label{tab:nico_sota}
\input{tab/sota_nico}
\end{subtable}
\end{minipage}%
\begin{minipage}{.5\textwidth}
\centering
\begin{subtable}{\textwidth}
\caption{\ltll}
\label{tab:leuven_sota}
\input{tab/sota_leuven}
\end{subtable}
\end{minipage}
\vspace{-5pt}
\caption{\emph{Domain conversion mAP (\%)} on four datasets. For each source domain (columns), we report the average mAP over every target domain. All methods are run by us. \Th{AVG}: average mAP over all source-target domain combinations.
\textbf{Bold:} best, \ts{magenta}: second-best. 
\label{tab:soa}
\vspace{-10pt}
}
\end{table*}

\subsection{Datasets, networks and evaluation protocol}
\label{sec:data}

We target a range of domain conversion applications where classes can be defined at the \emph{category level}~\cite{imagenet_r, minidomainnet, nico++} or \emph{instance level}~\cite{leuven}, and domains can be defined as \emph{styles}~\cite{imagenet_r, minidomainnet, leuven} or \emph{context/environment}~\cite{nico++}. 

\paragraph{Datasets for domain conversion.} ImageNet-R~\cite{imagenet_r}
has renditions of 200 \imagenet~\cite{dsl+09} classes comprising 30,000 images. Following \picword, we consider four domains: \emph{cartoon}, \emph{origami}, \emph{sculpture}, and \emph{toy}. We extend the benchmark by using every domain as the source.

MiniDomainNet~\cite{minidomainnet} is a subset of DomainNet~\cite{pbx19} with about 140,000 images, 126 classes, and four domains: \emph{clipart}, \emph{painting}, \emph{real}, and \emph{sketch}. Although this is a classification dataset, we adapt it for retrieval by using the official test set as our query set and the rest as the database.

Nico++~\cite{nico++} is an out-of-distribution classification dataset with $88,866$ real photographs, $60$ categories, and six domains: \emph{autumn}, \emph{dim light}, \emph{grass}, \emph{outdoor}, \emph{rock}, and \emph{water}. We adapt this dataset for retrieval by using a random $10\%$ as a query set and the rest as the database.

Large time lags location (LTLL)~\cite{leuven} contains images of $25$ locations taken over a range of more than $150$ years: $225$ historical and $275$ modern. We repurpose the dataset for retrieval by defining two domains: \emph{today} and \emph{archive}. 

\paragraph{Network.} We use the OpenAI pre-trained CLIP with a ViT-L/14 image encoder~\cite{dbk+21}.

\paragraph{Evaluation Protocol.} 
Unlike Precision@$k$ and Recall@$k$, commonly used in literature and focusing on specific points in a ranked list, mean Average Precision (mAP) provides a more comprehensive assessment by considering precision across the entire ranking. Thus, we choose mAP as our evaluation metric. We use Recall@$k$ only to compare our method with existing methods whose implementation or models are not publicly available.

\paragraph{Text/visual memory and \ours hyper-parameters.} 
We use the $20k$ words of the Open Images V7 dataset \cite{open_images} as our text memory. It is sufficiently large and is used in zero-shot recognition~\cite{cohen2022my} and by \searle~\cite{searle}, allowing direct comparison between the methods.
Unless otherwise stated, we use $k=20$, $n=7$, $m=7$. We use the image database as visual memory.
See the appendix for experiments with an external visual memory.

 \begin{table}
 \centering
 \input{tab/recall_table}
 \vspace{-5pt}
 \caption{\emph{Domain conversion Recall@$k$ (\%)} on \imagenetr. 
  \Th{AVG}: average performance over all target domains and ``photo'' as the source. 
  Top: methods that require training. Bottom: training-free methods.
  $^\dagger$: run by us. 
  \textbf{Bold:} best, \ts{magenta}: second-best. 
  \vspace{-10pt}}
 \label{tab:recall}
 \end{table}

\subsection{Simple Baselines}
\label{sec:base}
We include the following simple baselines for comparison. More advanced baselines can be found in the appendix.

\paragraph{Unimodal.}
Unimodal-query baselines rely only on similarity using one of the query modalities: \emph{text-only} by $h_T(y, t) := g(t)$ and \emph{visual-only} by $h_V(y, t) := f(y)$, referred to as ``Text'' and ``Image'' respectively in~\autoref{tab:soa}. Both are expected to fail as the final similarity misses aspects of the composed query.

\paragraph{Product.}
This baseline combines the two unimodal approaches by using the product of the corresponding similarities. It is referred to as ``Text $\times$ Image'' in~\autoref{tab:soa}.

 \footnotetext{The original work reports results in an incorrect way (recall defined differently~\cite{sxj+16}: equal to 1 if at least one relevant example is retrieved), which we confirm via code inspection and communication with the author.}

\paragraph{Sum.}
A common baseline in the literature that combines the two unimodal approaches by summation, \ie $h_S(y, t):= g(t) + f(y)$, referred to as ``Text + Image'' in~\autoref{tab:soa}. The problem is that the text and image embeddings follow very different distributions.
 
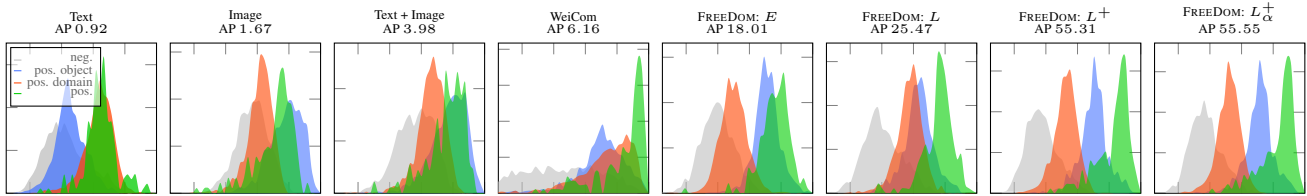
\begin{figure*}[t]
\centering
\input{fig/hist}
\vspace{-15pt}
\caption{\emph{Histogram of similarities} between a query and database images: \textcolor{lightgray}{negative} (wrong object and domain); \textcolor{RoyalBlue1}{positive} only \wrt the \textcolor{RoyalBlue1}{object} (correct object, wrong domain); \textcolor{OrangeRed1}{positive} only \wrt the \textcolor{OrangeRed1}{domain} (wrong object, correct domain); \textcolor{Green3}{positive} (correct object and domain). $E$: early fusion; $L$: late fusion; $L^{+}$: late fusion with memory-based expansion; $L^{+}_{\alpha}$: late fusion with memory-based expansion and frequencies as weights; AP: average precision. For better visualization, we sample an equal number of negatives, positives \wrt object, and positives \wrt domain, while the values in the histogram of positives are multiplied by 10. \minidn; text query: ``clipart''.
\label{fig:hist}
}
\vspace{-15pt}
\end{figure*}

\subsection{Competitors}
\label{sec:competitors}
We extensively compare \ours to recently proposed zero-shot composed image retrieval methods: \picword~\cite{pic2word}, \weicom~\cite{pke+24}, \compodiff~\cite{compodiff}, \searle~\cite{searle}, and \magic~\cite{magiclens}. The same ViT-L/14~\cite{dbk+21} image encoder is used for all competitors. \compodiff uses the text encoder of a ViT-G/14~\cite{dbk+21}. Although \ours uses the target domain's name as a textual query for every dataset, we tailor the textual query (prompt) per dataset for many of the competing methods. All methods are run and evaluated by us. Details for these methods are provided in the appendix.

\subsection{Experimental results}
\label{sec:results}

\paragraph{Comparison with SoTA.}
\ours outperforms all baselines and competitors (\autoref{tab:soa}) by a large margin. In particular, \ours outperforms the second-best method on \imagenetr by $15.87\%$ mAP, on \minidn by $14.32\%$, on \nico by $6.44\%$, and on \ltll by $6.64\%$. On \imagenetr, the second-best method is \searle. On \minidn, \compodiff and \searle are the second and third best, respectively, while \weicom performs lower than the simple baselines ``Text $\times$ Image'' and ``Text + Image''. On \nico, \searle and \magic are the only competitors that beat the simple baselines for more than $1\%$, while \picword performs lower than the simple baseline ``Text $\times$ Image''. On \ltll, \weicom is the second-best method. Interestingly, \magic is the best method for ``today'' as the source, but for ``archive'', it performs worse than the unimodal baseline ``Image''. The baseline ``Text $\times$ Image'' performs higher than \picword and \compodiff.

We conduct an additional experiment using SigLIP \cite{siglip} as a backbone. We get an mAP increase of 11.95, 5.72, 16.36, and 4.13 on \imagenetr, \nico, \minidn, and \ltll, respectively; see the appendix for details.

Additionally, in~\autoref{tab:recall}, we compare \ours with previous methods that evaluate with Recall@$k$ on \imagenetr.
\ours is the top-performing method among competitors that either requires training or is formed via an ensemble of several foundational models.

\paragraph{Qualitative analysis.}
\autoref{fig:hist} shows the histogram of similarities between a single query and the positives and negatives of different kinds of images. The unimodal baselines fail as expected. The sum baseline gives high importance to the image query. The normalized \weicom improves that to a small extent but performs poorly as well. Early fusion improves further but gives too much importance to the object since several \txlabels are merged with a single word for the domain. Late fusion significantly improves this imbalance. The use of proxy images and weighting further boosts the performance, visualized by the blue histogram moving to the right relative to the orange.

\subsection{Ablation study}
\label{sec:ablation}

\paragraph{Impact of hyper-parameters.}
\autoref{tab:nvsk} shows the impact of the number $k$ of proxy images and the number of nearest words $n$ from the vocabulary. The experiment is performed for a fixed number of \txlabels, $m=7$. None of the combinations with $k=1$ or $n=1$ is the best. Therefore, the two steps of the nearest neighbor search are meaningful. Note that $k=1$ corresponds to no expansion. The values $k=20$ and $n=7$ perform well across all datasets. In \autoref{tab:searelabla}, we show the impact of varying $m$, demonstrating the benefits of going beyond one \txlabel. The performance is stable for a large range of values.

\begin{table*}[t]
\centering
\input{tab/nvsk}
\vspace{-6pt}
\caption{\emph{The effect of the number of proxy images $k$ vs. the number of labels $n$ for each proxy} in \ours, measured in domain conversion mAP. Values $k$ $\in$ {\{10, 20}\} and $n$ $\in$ {\{7, 15}\} are better on average performance and competitive across datasets. Fixed $m=7$ used.
\vspace{-5pt}}
\label{tab:nvsk}
\end{table*}

\begin{table*}[t]
\centering
\vspace{-5pt}
\input{tab/ablations_w_searle}
\vspace{-6pt}
\caption{\emph{The impact of the number of selected \tlabels $m$} on each \ours component, measured in domain conversion mAP.
SRL: comparison to \searle that uses $m$ words from the vocabulary to guide the textual inversion optimization.
Components: $E$: early fusion; $L$: late fusion; $E^{+}$: early fusion with memory-based expansion; $L^{+}$ late fusion with memory-based expansion; $W$: late fusion with memory-based expansion and frequency-based weighting. Experiments with $k=20$ (if memory-based expansion is used) and $n=7$.}
\label{tab:searelabla}
\vspace{-10pt}
\end{table*}

\paragraph{Method components.}
In \autoref{tab:searelabla}, we show the impact of our components by adding them one by one, leading to the final method. Additionally, we compare to \searle, which performs test-time optimization of a learnable word token. This is a direct comparison between inversion in the discrete input space of text and the continuous latent space of word tokens.
\searle performs best if a single concept is used; concepts are similar to the \txlabels of our method. Our simplest variant with a single \txlabel performs better than \searle on three out of four datasets, showcasing the benefit of our textual inversion in discrete words. \ours benefits by using additional \txlabels and outperforms \searle on all datasets. Late fusion significantly boosts performance compared to early fusion, and so does expansion to proxy images. Using weights is beneficial on average and only slightly harmful on \ltll.

\subsection{Oracle experiments}
\label{sec:oracle}

\paragraph{Information injection}. 
In the inversion problem, a common challenge arises when the source domain appears within the \txlabels, causing conflicting domains in the query composition (\eg, ``cartoon origami''). Conversely, the correct query class or the source domain may not be found in the $m$ \txlabels. To study the impact of each, we conduct an oracle experiment and summarize the results in \autoref{tab:addomain}. In the first two columns (Object Gain), we compare early and late fusion with memory-based inversion after adding the name of the ground-truth class of the query to the $m$ \txlabels. Late fusion achieves almost twice the gain compared to early fusion, showing that it can benefit more from the correct information.
In the last two columns (Domain Loss), we compare the performance after including the source domain as a distractor in the \txlabels. Late fusion suffers almost half the loss compared to early fusion, showing more robust behavior to incorrect information.

\paragraph{Sensitivity to the vocabulary.}
We explore whether \ours strongly depends on having the most appropriate word for each query class in the vocabulary. To reflect that, we perform an oracle experiment where the name of the ground truth class of the query image is used to remove its $\ell$ nearest words from the vocabulary. After removing the $\ell=5$ most relevant words for each image query, \ours performs 23.52, 31.95, 23.58, and 30.81 mAP on \imagenetr, \minidn, \nico, and \ltll, respectively. Despite the lack of the most appropriate words, \ours is still the best method. For more details, see the appendix.

\paragraph{Performance upper-bound.} 
We use the ground-truth class of each image-query as a single \txlabel. In this case, \ours achieves 46.58, 34.00, 46.06, and 31.18 mAP on \imagenetr, \minidn, \nico, and \ltll, respectively, indicating space for improvement. However, this oracle experiment underperforms on \ltll, as class names do not always represent the object in this benchmark.

\begin{table}[t]
\centering
\vspace{-7pt}
\input{tab/ablations_add_domain}
\vspace{-6pt}
\caption{\emph{Oracle experiment} to study the impact of inliers and outliers in the $m$ \txlabels. The inlier represents the query object name, and the outlier represents the source domain name. Gain or loss by adding inliers or outliers in \txlabels, respectively, is reported.
Under memory-based expansion ($E^{+}$ and $L^{+}$), late fusion benefits more from inliers and is more robust to outliers.}
\vspace{-18pt}
\label{tab:addomain}

\end{table}

%% file: tab/sota_imagenet_r.tex
\centering
\scriptsize
\setlength{\tabcolsep}{5pt}
\renewcommand{\arraystretch}{.9}
\begin{tabular}{lrrrrrr}
\toprule
\Th{Method} &
\Th{Car}      &  
\Th{Ori}      &  
\Th{Pho}      &  
\Th{Scu}      &  
\Th{Toy}       &  
\Th{Avg}       \\
\midrule
Text      & 0.82  & 0.63  & 0.68  & 0.78  & 0.78   & 0.74  \\
Image     & 4.27  & 3.12  & 0.84  & 5.86  & 5.08  & 3.84  \\
Text $\times$ Image & 8.21 & 5.62 & 6.98 & 8.95 & 9.41 & 7.83 \\
Text + Image        & 6.61  & 4.45  & 2.17  & 9.18  & 8.62  & 6.21  \\ \midrule
\picword                  & 7.60  & 5.53  & 7.64  & 9.39  & 9.27  & 7.88  \\
\compodiff                & 13.71 & \ts{10.61} & 8.76 & 15.17 & \ts{16.17}  & 12.88  \\
\weicom                    & 10.07 & 7.61  & 10.06 & 11.26 & 13.38 & 10.47 \\
\searle           & \ts{18.11}  & 9.02  & 9.94 & \ts{17.26} & 15.83  & \ts{14.04}  \\
\magic & 7.79 & 6.33 & \ts{11.02} & 9.94 & 10.57 & 9.13 \\
\midrule \rowcolor{LightSteelBlue1}
\ours                      & \tb{35.97} & \tb{11.80} & \tb{27.97} & \tb{36.58} & \tb{37.21}  & \tb{29.91} \\ \bottomrule
\end{tabular}

%% file: tab/sota_minidn.tex
\centering
\scriptsize
\setlength{\tabcolsep}{7pt}
\renewcommand{\arraystretch}{.9}
\begin{tabular}{lrrrrr}
\toprule
\Th{Method} &
\Th{Clip}  &
\Th{Paint} &
\Th{Pho}    &
\Th{Ske}   &
\Th{Avg} \\ \midrule
Text                & 0.63  & 0.52  & 0.63  & 0.51    & 0.57  \\
Image               & 7.15  & 7.31  & 4.38  & 7.78    & 6.66  \\
Text $\times$ Image & 9.01  & 8.66  & 15.87 & 5.90    & 9.86 \\
Text + Image        & 9.59  & 9.97  & 9.22  & 8.53  & 9.33  \\ \midrule
\picword            & 13.39 & 8.63  & 17.96 & 8.03   & 12.00 \\
\compodiff          & 19.06 & \ts{24.27} & 23.41  & \ts{25.05}  & \ts{22.95} \\
\weicom             & 7.52  & 7.04  & 15.13 & 4.40   & 8.52  \\
\searle     & \ts{25.04} & 18.72 & 23.75 & 19.61  & 21.78 \\ 
\magic & 24.40 & 17.54 & \ts{28.59} & 9.71 & 20.06 \\
\midrule  \rowcolor{LightSteelBlue1}
\ours               & \tb{41.96} & \tb{31.65} & \tb{41.12} & \tb{34.36}  & \tb{37.27} \\ \bottomrule
\end{tabular}

%% file: tab/sota_nico.tex
\centering
\scriptsize
\setlength{\tabcolsep}{3.5pt}
\renewcommand{\arraystretch}{.9}
\begin{tabular}{lrrrrrrr}
\toprule
\Th{Method} &
\Th{Aut} &
\Th{Dim} &
\Th{Gra} &
\Th{Out} &
\Th{Roc} &
\Th{Wat} &
\Th{Avg} \\ \midrule
Text      & 1.00  & 0.99  & 1.15   & 1.23   & 1.10   & 1.05   & 1.09  \\
Image     & 6.45  & 4.85  & 5.67   & 7.67   & 7.65   & 5.65    & 6.32  \\ 
Text $\times$ Image & 8.24 & 6.36 & 12.11 & 12.71 & 10.46 & 8.84 & 9.79 \\
Text + Image        & 8.46 & 6.58 & 9.22  & 11.91 & 11.20 & 8.41   & 9.30  \\
\midrule
\picword                  & 9.79  & 8.09  & 11.24  & 11.27  & 11.01  & 7.16    & 9.76  \\
\compodiff                 & 10.07 & 7.83  & 10.53  & 11.41  & 11.93 & 10.15   & 10.32  \\
\weicom    & 8.58  & 7.39  & 13.04  & 13.17  & 11.32  & 9.73   & 10.54 \\ 
\searle         & 13.49 & 13.73 & 17.91  & 17.99  & 15.79  & 11.84  & 15.13 \\ 
\magic            & \ts{18.76} & \ts{15.17} & \ts{22.14} & \ts{23.61} & \ts{21.99} & \ts{16.30} & \ts{19.66} \\
\midrule \rowcolor{LightSteelBlue1}
\ours                      & \tb{24.35} & \tb{24.41} & \tb{30.06}  & \tb{30.51}  & \tb{26.92}  & \tb{20.37} & \tb{26.10} \\ \bottomrule
\end{tabular}

%% file: tab/sota_leuven.tex
\centering
\scriptsize
\setlength{\tabcolsep}{12pt}
\renewcommand{\arraystretch}{.9}
\begin{tabular}{lrrr}
\toprule
\Th{Method} &
\Th{Today} & \Th{Archive} & \Th{Avg} \\ \midrule
Text    & 5.28  & 6.16  & 5.72  \\
Image     & 8.47  & 24.51 & 16.49 \\ 
Text $\times$ Image & 16.42 & 29.90 & 23.16 \\ 
Text + Image        & 9.60  & 26.13 & 17.86 \\ \midrule
\picword                  & 17.86 & 24.67 & 21.27 \\
\compodiff                & 15.45 & 27.76 & 21.61 \\
\weicom                    & 24.56 & 28.63 & \ts{26.60} \\
\searle          & 20.82 & \ts{30.10} & 25.46 \\ 
\magic  & \tb{33.77} & 14.65 & 24.21 \\
\midrule \rowcolor{LightSteelBlue1}
\ours                      & \ts{30.95} & \tb{35.52} & \tb{33.24} \\ \bottomrule
\end{tabular}

%% file: tab/recall_table.tex
\scriptsize
\renewcommand{\arraystretch}{.9}
\begin{tabular}{lrrr} \toprule
\multirow{2}{*}{\Th{Method}} & \multirow{2}{*}{\Th{\# param}} & \multicolumn{2}{c}{\Th{Avg}} \\ \cmidrule{3-4} &
& \Th{R@10}  & \Th{R@50}  \\ \midrule
\picword    \cite{pic2word}              & 429M     & 10.05 & 23.23 \\
\picword  (CC-3M) \cite{pic2word}        & 429M     & 10.19 & 22.70 \\
\picword  (LAION 2B-en) \cite{pic2word}   & 429M    & 10.35 & 22.91 \\
ARTEMIS w/ CompoDiff dataset \cite{artemis} & 33M & 12.24 & 23.58 \\
CLIP4Cir w/ CompoDiff dataset \cite{combiner} & 178M & 12.11 & 23.79 \\
\compodiff (T5-XL) \cite{compodiff}        & 3B   & 11.14 & 21.76 \\
\compodiff (CLIP+T5-XL) \cite{compodiff}   & 3.6B   & 13.31 & 24.86 \\
\compodiff (CLIP)   \cite{compodiff}       & 568M   & 13.18 & 24.86 \\
Context-I2W \cite{context_i2w}             & 496M   & 12.90  & 27.60  \\
KEDs ~\cite{smz+24}                       & 428M  & 18.00  & 35.40  \\
\magic\footnotemark (original prompt)$^{\dagger}$~\cite{magiclens}        & 465M   & 8.61 & 21.60 \\
\magic$^\dagger$ ~\cite{magiclens}        & 465M  & 11.34 & 26.78 \\
\midrule
\weicom$^\dagger$ ~\cite{pke+24}                    & 428M        & 12.17 & 23.04 \\ 
\searle (default)$^\dagger$~\cite{searle}           & 428M  & 2.36  & 14.24 \\
\searle$^\dagger$~\cite{searle}            & 428M   & 11.76 & 30.96 \\ 
CIReVL ~\cite{cirevl}                     & 12.5B   & \ts{23.75}  & \ts{43.05} \\ \midrule
\rowcolor{LightSteelBlue1}
\ours                                      & 428M   & \textbf{27.54} & \textbf{47.05} \\ \bottomrule
\end{tabular}

%% file: fig/hist.tex
\input{fig/pgfplotsdata}
\pgfplotsset{
compat=1.11,
legend image code/.code={
\draw[mark repeat=2,mark phase=2]
plot coordinates {
(0cm,0cm)
(0.15cm,0cm)        
(0.3cm,0cm)        
};
}
}
\footnotesize
\begin{tabular}{@{\nssp}c@{\nssp}c@{\nssp}c@{\nssp}c@{\nssp}c@{\nssp}c@{\nssp}c@{\nssp}c@{\nssp}}
\begin{tikzpicture}
\begin{axis}[
    width=0.205\linewidth,
    height=0.205\linewidth,
    xmax = 1.00,
    xmin = 0.05,
    ymin = 0.0,
    grid=none,
    yticklabels={,,},
    xticklabels={,,},
    title style={align=center, yshift=-5pt},
    legend pos= {north west},
    legend style={cells={anchor=east}, font =\tiny, fill opacity=0.6, row sep=-4pt,inner sep=1pt,outer sep=0pt,legend image code/.code={\draw[mark repeat=2,mark phase=2,##1] (0cm,0cm) -- (0.1cm,0cm);}},
    title={\tiny Text \\[1pt] \tiny AP $0.92$},
]
\addplot[smooth,color=lightgray, fill = lightgray,fill opacity=0.75,draw=none,bar width = 0.045] table[x expr=\thisrow{bins}, y expr=\thisrow{fp_y}]  \histtext; \addlegendentry{neg.}
\addplot[smooth,color=RoyalBlue1, fill = RoyalBlue1,fill opacity=0.75,draw=none,bar width = 0.045] table[x expr=\thisrow{bins}, y expr=\thisrow{class_y}]  \histtext; \addlegendentry{pos. object}
\addplot[smooth,color=OrangeRed1, fill = OrangeRed1,fill opacity=0.75,draw=none,bar width = 0.045] table[x expr=\thisrow{bins}, y expr=\thisrow{domain_y}]  \histtext; \addlegendentry{pos. domain}
\addplot[smooth,color=Green3, fill = Green3,fill opacity=0.75,draw=none,bar width = 0.045] table[x expr=\thisrow{bins}, y expr=\thisrow{tp_y}]  \histtext; \addlegendentry{pos.}
\end{axis}
\end{tikzpicture}
&
\begin{tikzpicture}
\begin{axis}[
    width=0.205\linewidth,
    height=0.205\linewidth,
    xmax = 1.00,
    xmin = 0.05,
    ymin = 0.0,
    grid=none,
    yticklabels={,,},
    xticklabels={,,},
    title style={align=center, yshift=-5pt},
    legend pos= {north west},
    legend style={cells={anchor=east}, font =\scriptsize, fill opacity=0.8, row sep=-2.5pt},
    title={\tiny Image \\[1pt] \tiny AP $1.67$},
]
\addplot[smooth,color=lightgray, fill = lightgray,fill opacity=0.6,draw=none,bar width = 0.045] table[x expr=\thisrow{bins}, y expr=\thisrow{fp_y}]  \histimage; 
\addplot[smooth,color=RoyalBlue1, fill = RoyalBlue1,fill opacity=0.6,draw=none,bar width = 0.045] table[x expr=\thisrow{bins}, y expr=\thisrow{class_y}]  \histimage; 
\addplot[smooth,color=OrangeRed1, fill = OrangeRed1,fill opacity=0.6,draw=none,bar width = 0.045] table[x expr=\thisrow{bins}, y expr=\thisrow{domain_y}]  \histimage; 
\addplot[smooth,color=Green3, fill = Green3,fill opacity=0.6,draw=none,bar width = 0.045] table[x expr=\thisrow{bins}, y expr=\thisrow{tp_y}]  \histimage; 
\end{axis}
\end{tikzpicture}
&
\begin{tikzpicture}
\begin{axis}[
    width=0.205\linewidth,
    height=0.205\linewidth,
    xmax = 1.00,
    xmin = 0.05,
    ymin = 0.0,
    grid=none,
    yticklabels={,,},
    xticklabels={,,},
    title style={align=center, yshift=-5pt},
    legend pos= {north west},
    legend style={cells={anchor=east}, font =\scriptsize, fill opacity=0.8, row sep=-2.5pt},
    title={\tiny Text + Image \\[1pt] \tiny AP $3.98$},
]
\addplot[smooth,color=lightgray, fill = lightgray,fill opacity=0.6,draw=none,bar width = 0.045] table[x expr=\thisrow{bins}, y expr=\thisrow{fp_y}]  \histaddsum; 
\addplot[smooth,color=RoyalBlue1, fill = RoyalBlue1,fill opacity=0.6,draw=none,bar width = 0.045] table[x expr=\thisrow{bins}, y expr=\thisrow{class_y}]  \histaddsum; 
\addplot[smooth,color=OrangeRed1, fill = OrangeRed1,fill opacity=0.6,draw=none,bar width = 0.045] table[x expr=\thisrow{bins}, y expr=\thisrow{domain_y}]  \histaddsum; 
\addplot[smooth,color=Green3, fill = Green3,fill opacity=0.6,draw=none,bar width = 0.045] table[x expr=\thisrow{bins}, y expr=\thisrow{tp_y}]  \histaddsum; 
\end{axis}
\end{tikzpicture}
&
\begin{tikzpicture}
\begin{axis}[
    width=0.205\linewidth,
    height=0.205\linewidth,
    xmax = 1.00,
    xmin = 0.05,
    ymin = 0.0,
    grid=none,
    yticklabels={,,},
    xticklabels={,,},
    title style={align=center, yshift=-5pt},
    legend pos= {north west},
    legend style={cells={anchor=east}, font =\scriptsize, fill opacity=0.8, row sep=-2.5pt},
    title={\tiny \weicom \\[1pt] \tiny AP $6.16$},
]
\addplot[smooth,color=lightgray, fill = lightgray,fill opacity=0.6,draw=none,bar width = 0.045] table[x expr=\thisrow{bins}, y expr=\thisrow{fp_y}]  \histaddsumnorm; 
\addplot[smooth,color=RoyalBlue1, fill = RoyalBlue1,fill opacity=0.6,draw=none,bar width = 0.045] table[x expr=\thisrow{bins}, y expr=\thisrow{class_y}]  \histaddsumnorm; 
\addplot[smooth,color=OrangeRed1, fill = OrangeRed1,fill opacity=0.6,draw=none,bar width = 0.045] table[x expr=\thisrow{bins}, y expr=\thisrow{domain_y}]  \histaddsumnorm; 
\addplot[smooth,color=Green3, fill = Green3,fill opacity=0.6,draw=none,bar width = 0.045] table[x expr=\thisrow{bins}, y expr=\thisrow{tp_y}]  \histaddsumnorm; 
\end{axis}
\end{tikzpicture}
&
\begin{tikzpicture}
\begin{axis}[
    width=0.205\linewidth,
    height=0.205\linewidth,
    xmax = 1.00,
    xmin = 0.05,
    ymin = 0.0,
    grid=none,
    yticklabels={,,},
    xticklabels={,,},
    title style={align=center, yshift=-5pt},
    legend pos= {north west},
    legend style={cells={anchor=east}, font =\scriptsize, fill opacity=0.8, row sep=-2.5pt},
    title={\tiny \ours: $E$ \\[1pt] \tiny AP $18.01$},
]
\addplot[smooth,color=lightgray, fill = lightgray,fill opacity=0.6,draw=none,bar width = 0.045] table[x expr=\thisrow{bins}, y expr=\thisrow{fp_y}]  \histearlyfusion; 
\addplot[smooth,color=RoyalBlue1, fill = RoyalBlue1,fill opacity=0.6,draw=none,bar width = 0.045] table[x expr=\thisrow{bins}, y expr=\thisrow{class_y}]  \histearlyfusion; 
\addplot[smooth,color=OrangeRed1, fill = OrangeRed1,fill opacity=0.6,draw=none,bar width = 0.045] table[x expr=\thisrow{bins}, y expr=\thisrow{domain_y}]  \histearlyfusion; 
\addplot[smooth,color=Green3, fill = Green3,fill opacity=0.6,draw=none,bar width = 0.045] table[x expr=\thisrow{bins}, y expr=\thisrow{tp_y}]  \histearlyfusion; 
\end{axis}
\end{tikzpicture}
&
\begin{tikzpicture}
\begin{axis}[
    width=0.205\linewidth,
    height=0.205\linewidth,
    xmax = 1.00,
    xmin = 0.05,
    ymin = 0.0,
    grid=none,
    yticklabels={,,},
    xticklabels={,,},
    title style={align=center, yshift=-5pt},
    legend pos= {north west},
    legend style={cells={anchor=east}, font =\scriptsize, fill opacity=0.8, row sep=-2.5pt},
    title={\tiny \ours: $L$ \\[1pt] \tiny AP $25.47$},
]
\addplot[smooth,color=lightgray, fill = lightgray,fill opacity=0.6,draw=none,bar width = 0.045] table[x expr=\thisrow{bins}, y expr=\thisrow{fp_y}]  \histlatefusion; 
\addplot[smooth,color=RoyalBlue1, fill = RoyalBlue1,fill opacity=0.6,draw=none,bar width = 0.045] table[x expr=\thisrow{bins}, y expr=\thisrow{class_y}]  \histlatefusion; 
\addplot[smooth,color=OrangeRed1, fill = OrangeRed1,fill opacity=0.6,draw=none,bar width = 0.045] table[x expr=\thisrow{bins}, y expr=\thisrow{domain_y}]  \histlatefusion; 
\addplot[smooth,color=Green3, fill = Green3,fill opacity=0.6,draw=none,bar width = 0.045] table[x expr=\thisrow{bins}, y expr=\thisrow{tp_y}]  \histlatefusion; 
\end{axis}
\end{tikzpicture}
&
\begin{tikzpicture}
\begin{axis}[%
    width=0.205\linewidth,
    height=0.205\linewidth,
    xmax = 1.00,
    xmin = 0.05,
    ymin = 0.0,
    grid=none,
    yticklabels={,,},
    xticklabels={,,},
    title style={align=center, yshift=-5pt},
    legend pos= {north west},
    legend style={cells={anchor=east}, font =\scriptsize, fill opacity=0.8, row sep=-2.5pt},
    title={\tiny \ours: $L^{+}$ \\[1pt] \tiny AP $55.31$},
]
\addplot[smooth,color=lightgray, fill = lightgray,fill opacity=0.6,draw=none,bar width = 0.045] table[x expr=\thisrow{bins}, y expr=\thisrow{fp_y}]  \histlatefusionclean;
\addplot[smooth,color=RoyalBlue1, fill = RoyalBlue1,fill opacity=0.6,draw=none,bar width = 0.045] table[x expr=\thisrow{bins}, y expr=\thisrow{class_y}]  \histlatefusionclean; 
\addplot[smooth,color=OrangeRed1, fill = OrangeRed1,fill opacity=0.6,draw=none,bar width = 0.045] table[x expr=\thisrow{bins}, y expr=\thisrow{domain_y}]  \histlatefusionclean; 
\addplot[smooth,color=Green3, fill = Green3,fill opacity=0.6,draw=none,bar width = 0.045] table[x expr=\thisrow{bins}, y expr=\thisrow{tp_y}]  \histlatefusionclean;
\end{axis}
\end{tikzpicture}
&
\begin{tikzpicture}
\begin{axis}[
    width=0.205\linewidth,
    height=0.205\linewidth,
    xmax = 1.00,
    xmin = 0.05,
    ymin = 0.0,
    grid=none,
    yticklabels={,,},
    xticklabels={,,},
    title style={align=center, yshift=-5pt},
    legend pos= {north west},
    legend style={cells={anchor=east}, font =\scriptsize, fill opacity=0.8, row sep=-2.5pt},
    title={\tiny \ours: $L^{+}_{\alpha}$ \\[1pt] \tiny AP $55.55$},
]
\addplot[smooth,color=lightgray, fill = lightgray,fill opacity=0.6,draw=none,bar width = 0.045] table[x expr=\thisrow{bins}, y expr=\thisrow{fp_y}]  \histlatefusioncleanweight; 
\addplot[smooth,color=RoyalBlue1, fill = RoyalBlue1,fill opacity=0.6,draw=none,bar width = 0.045] table[x expr=\thisrow{bins}, y expr=\thisrow{class_y}]  \histlatefusioncleanweight; 
\addplot[smooth,color=OrangeRed1, fill = OrangeRed1,fill opacity=0.6,draw=none,bar width = 0.045] table[x expr=\thisrow{bins}, y expr=\thisrow{domain_y}]  \histlatefusioncleanweight; 
\addplot[smooth,color=Green3, fill = Green3,fill opacity=0.6,draw=none,bar width = 0.045] table[x expr=\thisrow{bins}, y expr=\thisrow{tp_y}]  \histlatefusioncleanweight; 
\end{axis}
\end{tikzpicture}
\end{tabular}

%% file: fig/pgfplotsdata.tex
\pgfplotstableread{
bins	domain_y	class_y	tp_y	fp_y
0.09188840538263321 0.0 0.0 0.0 0.08000000000000002
0.1168884053826332 0.0 0.03384094754653131 0.0 0.08000000000000002
0.1418884053826332 0.0 0.03384094754653131 0.0 0.12000000000000002
0.1668884053826332 0.0 0.10152284263959394 0.0 0.2800000000000001
0.1918884053826332 0.0 0.13536379018612524 0.0 0.6200000000000001
0.21688840538263318 0.020010005002501254 0.23688663282571917 0.0 0.8800000000000002
0.24188840538263318 0.020010005002501278 0.3722504230118448 0.0 1.2600000000000016
0.26688840538263314 0.020010005002501233 0.47377326565143785 0.21978021978021958 1.9399999999999982
0.29188840538263316 0.14007003501750864 0.7783417935702193 0.0 2.1999999999999984
0.3168884053826332 0.10005002501250639 1.2182741116751283 0.21978021978022008 2.760000000000004
0.34188840538263315 0.16008004002001022 1.6243654822335047 0.0 2.8600000000000034
0.3668884053826331 0.16008004002000986 2.6395939086294393 0.0 3.319999999999997
0.39188840538263314 0.24012006003001482 3.6548223350253775 0.0 2.839999999999997
0.41688840538263316 0.24012006003001532 4.737732656514389 0.8791208791208803 3.2600000000000047
0.44188840538263313 0.46023011505752937 5.346869712351952 0.0 2.8600000000000034
0.4668884053826331 0.6003001500750369 3.925549915397627 0.8791208791208783 2.4399999999999977
0.4918884053826331 0.8804402201100542 3.5532994923857837 0.21978021978021958 1.9399999999999982
0.5168884053826331 1.5007503751875924 2.504230118443314 0.6593406593406588 1.6799999999999984
0.5418884053826332 2.0410205102551346 2.3350253807106682 0.8791208791208822 1.5800000000000056
0.5668884053826331 2.4812406203101527 1.86125211505922 3.2967032967032934 0.8999999999999992
0.5918884053826331 3.4217108554277105 1.725888324873095 3.0769230769230744 1.0799999999999992
0.6168884053826331 4.502251125562798 0.8798646362098169 5.714285714285734 1.0600000000000038
0.641888405382633 5.262631315657824 0.744500846023688 4.83516483516483 0.9199999999999992
0.666888405382633 4.622311155577784 0.47377326565143785 6.373626373626367 0.7399999999999993
0.6918884053826331 4.402201100550272 0.20304568527918765 3.5164835164835133 0.9599999999999992
0.7168884053826331 3.6418209104552246 0.2707275803722502 3.2967032967032934 0.5799999999999994
0.7418884053826331 2.4612306153076626 0.13536379018612568 1.9780219780219852 0.3400000000000012
0.766888405382633 1.360680340170084 0.0 0.21978021978021958 0.2399999999999998
0.791888405382633 0.6603301650825407 0.0 0.6593406593406588 0.13999999999999987
0.8168884053826331 0.2801400700350185 0.0 0.8791208791208822 0.0
0.841888405382633 0.18009004502251108 0.0 0.6593406593406588 0.0
0.866888405382633 0.040020010005002465 0.0 0.8791208791208783 0.039999999999999966
0.891888405382633 0.060030015007503705 0.0 0.0 0.0
0.916888405382633 0.020010005002501233 0.0 0.21978021978021958 0.0
0.9418884053826331 0.0 0.0 0.4395604395604411 0.0
0.966888405382633 0.020010005002501233 0.0 0.0 0.0
1.0 0.0 0.0 0.0 0.0
}{\histtext}

\pgfplotstableread{
bins domain_y class_y tp_y fp_y
0.11658147722482681 0.020000000000000004 0.0 0.0 0.020000000000000004
0.1415814772248268 0.020000000000000004 0.0 0.0 0.020000000000000004
0.1665814772248268 0.020000000000000004 0.0 0.0 0.020000000000000004
0.1915814772248268 0.04000000000000001 0.0 0.0 0.0
0.2165814772248268 0.04000000000000001 0.0 0.21978021978021983 0.020000000000000004
0.24158147722482678 0.04000000000000001 0.03392705682782019 0.0 0.020000000000000004
0.2665814772248268 0.06000000000000008 0.0 0.21978021978022008 0.040000000000000056
0.29158147722482675 0.07999999999999993 0.0 0.21978021978021958 0.1199999999999999
0.31658147722482677 0.07999999999999993 0.03392705682782016 0.0 0.2399999999999998
0.3415814772248268 0.1400000000000002 0.0 0.0 0.48000000000000065
0.36658147722482676 0.06000000000000008 0.03392705682782023 0.43956043956044016 0.6000000000000008
0.3915814772248267 0.2399999999999998 0.06785411365564031 0.0 0.8599999999999992
0.41658147722482675 0.2999999999999997 0.03392705682782016 0.6593406593406588 1.0199999999999991
0.44158147722482677 0.6400000000000009 0.06785411365564047 0.21978021978022008 1.7200000000000022
0.46658147722482674 0.6200000000000008 0.16963528413910114 0.6593406593406602 2.0400000000000027
0.4915814772248267 1.279999999999999 0.27141645462256125 0.21978021978021958 3.199999999999997
0.5165814772248267 1.3999999999999988 0.20356234096692094 0.6593406593406588 3.419999999999997
0.5415814772248267 2.239999999999998 0.33927056827820157 1.5384615384615372 3.3599999999999968
0.5665814772248268 3.740000000000013 0.542832909245125 0.6593406593406617 3.8400000000000136
0.5915814772248267 4.1199999999999966 0.44105173876166204 1.5384615384615372 3.8799999999999963
0.6165814772248267 5.7999999999999945 1.0517387616624247 1.3186813186813175 3.8999999999999964
0.6415814772248267 5.74000000000002 1.7981340118744764 1.7582417582417644 3.1600000000000112
0.6665814772248266 5.059999999999995 1.8659881255301087 1.978021978021976 2.3199999999999976
0.6915814772248267 3.7399999999999967 2.13740458015267 1.978021978021976 2.0199999999999982
0.7165814772248267 2.639999999999998 2.7480916030534326 3.5164835164835133 1.5799999999999987
0.7415814772248267 1.3599999999999988 2.815945716709073 5.05494505494505 1.0399999999999991
0.7665814772248267 0.4400000000000016 3.223070398642929 5.274725274725293 0.5400000000000019
0.7915814772248266 0.039999999999999966 3.867684478371498 3.956043956043952 0.37999999999999967
0.8165814772248267 0.0 3.9355385920271377 3.736263736263733 0.1199999999999999
0.8415814772248267 0.0 3.4944868532654914 2.6373626373626466 0.02000000000000007
0.8665814772248266 0.0 3.5284139100932963 1.0989010989010979 0.0
0.8915814772248266 0.0 3.019508057675994 0.0 0.0
0.9165814772248266 0.0 2.815945716709073 0.43956043956043916 0.0
0.9415814772248267 0.0 1.1195928753180653 0.0 0.0
0.9665814772248267 0.0 0.33927056827820307 0.0 0.0
1.0     0.0     0.0     0.0     0.0     
}{\histimage}

\pgfplotstableread{
bins domain_y class_y tp_y fp_y
0.1030275896191597 0.020000000000000004 0.0 0.0 0.020000000000000004
0.1280275896191597 0.0 0.0 0.0 0.0
0.1530275896191597 0.020000000000000004 0.0 0.0 0.06000000000000001
0.17802758961915968 0.04000000000000001 0.0 0.0 0.020000000000000004
0.20302758961915968 0.020000000000000004 0.0 0.0 0.0
0.22802758961915967 0.0 0.03395585738539899 0.0 0.12000000000000002
0.25302758961915967 0.06000000000000008 0.0 0.0 0.20000000000000026
0.27802758961915963 0.0 0.03395585738539895 0.0 0.2399999999999998
0.30302758961915965 0.039999999999999966 0.0 0.4419889502762427 0.4799999999999996
0.3280275896191597 0.020000000000000028 0.06791171477079806 0.0 0.44000000000000056
0.35302758961915964 0.020000000000000028 0.03395585738539903 0.0 0.6000000000000008
0.3780275896191596 0.15999999999999986 0.03395585738539895 0.0 0.9199999999999992
0.40302758961915963 0.19999999999999982 0.10186757215619686 0.22099447513812134 1.299999999999999
0.42802758961915965 0.2800000000000004 0.10186757215619707 0.22099447513812187 1.8400000000000025
0.4530275896191596 0.24000000000000032 0.06791171477079806 0.44198895027624374 2.400000000000003
0.4780275896191596 0.5599999999999995 0.2716468590831916 0.0 2.4199999999999977
0.5030275896191596 0.5999999999999994 0.3395585738539895 0.4419889502762427 2.7999999999999976
0.5280275896191596 1.139999999999999 0.2716468590831916 1.1049723756906067 2.9999999999999973
0.5530275896191597 1.1600000000000041 0.6112054329371839 0.4419889502762447 2.580000000000009
0.5780275896191596 2.259999999999998 0.6451612903225801 0.8839779005524854 2.9799999999999973
0.6030275896191596 3.219999999999997 0.9847198641765695 0.6629834254143641 3.239999999999997
0.6280275896191596 4.340000000000016 1.324278438030565 1.325966850828734 2.9200000000000106
0.6530275896191595 4.739999999999996 1.561969439728352 0.8839779005524854 2.9599999999999973
0.6780275896191595 5.219999999999995 2.037351443123937 1.7679558011049707 2.839999999999997
0.7030275896191596 4.439999999999996 2.4448217317487244 2.8729281767955777 2.3599999999999977
0.7280275896191596 4.239999999999997 2.5466893039049214 1.5469613259668495 1.4799999999999986
0.7530275896191596 3.280000000000012 3.1918505942275153 4.640883977900569 1.0800000000000038
0.7780275896191595 2.0399999999999983 3.531409168081491 3.7569060773480634 0.3399999999999997
0.8030275896191595 1.1599999999999988 3.735144312393885 3.5359116022099415 0.19999999999999982
0.8280275896191596 0.3400000000000012 3.735144312393901 4.640883977900569 0.1400000000000005
0.8530275896191595 0.13999999999999987 3.56536502546689 3.7569060773480634 0.0
0.8780275896191595 0.0 3.837011884550082 4.198895027624306 0.0
0.9030275896191595 0.0 2.716468590831916 1.3259668508287281 0.019999999999999983
0.9280275896191595 0.0 1.4601018675721549 0.8839779005524854 0.0
0.9530275896191596 0.0 0.7130730050933811 0.0 0.0
1.0     0.0     0.0     0.0     0.0    
}{\histaddsum}

\pgfplotstableread{
bins domain_y class_y tp_y fp_y
-0.015090434 0.0 0.0 0.0 0.0
0.009909565560519695 0.02002002002002002 0.0 0.0 0.3
0.0349095655605197 0.0 0.03404255319148936 0.0 0.58
0.0599095655605197 0.08008008008008005 0.0 0.0 0.6999999999999997
0.0849095655605197 0.0800800800800801 0.03404255319148937 0.0 0.9400000000000003
0.1099095655605197 0.0 0.06808510638297874 0.0 0.9000000000000002
0.1349095655605197 0.08008008008008001 0.102127659574468 0.0 1.259999999999999
0.15990956556051972 0.14014014014014017 0.13617021276595748 0.0 1.0000000000000002
0.1849095655605197 0.020020020020020023 0.03404255319148937 0.0 1.2000000000000002
0.2099095655605197 0.04004004004004005 0.10212765957446811 0.0 1.1800000000000002
0.2349095655605197 0.04004004004004005 0.13617021276595748 0.0 1.1200000000000003
0.2599095655605197 0.14014014014014003 0.13617021276595734 0.0 1.1599999999999988
0.2849095655605197 0.20020020020020002 0.034042553191489335 0.22988505747126417 1.4799999999999986
0.30990956556051974 0.18018018018018042 0.17021276595744703 0.0 1.2600000000000016
0.3349095655605197 0.12012012012012002 0.23829787234042535 0.22988505747126417 1.0599999999999992
0.35990956556051973 0.2802802802802807 0.17021276595744703 0.0 1.500000000000002
0.3849095655605197 0.28028028028028007 0.204255319148936 0.22988505747126417 1.4399999999999986
0.4099095655605197 0.44044044044044006 0.23829787234042535 0.22988505747126417 1.259999999999999
0.43490956556051974 0.5205205205205212 0.6127659574468093 0.2298850574712647 1.1800000000000015
0.4599095655605197 0.6006006006006 0.37446808510638263 0.22988505747126417 1.279999999999999
0.48490956556051973 0.6206206206206214 0.6468085106382987 0.9195402298850588 1.1400000000000017
0.5099095655605197 0.6806806806806801 0.8510638297872333 0.6896551724137925 1.3799999999999988
0.5349095655605197 0.8408408408408401 0.714893617021276 0.6896551724137925 1.3799999999999988
0.5599095655605197 0.9809809809809801 1.3957446808510627 0.45977011494252834 1.4399999999999986
0.5849095655605198 1.0210210210210202 1.3617021276595733 0.45977011494252834 1.1799999999999988
0.6099095655605198 1.1211211211211252 1.5319148936170266 0.4597701149425304 1.380000000000005
0.6349095655605197 1.46146146146146 2.4170212765957424 0.0 1.5599999999999987
0.6599095655605197 1.7417417417417402 2.7914893617021255 1.1494252873563209 1.1599999999999988
0.6849095655605197 1.9419419419419401 3.3702127659574437 1.1494252873563209 1.1999999999999988
0.7099095655605198 2.2422422422422406 3.8127659574468056 0.0 1.0199999999999991
0.7349095655605198 2.402402402402411 2.961702127659585 1.149425287356326 1.1400000000000041
0.7599095655605197 2.4824824824824803 2.9617021276595716 1.6091954022988493 1.0399999999999991
0.7849095655605197 2.38238238238238 2.485106382978721 1.379310344827585 0.9399999999999992
0.8099095655605197 2.6826826826826804 2.1787234042553174 1.1494252873563209 0.7399999999999993
0.8349095655605198 2.7827827827827805 2.076595744680849 2.528735632183906 0.9399999999999992
0.8599095655605198 3.2632632632632745 1.5659574468085162 2.298850574712652 0.46000000000000163
0.8849095655605197 2.86286286286286 1.4978723404255305 2.528735632183906 0.4599999999999996
0.9099095655605197 2.7227227227227204 0.9872340425531905 5.747126436781604 0.35999999999999965
0.9349095655605197 1.9019019019019003 0.8851063829787227 7.586206896551718 0.2199999999999998
0.9599095655605198 0.6006006006006 0.6808510638297867 6.66666666666666 0.05999999999999995
1.0     0.0     0.0     0.0     0.0    
}{\histaddsumnorm}

\pgfplotstableread{
bins domain_y class_y tp_y fp_y
0.01758679933845997 0.0 0.0 0.0 0.04
0.04258679933845997 0.0 0.0 0.0 0.0
0.06758679933845997 0.0 0.0 0.0 0.0
0.09258679933845998 0.020000000000000004 0.0 0.0 0.06000000000000001
0.11758679933845997 0.020000000000000004 0.0 0.0 0.06000000000000001
0.14258679933845997 0.019999999999999983 0.0 0.0 0.19999999999999982
0.16758679933846 0.0 0.0 0.0 0.2800000000000001
0.19258679933845999 0.020000000000000004 0.0 0.2209944751381216 0.6400000000000001
0.21758679933845998 0.0 0.0 0.0 0.5600000000000002
0.24258679933845997 0.04000000000000001 0.0 0.0 1.2000000000000002
0.26758679933845997 0.07999999999999993 0.0 0.0 2.0199999999999982
0.29258679933846 0.17999999999999983 0.0 0.0 2.219999999999998
0.31758679933846 0.34000000000000047 0.0 0.0 2.940000000000004
0.34258679933846 0.6599999999999995 0.033840947546531275 0.0 3.279999999999997
0.36758679933846 1.3600000000000019 0.03384094754653135 0.0 3.720000000000005
0.39258679933845997 2.219999999999998 0.033840947546531275 0.0 3.9999999999999964
0.41758679933846 2.7599999999999976 0.06768189509306255 0.0 3.9199999999999964
0.44258679933846 3.780000000000005 0.10152284263959405 0.0 3.4600000000000044
0.46758679933846 5.079999999999996 0.0 0.4419889502762427 2.519999999999998
0.49258679933846 4.540000000000006 0.2030456852791881 0.0 2.440000000000003
0.51758679933846 4.9199999999999955 0.3045685279187814 0.22099447513812134 1.9399999999999982
0.54258679933846 4.099999999999996 0.3045685279187814 0.0 1.6399999999999986
0.56758679933846 3.519999999999997 0.9813874788494068 0.0 0.7999999999999993
0.59258679933846 2.1799999999999984 1.3197969543147197 1.1049723756906067 0.7799999999999994
0.6175867993384601 1.660000000000006 1.7597292724196338 0.8839779005524894 0.4800000000000017
0.64258679933846 1.0799999999999992 3.3164128595600646 1.1049723756906067 0.2999999999999997
0.66758679933846 0.6399999999999995 4.737732656514378 2.430939226519335 0.1199999999999999
0.69258679933846 0.31999999999999973 6.023688663282567 2.6519337016574562 0.13999999999999987
0.71758679933846 0.19999999999999982 5.109983079526222 3.977900552486184 0.17999999999999983
0.7425867993384601 0.04000000000000014 5.313028764805434 4.419889502762446 0.04000000000000014
0.76758679933846 0.039999999999999966 3.7901861252115028 4.198895027624306 0.0
0.79258679933846 0.05999999999999995 2.0642978003384074 4.640883977900548 0.019999999999999983
0.81758679933846 0.05999999999999995 1.8274111675126887 5.3038674033149125 0.0
0.84258679933846 0.039999999999999966 1.0152284263959381 3.5359116022099415 0.0
0.8675867993384601 0.0 0.7106598984771598 2.209944751381223 0.0
0.89258679933846 0.0 0.47377326565143785 1.3259668508287281 0.0
0.91758679933846 0.019999999999999983 0.3045685279187814 0.4419889502762427 0.0
0.94258679933846 0.0 0.06768189509306255 0.6629834254143641 0.0
0.96758679933846 0.0 0.10152284263959382 0.22099447513812134 0.0
1.0     0.0     0.0     0.0     0.0    
}{\histearlyfusion}

\pgfplotstableread{
bins domain_y class_y tp_y fp_y
0.07650424540042877 0.0 0.0 0.0 0.06000000000000001
0.10150424540042877 0.0 0.0 0.0 0.04000000000000001
0.12650424540042876 0.0 0.0 0.0 0.12000000000000002
0.15150424540042876 0.0 0.0 0.0 0.3400000000000001
0.17650424540042875 0.020000000000000004 0.0 0.0 0.5000000000000001
0.20150424540042874 0.04000000000000001 0.0 0.0 0.5000000000000001
0.22650424540042874 0.0 0.0 0.0 1.0200000000000014
0.2515042454004287 0.09999999999999991 0.0 0.0 1.6199999999999986
0.2765042454004287 0.039999999999999966 0.0 0.0 2.1599999999999984
0.30150424540042875 0.10000000000000013 0.03384094754653135 0.0 2.500000000000003
0.3265042454004287 0.16000000000000023 0.03384094754653135 0.0 3.140000000000004
0.3515042454004287 0.19999999999999982 0.16920473773265637 0.0 3.7399999999999967
0.3765042454004287 0.39999999999999963 0.06768189509306255 0.44444444444444403 3.8999999999999964
0.4015042454004287 0.3800000000000005 0.23688663282571945 0.22222222222222254 2.980000000000004
0.4265042454004287 0.700000000000001 0.6429780033840956 0.0 2.8600000000000034
0.45150424540042866 1.319999999999999 0.4399323181049066 0.0 2.639999999999998
0.4765042454004287 1.8599999999999983 0.9475465313028757 0.22222222222222202 2.1199999999999983
0.5015042454004287 2.4199999999999977 0.9813874788494068 0.0 1.8199999999999985
0.5265042454004287 3.6400000000000126 1.827411167512697 0.22222222222222301 1.8800000000000066
0.5515042454004286 5.019999999999995 2.0304568527918763 0.22222222222222202 1.339999999999999
0.5765042454004287 5.159999999999996 3.4179357021996584 0.22222222222222202 1.4199999999999986
0.6015042454004287 5.880000000000021 4.839255499153993 0.6666666666666691 1.0600000000000038
0.6265042454004286 4.799999999999995 4.940778341793566 1.3333333333333321 0.9399999999999992
0.6515042454004286 3.199999999999997 5.279187817258879 1.7777777777777761 0.5199999999999996
0.6765042454004286 2.1599999999999984 4.365482233502534 2.2222222222222205 0.2399999999999998
0.7015042454004287 1.1799999999999988 2.9103214890016895 1.7777777777777761 0.31999999999999973
0.7265042454004287 0.6200000000000021 2.2673434856176056 2.6666666666666763 0.1400000000000005
0.7515042454004286 0.15999999999999986 1.5905245346869699 6.222222222222217 0.05999999999999995
0.7765042454004286 0.3399999999999997 0.8798646362098131 6.222222222222217 0.0
0.8015042454004286 0.04000000000000014 0.7783417935702227 5.333333333333353 0.02000000000000007
0.8265042454004285 0.019999999999999983 0.5414551607445004 3.7777777777777746 0.0
0.8515042454004286 0.0 0.4060913705583753 2.4444444444444424 0.0
0.8765042454004286 0.019999999999999983 0.33840947546531275 1.5555555555555542 0.0
0.9015042454004286 0.0 0.033840947546531275 1.7777777777777761 0.0
0.9265042454004286 0.0 0.0 0.44444444444444603 0.0
0.9515042454004285 0.019999999999999983 0.0 0.22222222222222202 0.0
1.0     0.0     0.0     0.0     0.0    
}{\histlatefusion}

\pgfplotstableread{
bins domain_y class_y tp_y fp_y
0.12945888936519623 0.0 0.0 0.0 0.08000000000000002
0.15445888936519622 0.0 0.0 0.0 0.14000000000000004
0.17945888936519622 0.0 0.0 0.0 0.46000000000000013
0.2044588893651962 0.0 0.0 0.0 0.9600000000000002
0.2294588893651962 0.05999999999999995 0.0 0.0 1.339999999999999
0.25445888936519623 0.06000000000000008 0.0 0.0 1.9600000000000024
0.2794588893651962 0.12000000000000016 0.0 0.0 3.4000000000000044
0.30445888936519616 0.17999999999999983 0.033840947546531275 0.0 3.579999999999997
0.3294588893651962 0.2599999999999998 0.06768189509306255 0.0 3.9599999999999964
0.3544588893651962 0.4000000000000005 0.0 0.0 4.120000000000005
0.3794588893651962 0.740000000000001 0.10152284263959405 0.0 3.8800000000000052
0.40445888936519614 1.4199999999999986 0.033840947546531275 0.0 3.619999999999997
0.42945888936519616 1.5599999999999987 0.06768189509306255 0.22598870056497156 2.839999999999997
0.4544588893651962 2.820000000000004 0.16920473773265673 0.0 2.1000000000000028
0.47945888936519615 3.780000000000005 0.23688663282571945 0.22598870056497208 1.8000000000000025
0.5044588893651961 5.559999999999995 0.1353637901861251 0.22598870056497156 1.6199999999999986
0.5294588893651961 5.739999999999995 0.9475465313028757 0.4519774011299431 1.0999999999999992
0.5544588893651962 6.039999999999995 0.744500846023688 0.4519774011299431 1.139999999999999
0.5794588893651962 4.6000000000000165 1.4551607445008512 0.4519774011299451 0.7200000000000026
0.6044588893651961 3.1599999999999975 1.5228426395939072 0.22598870056497156 0.5799999999999994
0.6294588893651961 1.8599999999999983 2.4365482233502513 1.129943502824858 0.2999999999999997
0.6544588893651961 0.9800000000000035 3.925549915397645 0.4519774011299451 0.2000000000000007
0.679458889365196 0.5199999999999996 4.297800338409472 1.581920903954801 0.05999999999999995
0.7044588893651961 0.07999999999999993 6.700507614213191 1.3559322033898293 0.039999999999999966
0.7294588893651961 0.0 5.245346869712347 1.581920903954801 0.0
0.7544588893651961 0.019999999999999983 4.8054145516074405 1.8079096045197725 0.0
0.7794588893651961 0.02000000000000007 1.827411167512697 2.033898305084753 0.0
0.804458889365196 0.019999999999999983 1.7935702199661576 1.3559322033898293 0.0
0.8294588893651961 0.0 0.9137055837563444 3.615819209039545 0.0
0.8544588893651961 0.0 1.0490693739424741 4.971751412429397 0.0
0.879458889365196 0.0 0.5414551607445004 6.553672316384175 0.0
0.904458889365196 0.0 0.6768189509306255 6.553672316384175 0.0
0.929458889365196 0.0 0.23688663282571892 3.163841807909602 0.0
0.9544588893651961 0.0 0.033840947546531275 1.581920903954801 0.0
1.0     0.0     0.0     0.0     0.0    
}{\histlatefusionclean}

\pgfplotstableread{
bins domain_y class_y tp_y fp_y
0.11408969014883041 0.0 0.0 0.0 0.04000000000000001
0.1390896901488304 0.0 0.0 0.0 0.18000000000000002
0.1640896901488304 0.0 0.0 0.0 0.38000000000000006
0.1890896901488304 0.0 0.0 0.0 0.36000000000000004
0.2140896901488304 0.020000000000000004 0.0 0.0 0.9400000000000003
0.23908969014883039 0.08000000000000002 0.0 0.0 1.8800000000000006
0.2640896901488304 0.06000000000000008 0.0 0.0 2.520000000000003
0.28908969014883035 0.09999999999999991 0.0 0.0 3.3399999999999967
0.31408969014883037 0.2599999999999998 0.06768189509306255 0.0 3.7999999999999967
0.3390896901488304 0.36000000000000043 0.03384094754653135 0.0 3.860000000000005
0.36408969014883036 0.5200000000000007 0.0 0.0 4.580000000000006
0.3890896901488303 0.8399999999999992 0.10152284263959382 0.0 3.7599999999999967
0.41408969014883035 1.339999999999999 0.06768189509306255 0.2247191011235953 3.659999999999997
0.43908969014883037 2.520000000000003 0.0676818950930627 0.0 2.2600000000000033
0.46408969014883034 3.900000000000005 0.1353637901861254 0.0 2.0000000000000027
0.4890896901488303 4.479999999999996 0.2707275803722502 0.2247191011235953 1.8999999999999984
0.5140896901488303 6.479999999999994 0.20304568527918765 0.4494382022471906 1.339999999999999
0.5390896901488303 5.739999999999995 0.744500846023688 0.2247191011235953 1.0999999999999992
0.5640896901488304 5.140000000000018 0.7445008460236914 0.6741573033707889 0.9200000000000033
0.5890896901488303 3.639999999999997 1.4551607445008448 0.2247191011235953 0.4799999999999996
0.6140896901488303 2.3799999999999977 1.6582064297800323 0.8988764044943812 0.37999999999999967
0.6390896901488303 1.1800000000000042 2.233502538071074 0.6741573033707889 0.08000000000000028
0.6640896901488302 0.5799999999999994 3.8578680203045654 1.1235955056179765 0.17999999999999983
0.6890896901488303 0.2599999999999998 4.500846023688659 0.8988764044943812 0.039999999999999966
0.7140896901488303 0.07999999999999993 6.565143824027067 1.3483146067415719 0.019999999999999983
0.7390896901488303 0.039999999999999966 5.4483925549915355 1.3483146067415719 0.0
0.7640896901488303 0.0 4.7377326565144 2.0224719101123667 0.0
0.7890896901488302 0.0 1.9627749576988136 2.022471910112358 0.0
0.8140896901488303 0.0 1.8274111675126887 2.022471910112358 0.0
0.8390896901488303 0.0 0.9137055837563485 2.921348314606752 0.0
0.8640896901488302 0.0 0.9813874788494068 5.617977528089883 0.0
0.8890896901488302 0.0 0.7106598984771567 6.7415730337078585 0.0
0.9140896901488302 0.0 0.5076142131979691 6.292134831460669 0.0
0.9390896901488303 0.0 0.20304568527918765 2.921348314606739 0.0
0.9640896901488303 0.0 0.0 1.1235955056179816 0.0
1.0     0.0     0.0     0.0     0.0    
}{\histlatefusioncleanweight}

%% file: tab/nvsk.tex
\resizebox{2\columnwidth}{!}
{
\tiny
\begin{tabular}{cccccc@{\msp}ccccc@{\msp}ccccc@{\msp}ccccc@{\msp}ccccc}
\toprule
 &
  \multicolumn{5}{c}{\Th{Avg}} &
  \multicolumn{5}{c}{\Th{ImageNet-R}} &
  \multicolumn{5}{c}{\Th{MiniDN}} &
  \multicolumn{5}{c}{\Th{NICO++}} &
  \multicolumn{5}{c}{\Th{LTLL}} \\ \cmidrule{2-26}
\diagbox[width=4em,height=2em]{\hspace{-5pt}$k$}{$n$}  & 
  1 &
  7 &
  15 &
  30 &
  45 &
  1 &
  7 &
  15 &
  30 &
  45 &
  1 &
  7 &
  15 &
  30 &
  45 &
  1 &
  7 &
  15 &
  30 &
  45 &
  1 &
  7 &
  15 &
  30 &
  45 \\ \midrule
1  & \cellcolor[HTML]{EA9999}25.1 & \cellcolor[HTML]{F8DFDF}28.0 & \cellcolor[HTML]{F8DFDF}28.0 & \cellcolor[HTML]{F8DFDF}28.0 & \cellcolor[HTML]{F8DFDF}28.0 & \cellcolor[HTML]{F5CECE}26.2 & \cellcolor[HTML]{F2C4C4}25.8 & \cellcolor[HTML]{F2C4C4}25.8 & \cellcolor[HTML]{F2C4C4}25.8 & \cellcolor[HTML]{F2C4C4}25.8 & \cellcolor[HTML]{EA9999}30.2 & \cellcolor[HTML]{F0B9B9}32.1 & \cellcolor[HTML]{F0B9B9}32.1 & \cellcolor[HTML]{F0B9B9}32.1 & \cellcolor[HTML]{F0B9B9}32.1 & \cellcolor[HTML]{EA9999}19.4 & \cellcolor[HTML]{F6D5D5}23.2 & \cellcolor[HTML]{F6D5D5}23.2 & \cellcolor[HTML]{F6D5D5}23.2 & \cellcolor[HTML]{F6D5D5}23.2 & \cellcolor[HTML]{F6D5D5}24.5 & \cellcolor[HTML]{D9EAD1}30.8 & \cellcolor[HTML]{D9EAD1}30.8 & \cellcolor[HTML]{D9EAD1}30.8 & \cellcolor[HTML]{D9EAD1}30.8 \\
10 & \cellcolor[HTML]{F4F9F2}29.6 & \cellcolor[HTML]{99C884}31.5 & \cellcolor[HTML]{A7CF95}31.2 & \cellcolor[HTML]{B7D8A9}30.9 & \cellcolor[HTML]{D9EBD1}30.1 & \cellcolor[HTML]{C9E2BE}29.1 & \cellcolor[HTML]{93C47D}30.1 & \cellcolor[HTML]{A9D098}29.7 & \cellcolor[HTML]{FFFFFF}28.1 & \cellcolor[HTML]{F8E0E0}26.9 & \cellcolor[HTML]{FCF1F1}35.4 & \cellcolor[HTML]{DEEDD7}36.7 & \cellcolor[HTML]{FFFFFF}36.2 & \cellcolor[HTML]{FCF1F1}35.4 & \cellcolor[HTML]{F9E5E5}34.7 & \cellcolor[HTML]{F7DBDB}23.6 & \cellcolor[HTML]{FEFCFC}25.7 & \cellcolor[HTML]{FEFEFE}25.8 & \cellcolor[HTML]{FEFBFB}25.6 & \cellcolor[HTML]{FDF6F6}25.3 & \cellcolor[HTML]{E7F2E2}30.1 & \cellcolor[HTML]{A3CD91}33.5 & \cellcolor[HTML]{ABD19A}33.1 & \cellcolor[HTML]{93C47D}34.3 & \cellcolor[HTML]{A1CC8E}33.6 \\
20 & \cellcolor[HTML]{F4F9F2}29.6 & \cellcolor[HTML]{93C47D}31.6 & \cellcolor[HTML]{9FCB8B}31.4 & \cellcolor[HTML]{CBE3C1}30.4 & \cellcolor[HTML]{FEFEFE}29.3 & \cellcolor[HTML]{C4DFB8}29.2 & \cellcolor[HTML]{9ECA8B}29.9 & \cellcolor[HTML]{BFDCB2}29.3 & \cellcolor[HTML]{FBEDED}27.4 & \cellcolor[HTML]{F3C6C6}25.9 & \cellcolor[HTML]{FFFFFF}36.2 & \cellcolor[HTML]{B5D7A6}37.3 & \cellcolor[HTML]{D7E9CF}36.8 & \cellcolor[HTML]{FDF8F8}35.8 & \cellcolor[HTML]{FBEEEE}35.2 & \cellcolor[HTML]{F9E4E4}24.2 & \cellcolor[HTML]{DCECD4}26.1 & \cellcolor[HTML]{CDE4C3}26.2 & \cellcolor[HTML]{DCECD4}26.1 & \cellcolor[HTML]{FEFCFC}25.7 & \cellcolor[HTML]{FEFCFC}28.6 & \cellcolor[HTML]{A9D098}33.2 & \cellcolor[HTML]{A9D098}33.2 & \cellcolor[HTML]{B9D9AB}32.4 & \cellcolor[HTML]{E3F0DD}30.3 \\
30 & \cellcolor[HTML]{F6FAF4}29.5 & \cellcolor[HTML]{BEDCB1}30.7 & \cellcolor[HTML]{C1DDB4}30.7 & \cellcolor[HTML]{F9FCF7}29.5 & \cellcolor[HTML]{F9E2E2}28.1 & \cellcolor[HTML]{CFE5C5}29.0 & \cellcolor[HTML]{A4CD91}29.8 & \cellcolor[HTML]{C9E2BE}29.1 & \cellcolor[HTML]{F9E5E5}27.1 & \cellcolor[HTML]{F1BCBC}25.5 & \cellcolor[HTML]{E4F1DF}36.6 & \cellcolor[HTML]{A1CC8E}37.6 & \cellcolor[HTML]{C3DEB6}37.1 & \cellcolor[HTML]{FFFFFF}36.2 & \cellcolor[HTML]{FCF1F1}35.4 & \cellcolor[HTML]{FAE9E9}24.5 & \cellcolor[HTML]{BFDCB2}26.3 & \cellcolor[HTML]{B0D4A0}26.4 & \cellcolor[HTML]{BFDCB2}26.3 & \cellcolor[HTML]{F8FCF7}25.9 & \cellcolor[HTML]{FDF5F5}27.9 & \cellcolor[HTML]{FBFDFA}29.1 & \cellcolor[HTML]{E9F3E4}30.0 & \cellcolor[HTML]{FDF8F8}28.2 & \cellcolor[HTML]{F8E1E1}25.7 \\
40 & \cellcolor[HTML]{FCF4F4}28.9 & \cellcolor[HTML]{FAFCF9}29.4 & \cellcolor[HTML]{FEFFFE}29.3 & \cellcolor[HTML]{F7DBDB}27.9 & \cellcolor[HTML]{F2C1C1}26.8 & \cellcolor[HTML]{E4F1DF}28.6 & \cellcolor[HTML]{B4D6A5}29.5 & \cellcolor[HTML]{D4E8CC}28.9 & \cellcolor[HTML]{F7DBDB}26.7 & \cellcolor[HTML]{EDAAAA}24.8 & \cellcolor[HTML]{DEEDD7}36.7 & \cellcolor[HTML]{9AC886}37.7 & \cellcolor[HTML]{BCDBAE}37.2 & \cellcolor[HTML]{FFFFFF}36.2 & \cellcolor[HTML]{FBEFEF}35.3 & \cellcolor[HTML]{FBECEC}24.7 & \cellcolor[HTML]{B0D4A0}26.4 & \cellcolor[HTML]{A2CC8F}26.5 & \cellcolor[HTML]{B0D4A0}26.4 & \cellcolor[HTML]{DCECD4}26.1 & \cellcolor[HTML]{F8DEDE}25.4 & \cellcolor[HTML]{F5D1D1}24.1 & \cellcolor[HTML]{F6D7D7}24.7 & \cellcolor[HTML]{F1BEBE}22.1 & \cellcolor[HTML]{EFB2B2}20.8 \\
50 & \cellcolor[HTML]{F7DCDC}27.9 & \cellcolor[HTML]{FBEFEF}28.7 & \cellcolor[HTML]{FBEDED}28.6 & \cellcolor[HTML]{F2C4C4}26.9 & \cellcolor[HTML]{EEADAD}25.9 & \cellcolor[HTML]{FFFFFF}28.1 & \cellcolor[HTML]{C4DFB8}29.2 & \cellcolor[HTML]{E4F1DF}28.6 & \cellcolor[HTML]{F5CECE}26.2 & \cellcolor[HTML]{EA9999}24.1 & \cellcolor[HTML]{D7E9CF}36.8 & \cellcolor[HTML]{93C47D}37.8 & \cellcolor[HTML]{AED39E}37.4 & \cellcolor[HTML]{F9FCF7}36.3 & \cellcolor[HTML]{FCF1F1}35.4 & \cellcolor[HTML]{FBEEEE}24.8 & \cellcolor[HTML]{A2CC8F}26.5 & \cellcolor[HTML]{93C47D}26.6 & \cellcolor[HTML]{A2CC8F}26.5 & \cellcolor[HTML]{DCECD4}26.1 & \cellcolor[HTML]{F1BCBC}21.8 & \cellcolor[HTML]{F0B6B6}21.2 & \cellcolor[HTML]{F1BBBB}21.7 & \cellcolor[HTML]{EA9C9C}18.5 & \cellcolor[HTML]{EA9999}18.1

\\ \bottomrule
\end{tabular}
}

%% file: tab/ablations_w_searle.tex
\resizebox{2\columnwidth}{!}
{
\tiny
\begin{tabular}{lccccccccccccccccccccccccc}
\toprule
 & 
  \multicolumn{5}{c}{\Th{Avg}} &
  \multicolumn{5}{c}{\Th{ImageNet-R}} &
  \multicolumn{5}{c}{\Th{MiniDn}} &
  \multicolumn{5}{c}{\Th{NICO++}} &
  \multicolumn{5}{c}{\Th{LTLL}} \\ \cmidrule{2-26}
$m$ &
  1 &
  3 &
  7 &
  10 &
  15 &
  1 &
  3 &
  7 &
  10 &
  15 &
  1 &
  3 &
  7 &
  10 &
  15 &
  1 &
  3 &
  7 &
  10 &
  15 &
  1 &
  3 &
  7 &
  10 &
  15 \\ \midrule
SRL & \cellcolor[HTML]{EEACAC}19.5 & \cellcolor[HTML]{EDAAAA}19.3 & \cellcolor[HTML]{ECA3A3}18.7 & \cellcolor[HTML]{EB9E9E}18.2 & \cellcolor[HTML]{EA9999}17.7 & \cellcolor[HTML]{EB9E9E}9.3  & \cellcolor[HTML]{EA9B9B}8.9  & \cellcolor[HTML]{EA9999}8.5  & \cellcolor[HTML]{EA9999}8.4  & \cellcolor[HTML]{ECA3A3}10.2 & \cellcolor[HTML]{F0B9B9}24.3 & \cellcolor[HTML]{F0B8B8}24.2 & \cellcolor[HTML]{EDAAAA}22.7 & \cellcolor[HTML]{ECA3A3}21.9 & \cellcolor[HTML]{EA9999}20.8 & \cellcolor[HTML]{F3C5C5}15.9 & \cellcolor[HTML]{F3C5C5}15.9 & \cellcolor[HTML]{F3C6C6}16.0 & \cellcolor[HTML]{F3C6C6}16.0 & \cellcolor[HTML]{EEB1B1}13.7 & \cellcolor[HTML]{FCF2F2}28.4 & \cellcolor[HTML]{FBEFEF}28.2 & \cellcolor[HTML]{F9E5E5}27.5 & \cellcolor[HTML]{F6D7D7}26.5 & \cellcolor[HTML]{F6D3D3}26.2 \\
E   & \cellcolor[HTML]{FAEAEA}25.1 & \cellcolor[HTML]{FBEEEE}25.5 & \cellcolor[HTML]{F2C0C0}21.3 & \cellcolor[HTML]{EDABAB}19.4 & \multicolumn{1}{l}{-}        & \cellcolor[HTML]{FAFDF9}26.2 & \cellcolor[HTML]{FCF4F4}24.2 & \cellcolor[HTML]{F4CDCD}17.5 & \cellcolor[HTML]{F1BFBF}15.1 & \multicolumn{1}{l}{-}        & \cellcolor[HTML]{FCF0F0}30.2 & \cellcolor[HTML]{FEFCFC}31.5 & \cellcolor[HTML]{F4CBCB}26.2 & \cellcolor[HTML]{EEADAD}23.0 & \multicolumn{1}{l}{-}        & \cellcolor[HTML]{F9E6E6}19.4 & \cellcolor[HTML]{F3C9C9}16.3 & \cellcolor[HTML]{ECA3A3}12.2 & \cellcolor[HTML]{EA9999}11.1 & \multicolumn{1}{l}{-}        & \cellcolor[HTML]{F1BCBC}24.5 & \cellcolor[HTML]{F4F9F2}29.8 & \cellcolor[HTML]{FFFFFF}29.3 & \cellcolor[HTML]{FCF1F1}28.3 & \multicolumn{1}{l}{-}        \\
E+  & \cellcolor[HTML]{FEFEFE}26.9 & \cellcolor[HTML]{E6F1E1}28.0 & \cellcolor[HTML]{F6D5D5}23.2 & \cellcolor[HTML]{F0B8B8}20.6 & \multicolumn{1}{l}{-}        & \cellcolor[HTML]{C0DDB3}28.5 & \cellcolor[HTML]{E0EED9}27.2 & \cellcolor[HTML]{F6D7D7}19.2 & \cellcolor[HTML]{F3C7C7}16.4 & \multicolumn{1}{l}{-}        & 34.9                         & \cellcolor[HTML]{BDDBAF}35.4 & \cellcolor[HTML]{F9E1E1}28.6 & \cellcolor[HTML]{F1BEBE}24.8 & \multicolumn{1}{l}{-}        & 22.3                         & \cellcolor[HTML]{F7DADA}18.1 & \cellcolor[HTML]{EEADAD}13.3 & \cellcolor[HTML]{EBA1A1}12.0 & \multicolumn{1}{l}{-}        & 22.0                         & \cellcolor[HTML]{D3E7CA}31.3 & \cellcolor[HTML]{CCE3C1}31.6 & \cellcolor[HTML]{FEFEFE}29.2 & \multicolumn{1}{l}{-}        \\
L   & \cellcolor[HTML]{FAEAEA}25.1 & \cellcolor[HTML]{E5F1E0}28.1 & \cellcolor[HTML]{E7F2E3}28.0 & \cellcolor[HTML]{F4F9F2}27.4 & \cellcolor[HTML]{FDF8F8}26.3 & \cellcolor[HTML]{FAFDF9}26.2 & \cellcolor[HTML]{DEEDD7}27.3 & \cellcolor[HTML]{FEFDFD}25.8 & \cellcolor[HTML]{FDF7F7}24.7 & \cellcolor[HTML]{FBEDED}23.0 & \cellcolor[HTML]{FBF0F0}30.2 & \cellcolor[HTML]{E9F3E5}33.0 & \cellcolor[HTML]{FAFCF9}32.1 & \cellcolor[HTML]{FDF6F6}30.9 & \cellcolor[HTML]{F9E4E4}28.9 & \cellcolor[HTML]{F9E6E6}19.4 & \cellcolor[HTML]{EFF6EB}22.7 & \cellcolor[HTML]{E1EFDB}23.2 & \cellcolor[HTML]{EBF5E7}22.8 & \cellcolor[HTML]{FEFDFD}21.9 & \cellcolor[HTML]{F1BCBC}24.5 & \cellcolor[HTML]{FFFFFF}29.3 & \cellcolor[HTML]{DEEDD8}30.8 & \cellcolor[HTML]{D6E9CD}31.2 & \cellcolor[HTML]{CDE4C2}31.6 \\
L+  & \cellcolor[HTML]{FFFFFF}26.9 & \cellcolor[HTML]{A9D097}30.7 & \cellcolor[HTML]{9AC886}31.3 & \cellcolor[HTML]{ADD29C}30.5 & \cellcolor[HTML]{DDEDD6}28.4 & \cellcolor[HTML]{BFDCB2}28.5 & \cellcolor[HTML]{9ECA8A}29.8 & \cellcolor[HTML]{B0D4A0}29.1 & \cellcolor[HTML]{D1E6C8}27.8 & \cellcolor[HTML]{FEFAFA}25.3 & 34.9                         & \cellcolor[HTML]{95C580}37.6 & \cellcolor[HTML]{ACD29B}36.3 & \cellcolor[HTML]{C7E1BC}34.9 & \cellcolor[HTML]{FBFDFA}32.0 & 22.3                         & \cellcolor[HTML]{A3CD90}25.5 & \cellcolor[HTML]{9DCA89}25.7 & \cellcolor[HTML]{ABD29A}25.2 & \cellcolor[HTML]{CFE5C5}23.9 & 22.0                         & \cellcolor[HTML]{F2F8F0}29.9 & \cellcolor[HTML]{93C47D}34.2 & \cellcolor[HTML]{93C47D}34.2 & \cellcolor[HTML]{B9D9AB}32.5 \\
W   & \cellcolor[HTML]{FFFFFF}26.9 & \cellcolor[HTML]{AAD199}30.6 & \cellcolor[HTML]{93C47D}31.6 & \cellcolor[HTML]{95C57F}31.6 & \cellcolor[HTML]{A0CB8D}31.1 & \cellcolor[HTML]{BFDCB2}28.5 & \cellcolor[HTML]{93C47D}30.2 & \cellcolor[HTML]{9BC987}29.9 & \cellcolor[HTML]{AAD198}29.4 & \cellcolor[HTML]{C1DEB5}28.4 & 34.9                         & \cellcolor[HTML]{93C47D}37.7 & \cellcolor[HTML]{9BC987}37.3 & \cellcolor[HTML]{A3CD90}36.8 & \cellcolor[HTML]{AFD49F}36.2 & 22.3                         & \cellcolor[HTML]{A2CD8F}25.6 & \cellcolor[HTML]{93C47D}26.1 & \cellcolor[HTML]{95C57F}26.1 & \cellcolor[HTML]{99C784}25.9 & 22.0                         & \cellcolor[HTML]{FEFCFC}29.1 & \cellcolor[HTML]{A9D097}33.2 & \cellcolor[HTML]{98C783}34.0 & \cellcolor[HTML]{9CC988}33.8

 \\ \bottomrule
\end{tabular}
}

%% file: tab/ablations_add_domain.tex
\centering
\footnotesize
\setlength{\tabcolsep}{10pt}
\begin{tabular}{lcccc}
\toprule
\mr{2}{\Th{Dataset}} & \mc{2}{\Th{Object Gain}} & \mc{2}{\Th{Domain Loss}} \\ \cmidrule{2-5}
           & \Th{$E^{+}$} & \Th{$L^{+}$} & \Th{$E^{+}$}  & \Th{$L^{+}$}  \\ \midrule
\imagenetr & +0.72 & +1.16 & -4.30 & -1.64 \\
\nico       & +0.21 & +0.40 & -0.66 & -1.29 \\
MiniDN    & +0.72 & +0.59 & -7.43 & -2.79 \\
\ltll     & +1.56 & +4.00 & -2.86 & -1.58 \\ \midrule
Avg      & +0.80 & +1.54 & -3.81 & -1.83 \\ \bottomrule
\end{tabular}

%% file: tex/conclusions.tex
\section{Conclusions}
\label{sec:conclusions}

We introduced \ours, a training-free composed image retrieval method for domain conversion based on a pre-trained CLIP model. The key component is the textual inversion of the query image based on soft assignment to a sparse vocabulary of words. Our detailed ablations show the importance of every component of the method and its robustness to the choice of hyper-parameters.
We also introduced three new benchmarks with different domain types, providing a broad testbed for future research. Despite its zero supervision, data, or training requirements, \ours outperforms the state-of-the-art methods by a large margin.

%% file: tex/appendix.tex
\section{Appendix}
\subsection{Method details}
\ours is presented in Algorithm \ref{alg:overview} for further clarity. The multiple text inversion, lines 10 - 12, is efficiently executed by a single NN-search for a set of queries whose GPU implementations are readily available (\eg, FAISS).

\subsection{Competing methods}
We provide the implementation details of the literature methods used in the main paper. 

\paragraph{\picword~\cite{pic2word}} achieves textual inversion in the latent space of the text tokens through a three-layered MLP. In every experiment with \picword, we use the officially pre-trained mapping network released by the authors. For ImageNet-R and MiniDN, the composed query has the same format as in the original paper: ``a [\emph{target domain}] of *'', \eg ``a cartoon of *''. For NICO++, the composed query is ``a * in [\emph{target domain}]'', \eg ``a * in autumn''.  Finally, for the LTLL dataset, the composed query ``a [\emph{target domain}] photo of *'' is used, \eg ``a today photo of *''.

\paragraph{\searle~\cite{searle}} performs textual inversion by test-time optimization to represent query images in the latent space of the vector tokens. We opt for the optimization variant instead of their feed-forward network since it is shown to perform better. We use the publicly released official implementation for our experiments. We refer to the version with default optimization hyper-parameters as ``\searle (default)'' and to our improved hyper-parameters by ``\searle''. 
Each query image is associated with different concepts retrieved from a vocabulary, which is the same as the \txlabels of our method. 
We refer to the number of those concepts by $m$ in \autoref{tab:searelabla}. The final composed queries are adapted for each dataset in the same way as for \picword.
We perform a hyper-parameter search for learning rate in \{0.2, 0.02, 0.002, 0.0002\}, iterations in \{5, 10, 50, 200, 350, 500\}, and the number of textual labels $m$ in \{1, 3, 7, 10, 15\}. The best results across all datasets are for $lr=0.0002$, $iters=350$, and $m=1$.  

\paragraph{\compodiff~\cite{compodiff}} is built on top of a frozen CLIP. We follow the publicly released official implementation for our experiments. We use the officially pre-trained denoising Transformer released by the authors. We do not use any masks or any mixed text condition. The query text includes only the target domain word, i.e., ``[\emph{target domain}]''.

\begin{algorithm}
\caption{\ours.\label{alg:overview}}
\input{tex/algo}
\end{algorithm}

\paragraph{\weicom~\cite{pke+24}} is a composed image retrieval method specialized for remote sensing. It fits a normal distribution to the similarities between the text query $g(t)$ and all the database images $f(x)$ for $x \in X$, and similarly for the image query $f(y)$. It uses each distribution's corresponding cumulative distribution function to transform the similarities closer to the uniform distribution. It then combines the similarities by summation.

\paragraph{\magic~\cite{magiclens}} is a composed image retrieval method that fine-tunes a VLM model on triplets collected from the internet, assuming that images from the same website share implicit relationships describable by textual instructions. We use the CLIP-L variant from the official code and evaluate its performance with two settings: the default prompt, ``find this object in [\emph{target domain}]'', which performed poorly across datasets, and prompts tailored per dataset. The best prompts were: ``a [\emph{target domain}] of this'' (ImageNet-R), ``a [\emph{target domain}] of'' (MiniDN), ``in [\emph{target domain}]'' (NICO++), and ``a [\emph{target domain}] photo of'' (LTLL). Results are reported as ``\magic (original prompt)'' and ``\magic''.

\subsection{Advanced baselines}
\label{sec:s_base}
In addition to the simple baselines in the main paper, we present the following more ''advanced`` baselines and summarize their performance in \autoref{tab:advanced}.

\paragraph{InstructPix2Pix}.~\cite{instructpix2pix} In this baseline, InstructPix2Pix is used to generate an image from our visual and textual queries. Then, retrieval is done by image-to-image similarities. The performance of this baseline is low, indicating that the combination of the two modalities through the visual encoder is sub-optimal. We qualitatively observe that although several of the generated images are quite successful, many are completely unsuccessful. 

\paragraph{\ours w/ img-cap}. In this baseline, we assume access to a dataset of image-caption pairs; the first 40M images and captions of LAION 400M~\cite{laion400m} are used. This set forms a joint visual-textual memory. Proxy images are retrieved from this memory, and their captions are treated as the text labels of textual inversion. Then, they are combined with the query text, and late fusion follows with weights equal to the similarities between the query and the memory images. The hyperparameters are the same as our standard \ours. Interestingly, this baseline surpasses \ours on \ltll for the case of ''today`` as the source domain.

\paragraph{\ours w/ captioners.} In this baseline, two captioners are used, namely BLIP~\cite{blip} and BLIP2~\cite{blip2}. Each captioner captions every query image, and the results are used as the two text labels for the image. Subsequently, our standard processing pipeline is followed. The similarities of each caption are used with the query image as weights for late fusion. This baseline uses extra architectures, is $15$ times slower than the standard \ours, and is consistently worse.

\begin{table}[t]
    \centering
    \begin{minipage}{\columnwidth}
        \centering
        \resizebox{\columnwidth}{!}{%
        \input{tab_supp/advanced_imagenet_r}
        }
        \subcaption{\imagenetr}
        \label{tab:advanced_imgnet_sota}
    \end{minipage}
    \vfill
    \begin{minipage}{\columnwidth}
        \centering
        \resizebox{\columnwidth}{!}{%
        \input{tab_supp/advanced_nico}
        }
        \subcaption{\nico}
        \label{tab:advanced_nico_sota}
    \end{minipage}
    \vfill
    \begin{minipage}{\columnwidth}
        \centering
        \resizebox{\columnwidth}{!}{%
        \input{tab_supp/advanced_minidn}
        }
        \subcaption{\minidn}
        \label{tab:advanced_minidn_sota}
    \end{minipage}
    \vfill
    \begin{minipage}{\columnwidth}
        \centering
        \resizebox{\columnwidth}{!}{%
        \input{tab_supp/advanced_leuven}
        }
        \subcaption{\ltll}
        \label{tab:advanced_leuven_sota}
    \end{minipage}
    \vspace{-10pt}
    \caption{\emph{Evaluation of advanced baselines.}
    }
    \label{tab:advanced}
    \vspace{-15pt}
\end{table}

\subsection{Additional results}

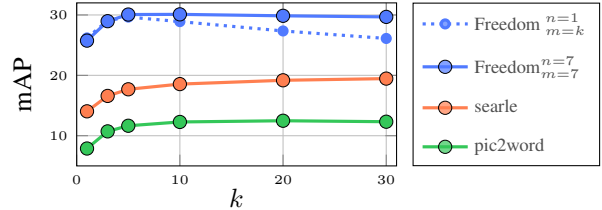
\begin{figure}[t]
\centering
\input{fig_supp/freedom_w_other_inversions}
\vspace{-10pt}
\caption{\emph{Different Inversions} with visual memory expansion and late-fusion on \imagenetr
\label{fig:different_inversions} 
\vspace{-5pt}}
\end{figure}

\paragraph{Impact of \ours components to different inversion methods.}
The three main components of \ours are text memory-based inversion, visual memory-based expansion, and late fusion. 
We apply the last two components on top of different inversion methods, whenever applicable, \ie, with SEARLE and Pic2Word.
Incorporating the two \ours components (using $m=k$, while $n$ is equal to $1$ inherently for both methods) improves both methods, while our text memory-based inversion performs consistently the best. 
This experiment is summarized in \autoref{fig:different_inversions}.
We follow the \ours workflow: A visual memory is used to enrich the query with $k$ images. Then, inversion follows as \ours, \searle, and \picword. Finally, the combination is done by late fusion. 
Although the memory-based inversion is more sensitive for large $k$ (dotted blue), our design choice of having a fixed number of final words ($m=7$) makes \ours robust. 

\paragraph{Impact of visual memory.}
We demonstrate the performance of \ours with different visual memories in \autoref{tab:laion}.
Compared to no visual memory, every other option improves the performance on average. Furthermore, every visual memory is advantageous for every individual dataset except for the case of ImageNet-R with LAION 40M due to the low availability of images in specific domains such as \emph{origami}.
Therefore, the efficacy of the memory remains robust even when dealing with unstructured datasets such as the image part of LAION. Additionally, including task-relevant images, even in small proportions, proves advantageous. The best improvements are achieved using the database as memory, which is our default choice. 

\begin{table}[t]
\centering
\input{tab_supp/extended_memory}
\vspace{-5pt}
\caption{\emph{Impact of the visual memory:} Comparing performance between no visual memory, the database as visual memory, a 40M-image LAION~\cite{laion400m} visual memory, and their union.}
\vspace{-5pt}
\label{tab:laion}
\end{table}

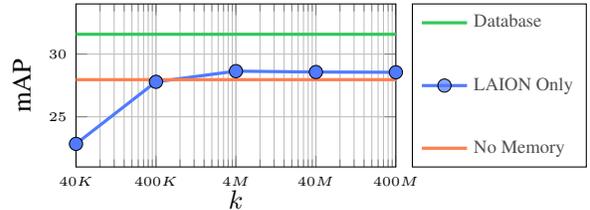
\begin{figure}[t]
\centering
\input{fig_supp/increasing_laion}
\vspace{-10pt}
\caption{\emph{Impact of the visual memory:} Performance comparison between no visual memory, the database as visual memory, and visual memory comprising LAION~\cite{laion400m} images of various sizes.}
\vspace{-10pt}
\label{fig:incr_laion} 
\end{figure}

We also study the effect of the size of the visual memory. We choose LAION subsets of size $40k$, $400k$, $4M$, $40M$, and $400M$ as visual memories. The average mAP of \imagenetr, \nico, \minidn, and \ltll is reported in \autoref{fig:incr_laion}. The database as a visual memory is the upper bound for this experiment, given that it is curated for the task. A performance saturation is observed for the visual memory of size $4M$, which surpasses the performance of the no-visual-memory baseline \ours (k=1). This supports the idea that visual memory is beneficial even when not curated. It also suggests a practical upper bound for the visual memory size. Notably, \ours, with a visual memory size of $4M$,  has a query latency of $24.8ms$ for \imagenetr.

\begin{figure}[t]
\centering
\vspace{-5pt}
\input{fig_supp/times}
\caption{\emph{Performance \vs query time.} Different variants of \ours are shown by varying hyper-parameter ($k$, $m$, $n$) values and textual memory size. \ours uses textual memory with size 20k (default) or 236k (reported).} 
\vspace{-5pt}
\label{fig:latency}
\end{figure}
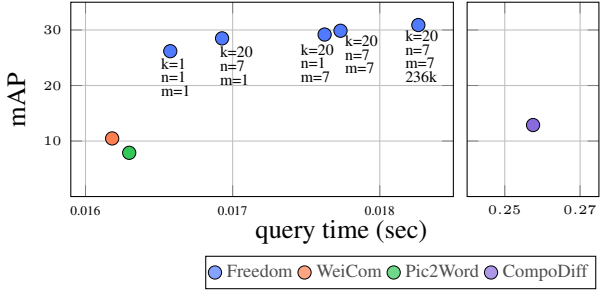

\begin{table}[t]
 \centering
 \input{tab_supp/sensitivity}
 \vspace{-5pt}
 \caption{\emph{Oracle experiment} to study the sensitivity of \ours by removing the $\ell$ words closest to each query's class label from the vocabulary. We report the mAP (\%) reduction.}
 \label{tab:removebest}
 \vspace{-12pt}
 \end{table}
 
\paragraph{Query time}. \autoref{fig:latency} presents the latency comparison between \ours and the competitive methods: \weicom, \picword, and \compodiff on \imagenetr. \compodiff achieves an mAP of 12.9 while being significantly slower, with a latency of $257.5ms$. In contrast, \weicom, \picword, and \ours (with m=1, n=1, and k=1) exhibit similar latencies (16.2ms, 16.3ms, and 16.6ms, respectively), but \ours outperforms the rest with an mAP of 26.18 compared to 10.47 for \weicom and 7.88 for \picword. Increasing \ours hyperparameters to k=20 and n=7 yields an additional 2.31 mAP with a minimal latency cost of 0.3ms. The variant with k=20, m=7, n=1 shows a further latency increase of 0.7ms, leading to a 0.69 mAP gain. The default \ours configuration (k=20, m=7, n=7) achieves 29.86 mAP with a total latency of 17.7ms. Expanding the text memory to 236k words \cite{nltk} results in an mAP of 30.89 with a final latency of 18.3ms.

To test the scalability of \ours, on top of the 236k text memory, we artificially enlarge the visual memory to 1M. The latency increases by only $1.8ms$ to a total of $20.1ms$. By expanding the database to 1M, we get a latency of $24.5ms$. Even in a database of two magnitudes larger (\imagenetr VS our artificial 1M database), \ours is more than $10$ times faster than \compodiff.
 
\paragraph{Oracle experiment.} We demonstrate the robustness of \ours with respect to the choice of textual memory. As an oracle experiment, the $\ell$ words closest to each query's class label are removed from our vocabulary, and the results with the remaining are reported. The performance reduction (mAP) for $\ell = 1, 5, 10, 20$ is summarized in \autoref{tab:removebest}. Despite the performance drop, \ours still outperforms the state-of-the-art on all datasets, even for $\ell = 20$. In \autoref{tab:removedwords}, we present examples of excluded words for this experiment.

 \begin{table}
  \centering
  \input{tab_supp/removed_words}
  \vspace{-5pt}
  \caption{\emph{Example words removed for the robustness ablation experiment}. The first column shows the class names of queries. The remaining columns (ranked in descending order) show top-ranked words from the textual memory.}
  \vspace{-10pt}
  \label{tab:removedwords}
  \end{table}

\paragraph{Memory-based inversion examples.}
In \autoref{tab:memory_inversion}, we present some examples of the memory-based inversion of \ours, including the inverted text and the corresponding frequency as weight.

 \begin{table*}
  \centering
  \input{tab_supp/memory_inversion}
  \vspace{-5pt}
  \caption{\emph{Memory-based inversion}. Examples of query images alongside their inverted text and corresponding weights of \ours.}
  \vspace{-5pt}
  \label{tab:memory_inversion}
  \end{table*}

\paragraph{SigLIP as the backbone.}
We test the transferability of \ours to other backbones by using features from SigLIP \cite{siglip}, and the results are summarized in \autoref{tab:siglip}. We observe a significant increase in all datasets in the $4.13$ to $16.36$ mAP range without additional tuning.

\begin{table*}[t]
\begin{center}
\begin{minipage}{.5\textwidth}
\centering
\begin{subtable}{\textwidth}
\caption{\imagenetr}
\label{tab:siglip_imgnet_sota}
\input{tab_supp/siglip_imagenet_r}
\end{subtable}
\end{minipage}%
\begin{minipage}{.5\textwidth}
\centering
\begin{subtable}{\textwidth}
\caption{\minidn}
\label{tab:siglip_minidn_sota}
\input{tab_supp/siglip_minidn}
\end{subtable}
\end{minipage}

\vspace{0pt} 

\begin{minipage}{.5\textwidth}
\centering
\begin{subtable}{\textwidth}
\caption{\nico}
\label{tab:siglip_nico_sota}
\input{tab_supp/siglip_nico}
\end{subtable}
\end{minipage}%
\begin{minipage}{.5\textwidth}
\centering
\begin{subtable}{\textwidth}
\caption{\ltll}
\label{tab:siglip_leuven_sota}
\input{tab_supp/siglip_leuven}
\end{subtable}
\end{minipage}
\vspace{-10pt}
\caption{\emph{Domain conversion mAP (\%)} on four datasets, with SigLIP as a backbone. The best is denoted in bold. \label{tab:siglip}
\vspace{-20pt}
}
\end{center}
\end{table*}

\paragraph{Datasets for general composed image retrieval.}
In this work, we focus on the domain conversion task, motivated by the significance of its applications.
Addressing the challenges of this task, particularly the utilization of bi-modal queries and open-world recognition across domains and objects, proves to be non-trivial. Given that our method handles these challenges well and considering that these challenges extend universally to the general composed image retrieval, we evaluate \ours on benchmarks of the general task: \fashioniq~\cite{fashioniq}, \cirr~\cite{cirr}, and \circo~\cite{searle}. The results are summarized in \autoref{tab:othercir}.

Even though \ours is training-free and its scope is domain conversion, the results indicate that it is comparable with some general methods. Specifically, compared to \picword, \searle, and \compodiff, \ours underperforms in Fashion-IQ, it performs comparably well in CIRR, and is the best approach in CIRCO.

\begin{table*}[t]
\begin{center}
\begin{minipage}{.49\textwidth}
\centering
\begin{subtable}{\textwidth}
\caption{\Th{CIRR}}
\label{tab:general_cirr}
\input{tab_supp/general_cirr}
\end{subtable}
\end{minipage}
\begin{minipage}{.49\textwidth}
\centering
\begin{subtable}{\textwidth}
\caption{\Th{CIRCO}}
\label{tab:general_circo}
\input{tab_supp/general_circo}

\end{subtable}
\end{minipage}

\vspace{0pt}

\begin{minipage}{1\textwidth}
\centering
\begin{subtable}{\textwidth}
\caption{\Th{Fashion-IQ}}
\label{tab:general_fashion}
\input{tab_supp/general_fashion}

\end{subtable}
\end{minipage}
\caption{\emph{Composed image retrieval beyond domain conversion:} We evaluate \ours on the three most popular benchmarks for general composed image retrieval. We denote with $^*$ the optimized parameters of \ours obtained through hyperparameter tuning on the validation set of \Th{CIRR}.
}
\vspace{-10pt}
\label{tab:othercir}
\end{center}
\end{table*}

\begin{table*}
\centering
\input{tab_supp/recall_tables}
\caption{\emph{Domain conversion evaluated by Recall@$k$ (\%)} on \imagenetr. Comparison of \ours with baselines and competitors. Source domain: \Th{Photo}; target domains: \Th{Cartoon}, \Th{Origami}, \Th{Toy}, and \Th{Sculpture}. \Th{AVG}: average Recall@$10$ and Recall@$50$ over all target domains. \textbf{Bold:} best, \ts{magenta}: second best.}
\label{tab:full_recall}
\end{table*}
  
\paragraph{Detailed results}. Following the literature~\cite{pic2word, compodiff, searle}, we evaluate on \imagenetr, using only the \Th{Photo} domain as source, and measure Recall@$k$. We compare with baselines and competitors in~\autoref{tab:full_recall}. The baselines and the \searle experiments are performed by us, \picword performance is reported from the original paper, the rest of the  \picword experiments, ARTEMIS~\cite{artemis}, CLIP4CIR~\cite{combiner}, and \compodiff are reported from the \compodiff paper. \ours outperforms all baselines and competitors by a large margin. \cirevl is the second best, even though it uses architectures with an estimated number of parameters of two orders of magnitude higher than \ours. 

\autoref{tab:full_soa} shows exhaustive results for all source-target domain combinations on \imagenetr, \nico, and \minidn. We compare \ours with baselines and competitors. \ours outperforms all baselines and competitors on all datasets. On \imagenetr (\autoref{tab:full_imgnet_sota}), \searle is the second best, while \compodiff, \weicom, and \magic surpass it for specific source/target couples. On \nico (\autoref{tab:full_nico_sota}), \magic is the second best. On \minidn (\autoref{tab:full_minidn_sota}), \compodiff and \searle are the second and third-best methods, respectively.

\subsection{Visualizations}

\autoref{fig:freedom_solo_imagenet_r} shows visualizations of the top-ranked database images of \ours on \imagenetr. We use \Th{Photo} as the source domain and convert it to any target domain. \ours can retrieve correct images in all cases. ~\autoref{fig:freedom_solo_sbir} shows visualizations of the top-ranked database images of \ours on \minidn. We perform \Th{Sketch} $\rightarrow$ \Th{Photo} conversion, i.e., sketch-based image retrieval~\cite{dey2019doodle, dutta2019semantically, yelamarthi2018zero}. Interestingly, \ours is performing well in this task, in contrast to \picword~\cite{pic2word}.

Furthermore, we present challenging cases where state-of-the-art methods underperform, and the performance of \ours is demonstrated. \autoref{fig:freedom_competitors_ltll} shows visualizations of the top-ranked database images of \ours \vs competitors on the instance-level dataset \ltll. \Th{Archive} $\rightarrow$ \Th{Today} and \Th{Today} $\rightarrow$ \Th{Archive} domain conversions are performed. We observe that the competitors confuse both domains and instances. ~\autoref{fig:freedom_competitors_nico} shows visualizations of the top-ranked database images of \ours \vs competitors on \nico. \Th{Autumn} $\rightarrow$ \Th{Dimlight} and \Th{Grass} $\rightarrow$ \Th{Autumn} domain conversions are performed. \ours has the best retrieval results, while the competitors fail almost everywhere. 

In our visual examples, we excluded exact duplicates, and we performed aspect ratio changes for better presentation.

\newpage
\begin{table*}[htbp]
    \centering
    \begin{minipage}{\textwidth}
        \centering
        \resizebox{0.8\textwidth}{!}{%
        \renewcommand{\arraystretch}{0.75}%
        \input{tab_supp/sota_imagenet_r}
        }
        \subcaption{\imagenetr}
        \label{tab:full_imgnet_sota}
    \end{minipage}

    \begin{minipage}{\textwidth}
        \centering
        \resizebox{0.8\textwidth}{!}{%
         \renewcommand{\arraystretch}{0.75}%
        \input{tab_supp/sota_nico}
        }
        \subcaption{\nico}
        \label{tab:full_nico_sota}
    \end{minipage}

    \begin{minipage}{\textwidth}
        \centering
        \resizebox{0.8\textwidth}{!}{%
         \renewcommand{\arraystretch}{0.75}%
        \input{tab_supp/sota_minidn}
        }
        \subcaption{\minidn}
        \label{tab:full_minidn_sota}
    \end{minipage}
    \caption{\emph{Domain conversion evaluated by mAP (\%)} on three datasets. Comparison of \ours with baselines and competitors. Across all methods, rows and columns represent the source and target domains, respectively. \Th{AVG}: average mAP over respective source-target domain combinations. \textbf{Bold:} best, \ts{magenta}: second-best.}
    \label{tab:full_soa}
\end{table*}

\newpage
\input{fig_supp/retrievals_freedom_solo_imagenet_r}
\input{fig_supp/retrievals_freedom_solo_sbir}
\input{fig_supp/retrievals_competitors_ltll}
\input{fig_supp/retrievals_competitors_nico}
\clearpage

\clearpage

%% file: tex/algo.tex
\fontsize{7pt}{9pt}\selectfont
\algrenewcommand\algorithmicindent{1em}%
\begin{algorithmic}[1]
\color{black}
\Procedure{FreeDom}{y, t, X, V, Z}
\color{gray}
\State $y: $ image query 
\State $t: $ text query
\State $V, \cV: $ textual memory (words vocabulary) and embeddings
\State $Z, \cZ: $ visual memory (external image set) and embeddings
\State $X, \cX: $ database images and embeddings
\color{black}
\State $\vy \gets f(y)$\comment{embedding of image query}
\State  $\{ \vy_1, \dots, \vy_k \} \gets \NN_k(\vy; \cZ)$\comment{$k$ nearest proxy images - including $\vy$}
\State $W^+ \gets \emptyset$\comment{collect all word inversions}
\For{$i \in 1 \ldots k$}\comment{loop over proxy images}
\State $W^+ \gets W^+ \cup \NN_n(\vy_i; \cV)$ \comment{invert proxy image - $n$ nearest words}
\EndFor
\State  $\{ \hat{w}_1, \dots, \hat{w}_m \}, \{ \hat{a}_1, \dots, \hat{a}_m \}  \gets \text{most-frequent}_m(W^+)$ \comment{$m$ most frequent words and frequencies}
\State $\vt \gets \text{zero vector}$
\For{$i \in 1 \ldots m$} \comment{loop over frequent words}
\State $\vt_i \gets g(\hat{w}_i \oplus t)$ \comment{composed query (\eg ``shark origami'') embedding}
\State $\vt \gets \vt + \hat{a}_i\vt_i$ \comment{aggregated query - equivalent to late fusion}
\EndFor
\State Rank images $x_j\in \cX$ based on similarity $\vt^\top x_j$ \comment{execute the final query}
\EndProcedure
\end{algorithmic}

%% file: tab_supp/advanced_imagenet_r.tex
\centering
\scriptsize
\setlength{\tabcolsep}{4pt}
\begin{tabular}{lrrrrrr}
\toprule
\Th{Method} &
\Th{Car}      & 
\Th{Ori}      & 
\Th{Pho}      & 
\Th{Scu}      & 
\Th{Toy}       &  
\Th{Avg}       \\
\midrule
 InstructPix2Pix & 3.90 & 5.70 & 1.97 & 5.70 & 5.62 & 4.58 \\
 \ours w/ img-cap & 15.11 & 6.70 & 19.77 & 18.08 & 16.58 & 15.24 \\
 \ours w/ captioners & 16.68 & 11.74 & 17.44 & 15.68 & 16.94 & 15.70 \\
\midrule \rowcolor{LightSteelBlue1}
\ours                      & \tb{35.97} & \tb{11.80} & \tb{27.97} & \tb{36.58} & \tb{37.21}  & \tb{29.91} \\ \bottomrule
\end{tabular}

%% file: tab_supp/advanced_nico.tex
\centering
\scriptsize
\setlength{\tabcolsep}{2.8pt}
\begin{tabular}{lrrrrrrr}
\toprule
\Th{Method} &
\Th{Aut} &
\Th{Dim} &
\Th{Gra} &
\Th{Out} &
\Th{Roc} &
\Th{Wat} &
\Th{Avg} \\ \midrule
 InstructPix2Pix & 4.18 & 2.66 & 4.60 & 4.78 & 5.19 & 3.56 & 4.16  \\
 \ours w/ img-cap & 15.56  & 11.64  & 19.34  & 19.18  & 17.56  & 13.81 & 16.18 \\
 \ours w/ captioners & 14.07  & 9.54  & 18.67  & 20.86  & 17.34  & 12.37 & 15.48 \\
\midrule \rowcolor{LightSteelBlue1}
\ours                      & \tb{24.35} & \tb{24.41} & \tb{30.06}  & \tb{30.51}  & \tb{26.92}  & \tb{20.37} & \tb{26.10} \\ \bottomrule
\end{tabular}

%% file: tab_supp/advanced_minidn.tex
\centering
\scriptsize
\setlength{\tabcolsep}{3pt}
\begin{tabular}{lrrrrr}
\toprule
\Th{Method} &
\Th{Clip}  &
\Th{Paint} &
\Th{Pho}    &
\Th{Ske}   &
\Th{Avg} \\ \midrule
 InstructPix2Pix & 8.57  & 8.86  & 7.08  & 7.20 &  7.93 \\
 \ours w/ img-cap & 21.88 & 17.54 & 31.78 & 15.35 & 21.64 \\
 \ours w/ captioners & 27.65 & 17.42 & 33.42 & 17.24 & 23.91 \\
\midrule  \rowcolor{LightSteelBlue1}
\ours                      & \tb{41.96} & \tb{31.65} & \tb{41.12} & \tb{34.36}  & \tb{37.27} \\ \bottomrule
\end{tabular}

%% file: tab_supp/advanced_leuven.tex
\centering
\scriptsize
\setlength{\tabcolsep}{6pt}
\begin{tabular}{lrrr}
\toprule
\Th{Method} &
\Th{Today} & \Th{Archive} & \Th{Avg} \\ \midrule
 InstructPix2Pix & 9.83  & 20.02  & 14.92 \\
 \ours w/ img-cap & \tb{42.58} & 19.16 & 30.87 \\
 \ours w/ captioners & 26.52 & 18.76 & 22.19 \\
\midrule \rowcolor{LightSteelBlue1}
\ours                      & 30.95 & \tb{35.52} & \tb{33.24} \\ \bottomrule
\end{tabular}

%% file: fig_supp/freedom_w_other_inversions.tex
\pgfplotsset{every tick label/.append style={font=\tiny}}
\tikzset{every mark/.append style={solid}}
\begin{tikzpicture}
\begin{axis}[
    width=0.7\linewidth,
    height=.45\linewidth,
    ymin=5, ymax=32,
    xmin=0, xmax=31,
    ylabel={mAP}, 
    xlabel={$k$},
    xlabel style = {yshift = 5pt},
    grid=both,
    legend style={
        cells={anchor=west}, 
        font=\scriptsize, 
        fill opacity=0.7, 
        row sep=1ex, 
        column sep=1ex,
        legend pos=north west,  
        at={(1.05,0.5)},  
        anchor=west, 
        /tikz/every even column/.append style={column sep=1ex},
        column sep=0pt, 
        column 1/.style={nodes={draw=none, fill=none}}, 
    },
    legend columns=1, 
]

\addplot[dotted, color=appleblue, mark=*, mark size=1.5, line width=1.2pt] coordinates {(1, 26.19) (3,29.08) (5,29.65) (10,28.9) (20,27.35) (30,26.11)};
\addlegendentry{Freedom {$^{n=1}_{m=k}$}};

\addplot[color=appleblue, mark=*, mark size=2.5, mark options={draw=black, line width=0.35pt}, line width=1.2pt] coordinates {(1, 25.75) (3,28.97) (5,30.07) (10,30.1) (20,29.86) (30,29.68)};
\addlegendentry{Freedom{$^{n=7}_{m=7}$}};

\addplot[color=appleorange, mark=*, mark size=2.5, mark options={draw=black, line width=0.35pt}, line width=1.2pt] coordinates {(1, 14.04) (3,16.59) (5,17.69) (10,18.55) (20,19.18) (30,19.46)};
\addlegendentry{searle};

\addplot[color=applegreen, mark=*, mark size=2.5, mark options={draw=black, line width=0.35pt}, line width=1.2pt] coordinates {(1, 7.88) (3,10.72) (5,11.65) (10,12.28) (20,12.48) (30,12.32)};
\addlegendentry{pic2word};

\end{axis}
\end{tikzpicture}

%% file: tab_supp/extended_memory.tex
\centering
\scriptsize
\setlength{\tabcolsep}{2.5pt}
\begin{tabular}{lccccc}
\toprule
\Th{Memory} &  \Th{Avg} & \Th{ImageNet-R} & \Th{MiniDn} & \Th{NICO++} & \Th{LTLL} \\ \midrule
\Th{No Memory} & 27.96 & 25.77 & 32.06 & 23.20 & 30.82 \\
\Th{LAION 40M} &28.57 & 25.00 & 33.85 & 24.31 & 31.11 \\
\Th{Database + LAION 40M} & 29.52 & 26.07 & 34.92 & 24.91 & 32.17 \\ 
\Th{Database} & 31.63 & 29.91 & 37.27 & 26.10 & 33.24 \\ \bottomrule
\end{tabular}

%% file: fig_supp/increasing_laion.tex
\pgfplotsset{every tick label/.append style={font=\tiny}}

\begin{tikzpicture}
\begin{axis}[
    width=0.7\linewidth,
    height=.45\linewidth,
    ymin=21, ymax=34,
    xmin=40000, xmax=400000000,
    ylabel={mAP}, 
    xlabel={$k$},
    xlabel style = {yshift = 5pt},
    xmode=log,
    xtick={40000, 400000, 4000000, 40000000, 400000000},
    ytick={25, 30},
    xticklabels={$40K$, $400K$, $4M$, $40M$, $400M$}, 
    scaled x ticks = true, 
    grid=both,
    legend style={
        cells={anchor=west}, 
        font=\scriptsize, 
        fill opacity=0.7, 
        row sep=3ex, 
        column sep=1ex,
        legend pos=north west,  
        at={(1.05,0.5)},  
        anchor=west, 
        /tikz/every even column/.append style={column sep=1ex},
        column sep=0pt, 
        column 1/.style={nodes={draw=none, fill=none}}, 
    },
    legend columns=1, 
]

\addplot[color=applegreen, line width=1.2pt] coordinates {(40000, 31.58) (400000, 31.58) (4000000, 31.58) (40000000, 31.58) (400000000, 31.58)};
\addlegendentry{Database};

\addplot[color=appleblue, mark=*, mark size=2.5, mark options={draw=black, line width=0.35pt}, line width=1.2pt] coordinates {(40000, 22.84) (400000, 27.79) (4000000, 28.64) (40000000, 28.57) (400000000, 28.55)};
\addlegendentry{LAION Only};

\addplot[color=appleorange, line width=1.2pt] coordinates {(40000, 27.96) (400000, 27.96) (4000000, 27.96) (40000000, 27.96) (400000000, 27.96)};
\addlegendentry{No Memory};

\end{axis}
\end{tikzpicture}

%% file: fig_supp/times.tex
\pgfplotsset{every tick label/.append style={font=\tiny}}

\begin{tikzpicture}
\begin{axis}[
    width=0.8\linewidth,
    height=.5\linewidth,
    name=plot1,
    ymin=0, ymax=35,
    xmin=0.0159, xmax=0.0185,
    ylabel={mAP}, 
    xlabel={query time (sec)},
    xlabel style = {yshift = 5pt, xshift=30pt},
    grid=both,
    scaled x ticks = false,
   xtick={0.016, 0.017, 0.018},
    ytick={10, 20, 30},
    xticklabels={0.016, 0.017, 0.018},
    xticklabel style={/pgf/number format/fixed}, 
    mark size=4pt,
    legend columns=-1
    ]

\addplot[color=appleblue, only marks, mark=*, mark size=2.5, mark options={draw=black, line width=0.35pt}, line width=1.2pt] coordinates {(0.016576, 26.18) (0.016928,28.49) (0.017625,29.18) (0.017733,29.86) (0.018262,30.89)};
\addplot[color=appleorange, only marks, mark=*, mark size=2.5, mark options={draw=black, line width=0.35pt}, line width=1.2pt] coordinates {(0.016181, 10.47)};
\addplot[color=applegreen, only marks, mark=*, mark size=2.5, mark options={draw=black, line width=0.35pt}, line width=1.2pt] coordinates {(0.016297, 7.88) };
\addplot[color=applepurple, only marks, mark=*, mark size=2.5, mark options={draw=black, line width=0.35pt}, line width=1.2pt] coordinates {(0.257517, 12.88) };

\node[anchor=south east, font=\fontsize{6}{4}\selectfont] at (axis cs:0.0176,16.5) {\parbox{2cm}{k=1 \linebreak n=1 \linebreak m=1}};
\node[anchor=south east, font=\fontsize{6}{4}\selectfont] at (axis cs:0.018,18.5) {\parbox{2cm}{k=20 \linebreak n=7 \linebreak m=1}};
\node[anchor=south east, font=\fontsize{6}{4}\selectfont] at (axis cs:0.01855,19) {\parbox{2cm}{k=20 \linebreak n=1 \linebreak m=7}};
\node[anchor=south east, font=\fontsize{6}{4}\selectfont] at (axis cs:0.01885,20.5) {\parbox{2cm}{k=20 \linebreak n=7 \linebreak m=7}};
\node[anchor=south east, font=\fontsize{6}{4}\selectfont] at (axis cs:0.01926,19) {\parbox{2cm}{k=20 \linebreak n=7 \linebreak m=7 \linebreak 236k}};
\end{axis}

\begin{axis}[
    width=0.4\linewidth,
    height=.5\linewidth,
    name=plot2,
    at={(plot1.right of east)},
    anchor=west,
    grid=both,
    ymin=0, ymax=35,
    xmin=0.24, xmax=0.275,
    xmode=normal,
    xshift=5pt,
    yticklabels=\empty,
    xtick={0.25, 0.27},
    ytick={10, 20, 30},
    xticklabel style={/pgf/number format/fixed}, 
    axis background/.style={fill=white},
    mark size=4pt,
    legend style={cells={anchor=west}, font =\scriptsize, fill opacity=0.7, row sep=-2.9pt, xshift=0pt, yshift=-21ex, inner sep=1pt},
    legend columns=-1
    ]
\addplot[color=appleblue, only marks, mark=*, mark size=2.5, mark options={draw=black, line width=0.35pt}, line width=1.2pt] coordinates {(0.016576, 26.18) (0.016928,28.49) (0.017625,29.18) (0.017733,29.86) (0.018262,30.89)};
\addlegendentry{Freedom};

\addplot[color=appleorange, only marks, mark=*, mark size=2.5, mark options={draw=black, line width=0.35pt}, line width=1.2pt] coordinates {(0.016181, 10.47)};
\addlegendentry{WeiCom};

\addplot[color=applegreen, only marks, mark=*, mark size=2.5, mark options={draw=black, line width=0.35pt}, line width=1.2pt] coordinates {(0.016297, 7.88) };
\addlegendentry{Pic2Word};

\addplot[color=applepurple, only marks, mark=*, mark size=2.5, mark options={draw=black, line width=0.35pt}, line width=1.2pt] coordinates {(0.257517, 12.88) };
\addlegendentry{CompoDiff};

\end{axis}
\end{tikzpicture}

%% file: tab_supp/sensitivity.tex
\centering
\scriptsize
\setlength{\tabcolsep}{12pt}
\begin{tabular}{lcccc}
\toprule
\mr{2}{\Th{Dataset}} & \mc{4}{ $l$ }
\\ \cmidrule{2-5}
           & 1 & 5 & 10  & 20  \\ \midrule
\Th{ImageNet-R} & -3.48 & -6.39 & -7.57 & -9.08 \\
\Th{NICO++}       & -1.03 & -2.52 & -3.35 & -4.42 \\
\Th{MiniDN}    & -2.08 & -5.32 & -6.77 & -8.37 \\
\Th{LTLL}     & -1.50 & -2.43 & -2.96 & -4.43 \\ \bottomrule
\end{tabular}

%% file: tab_supp/removed_words.tex
\centering
\tiny
\renewcommand{\arraystretch}{0.8}
\setlength{\tabcolsep}{.1pt}

\begin{tabular}{c@{\lsp}ccccc}
\toprule
\mc{6}{ \Th{ImageNet-R}} \\
\midrule
\Th{Object} & \Th{NN-1} & \Th{NN-2} & \Th{NN-3} & \Th{NN-4} & \Th{NN-5}
\\ \midrule

\textit{School bus} & \textit{School bus} & \textit{Bus} & \textit{Airport bus} & \textit{Minibus} & \textit{Bus driver} \\ 
\textit{Guillotine} & \textit{Guillotine} & \textit{Meat cutter} & \textit{Paper cutter} & \textit{Grindstone} & \textit{Lock} \\ 
\textit{Lawn mower} & \textit{Lawn mower} & \textit{Mower} & \textit{Riding mower} & \textit{Lawn} & \textit{Walk-behind mower} \\ 
\textit{African chameleon} & \textit{Chameleon} & \textit{Common chameleon} & \textit{Lizard} & \textit{Dragon lizard} & \textit{Reptile} \\ 
\textit{Basset} & \textit{Basset hound} & \textit{Basset artésien normand} & \textit{Beagle} & \textit{Spaniel} & \textit{Bulldog} \\ 
\textit{Beer glass} & \textit{Beer glass} & \textit{Pint glass} & \textit{Beer} & \textit{Wine glass} & \textit{Pint} \\ 
\textit{Collie} & \textit{Collie} & \textit{Australian collie} & \textit{Border collie} & \textit{Spaniel} & \textit{Wolf} \\ 
\textit{Golden retriever} & \textit{Golden retriever} & \textit{Retriever} & \textit{Goldendoodle} & \textit{Golden dream} & \textit{Puppy} \\

\midrule 
\mc{6}{ \Th{\nico} }
\\ \midrule 
\Th{Object} & \Th{NN-1} & \Th{NN-2} & \Th{NN-3} & \Th{NN-4} & \Th{NN-5}
\\ \midrule

\textit{Ostrich} & \textit{Ostrich} & \textit{Ostrich meat} & \textit{Emu} & \textit{Elephant} & \textit{Camel} \\ 
\textit{Bus} & \textit{Bus} & \textit{Airport bus} & \textit{School bus} & \textit{Bus driver} & \textit{Car} \\ 
\textit{Kangaroo} & \textit{Kangaroo} & \textit{Red kangaroo} & \textit{Koala} & \textit{Reindeer} & \textit{Camel} \\ 
\textit{Lifeboat} & \textit{Lifeboat} & \textit{Boat} & \textit{Speedboat} & \textit{Rescuer} & \textit{Jollyboat} \\ 
\textit{Airplane} & \textit{Airplane} & \textit{Aircraft} & \textit{Airliner} & \textit{Air travel} & \textit{Aviation} \\ 
\textit{Butterfly} & \textit{Butterfly} & \textit{Moths and butterflies} & \textit{Moth} & \textit{Insect} & \textit{Monarch butterfly} \\ 
\textit{Crocodile} & \textit{Crocodile} & \textit{Alligator} & \textit{Dinosaur} & \textit{Crocodilia} & \textit{Iguana} \\ 
\textit{Chair} & \textit{Chair} & \textit{Office chair} & \textit{Folding chair} & \textit{Club chair} & \textit{Throne}  \\

\midrule 
\mc{6}{ \Th{\minidn} }
\\ \midrule 
\Th{Object} & \Th{NN-1} & \Th{NN-2} & \Th{NN-3} & \Th{NN-4} & \Th{NN-5}
\\ \midrule
\textit{Sheep} & \textit{Sheep} & \textit{Wool} & \textit{Shepherd} & \textit{Livestock} & \textit{Flock} \\ 
\textit{Skateboard} & \textit{Skateboard} & \textit{Skateboarding} & \textit{Skate} & \textit{Skateboard deck} & \textit{Skateboarder} \\ 
\textit{Peanut} & \textit{Peanut} & \textit{Peanut butter} & \textit{Soy nut} & \textit{Bean} & \textit{Biscuit}  \\ 
\textit{Pig} & \textit{Pig} & \textit{Boar} & \textit{Pignolo} & \textit{Ham} & \textit{Rat} \\ 
\textit{Rhinoceros} & \textit{Rhinoceros} & \textit{Indian rhinoceros} & \textit{Hippopotamus} & \textit{Elephant} & \textit{Dinosaur} \\ 
\textit{Truck} & \textit{Truck} & \textit{Trailer truck} & \textit{Pickup truck} & \textit{Truck driver} & \textit{Truck racing} \\ 
\textit{Carrot} & \textit{Carrot} & \textit{Baby carrot} & \textit{Carrot cake} & \textit{Vegetable} & \textit{Root vegetable} \\ 
\textit{Pear} & \textit{Pear} & \textit{Asian pear} & \textit{European pear} & \textit{Apple} & \textit{Onion} \\ 

\midrule
\mc{6}{ \Th{\ltll} }
\\ \midrule
\Th{Object} & \Th{NN-1} & \Th{NN-2} & \Th{NN-3} & \Th{NN-4} & \Th{NN-5}
\\ \midrule
\textit{Notredame} & \textit{Cathedral} & \textit{Négociant} & \textit{Jesus} & \textit{Arena} & \textit{Château} \\ 
\textit{TempleTooth} & \textit{Tooth} & \textit{Temple} & \textit{Mouth} & \textit{Tombet} & \textit{Temple fade} \\ 
\textit{BigBen} & \textit{Clock} & \textit{Bell} & \textit{Man} & \textit{Bee} & \textit{Dollar} \\  
\textit{SacreCoeur} & \textit{Cemetery} & \textit{Tours (City)} & \textit{Church} & \textit{Grave} & \textit{Tomb} \\ 
\textit{TajMahal} & \textit{Elephant} & \textit{Tiger} & \textit{Mahlab} & \textit{Masala} & \textit{Naan} \\ 
\textit{Arcdetriomphe} & \textit{Triumphal arch} & \textit{Natural arch} & \textit{Arch} & \textit{Tunnel} & \textit{Gate} \\ 
\textit{Pettah} & \textit{Mustamakkara} & \textit{Pathiri} & \textit{Koottu} & \textit{Kozhukkatta} & \textit{Poriyal} \\ 
\textit{EiffelTower} & \textit{Skyscraper} & \textit{Tower} & \textit{Mountain} & \textit{Lighthouse} & \textit{Windmill} \\ 
\bottomrule
\end{tabular}

%% file: tab_supp/memory_inversion.tex
\footnotesize
\setlength{\tabcolsep}{4pt}
\renewcommand{\cellalign}{bc}
\begin{tabular}{lcccccccc} \\ \toprule
Query image &   NN-1  & NN-2       & NN-3   & NN-4  & NN-5   & NN-6& NN-7   \\ \midrule
\figsup[.07]{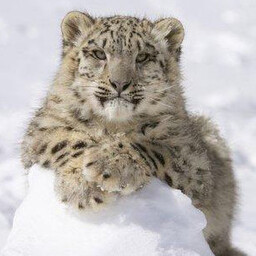} & 
 \makecell{\textit{Snow leopard} \\  1.00 \\   \phantom{}} &
  \makecell{\textit{Big cats} \\  0.80 \\   \phantom{}} &
   \makecell{\textit{Himalayan} \\  0.60 \\   \phantom{}} &
    \makecell{\textit{Clouded leopard} \\  0.45 \\   \phantom{}} &
     \makecell{\textit{Big cat} \\  0.35 \\  \phantom{}} &
      \makecell{\textit{Snowball} \\  0.25 \\   \phantom{}} &
       \makecell{\textit{Artic} \\  0.25 \\  \phantom{}}     
\\ \midrule
\figsup[.07]{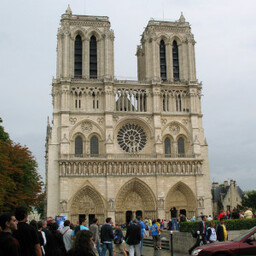} & 
 \makecell{\textit{Gothic} \\ \textit{architecture} \\ 1.00 \\  \phantom{}} &
  \makecell{\textit{Unesco world} \\ \textit{heritage site}\\ 1.00 \\  \phantom{}} &
   \makecell{\textit{Cathedral} \\  1.00 \\  \phantom{}} &
    \makecell{\textit{Medieval} \\ \textit{architecture} \\ 0.95 \\  \phantom{}} &
     \makecell{\textit{Classical} \\ \textit{architecture} \\ 0.90 \\  \phantom{}} &
      \makecell{\textit{Holy places} \\  0.90 \\  \phantom{}} &
       \makecell{\textit{Gothic} \\  0.30 \\  \phantom{}}     
\\ \midrule
\figsup[.07]{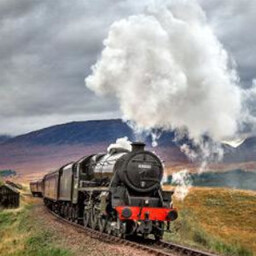} &
 \makecell{\textit{Steam engine} \\  1.00 \\  \phantom{}} &
  \makecell{\textit{Locomotive} \\  0.95 \\  \phantom{}} &
   \makecell{\textit{Train} \\  0.90 \\  \phantom{}} &
    \makecell{\textit{Steam} \\  0.75 \\  \phantom{}} &
     \makecell{\textit{Railway} \\  0.35 \\  \phantom{}} &
      \makecell{\textit{Railroad} \\ \textit{engineer} \\ 0.35 \\  \phantom{}} &
       \makecell{\textit{British rail} \\ \textit{class 81} \\ 0.35 \\  \phantom{}}     
\\ \midrule
\figsup[.07]{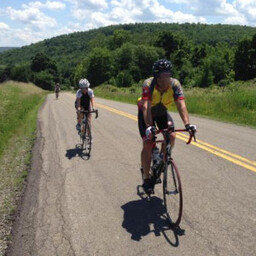} & 
 \makecell{\textit{Cross-country} \\ \textit{cycling} \\ 1.00 \\  \phantom{}} &
  \makecell{\textit{Endurance} \\ \textit{riding} \\ 0.85 \\  \phantom{}} &
   \makecell{\textit{Bicycles--Equipment} \\ \textit{and supplies} \\ 0.70 \\  \phantom{}} &
    \makecell{\textit{Road cycling} \\  0.70 \\  \phantom{}} &
     \makecell{\textit{Bicycle racing}  \\  0.65 \\  \phantom{}} &
      \makecell{\textit{Road bicycle} \\ \textit{racing} \\ 0.55 \\  \phantom{}} &
       \makecell{\textit{Endurance sports} \\  0.35 \\  \phantom{}} 
\\ \midrule
\figsup[.07]{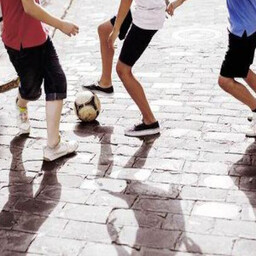} &
 \makecell{\textit{Soccer} \\  1.00 \\  \phantom{}} &
  \makecell{\textit{Street football} \\  0.93 \\  \phantom{}} &
   \makecell{\textit{Freestyle football} \\  0.93 \\  \phantom{}} &
    \makecell{\textit{Soccer kick} \\  0.64 \\  \phantom{}} &
     \makecell{\textit{Soccer ball}  \\  0.57 \\  \phantom{}} &
      \makecell{\textit{Kick (Sports)} \\  0.50 \\  \phantom{}} &
       \makecell{\textit{Street sports} \\  0.43 \\  \phantom{}} 
\\ \bottomrule

\end{tabular}

%% file: tab_supp/siglip_imagenet_r.tex
\centering
\scriptsize
\setlength{\tabcolsep}{6pt}
\begin{tabular}{lrrrrrr}
\toprule
\Th{Method} &
\Th{Car}      &
\Th{Ori}      &  
\Th{Pho}      &
\Th{Scu}      &
\Th{Toy}       &
\Th{Avg}       \\
\midrule
Text      & 0.88  & 0.80  & 0.62  & 0.95  & 0.90   & 0.83  \\
Image     & 4.97  & 3.70  & 0.84  & 8.18  & 7.40  & 5.02  \\
Text $\times$ Image & 6.57 & 4.34 & 4.89 & 6.46 & 7.46 & 5.94 \\
Text + Image        & 7.88  & 5.84  & 3.08  & 13.50  & 12.71  & 8.60  \\ \midrule
\rowcolor{LightSteelBlue1}
\ours                      & \tb{49.46} & \tb{27.12} & \tb{38.11} & \tb{47.52} & \tb{46.90}  & \tb{41.82} \\ \bottomrule
\end{tabular}

%% file: tab_supp/siglip_minidn.tex
\centering
\scriptsize
\setlength{\tabcolsep}{8pt}
\begin{tabular}{lrrrrr}
\toprule
\Th{Method} &
\Th{Clip}  &
\Th{Paint} &
\Th{Pho}    &
\Th{Ske}   &
\Th{Avg} \\ \midrule
Text      & 0.76  & 0.72  & 0.76  & 0.75    & 0.74  \\
Image     & 5.07  & 7.53  & 3.68  & 6.15    & 5.61  \\
Text $\times$ Image & 3.00 & 2.60 & 4.34 & 3.18 & 3.28 \\
Text + Image        & 7.79  & 11.33  & 10.80  & 9.02  & 9.74  \\ \midrule
\rowcolor{LightSteelBlue1}
\ours                      & \tb{57.14} & \tb{45.47} & \tb{59.71} & \tb{52.21}  & \tb{53.63} \\ \bottomrule
\end{tabular}

%% file: tab_supp/siglip_nico.tex
\centering
\scriptsize
\setlength{\tabcolsep}{4.5pt}
\begin{tabular}{lrrrrrrr}
\toprule
\Th{Method} &
\Th{Aut} &
\Th{Dim} &
\Th{Gra} &
\Th{Out} &
\Th{Roc} &
\Th{Wat} &
\Th{Avg} \\ \midrule
Text      & 1.08  & 1.13  & 1.04   & 1.26   & 1.10   & 1.11   & 1.12  \\
Image     & 6.19  & 5.19  & 5.42   & 7.67   & 7.44   & 5.62    & 6.25  \\ 
Text $\times$ Image & 2.31 & 2.91 & 3.26 & 3.53 & 3.25 & 2.90 & 3.03 \\
Text + Image        & 8.35 & 7.19 & 8.08  & 11.42 & 10.57 & 8.12   & 8.95  \\
\midrule \rowcolor{LightSteelBlue1}
\ours                      & \tb{30.28} & \tb{29.96} & \tb{33.86}  & \tb{37.16}  & \tb{33.14}  & \tb{26.49} & \tb{31.81} \\ \bottomrule
\end{tabular}

%% file: tab_supp/siglip_leuven.tex
\centering
\scriptsize
\setlength{\tabcolsep}{13.5pt}
\begin{tabular}{lrrr}
\toprule
\Th{Method} &
\Th{Today} & \Th{Archive} & \Th{Avg} \\ \midrule
Text    & 3.84  & 5.02  & 4.43  \\
Image     & 10.25  & 28.14 & 19.20 \\ 
Text $\times$ Image & 4.87 & 3.49 & 4.18 \\ 
Text + Image        & 10.16  & 26.73 & 18.44 \\ \midrule
\rowcolor{LightSteelBlue1}
\ours                      & \tb{27.45} & \tb{47.00} & \tb{37.22} \\ \bottomrule
\end{tabular}

%% file: tab_supp/general_cirr.tex
\centering
\fontsize{8}{6}\selectfont
\setlength{\tabcolsep}{9pt}
\begin{tabular}{lllll} 
\toprule
 \Th{Method}                  & R@1            & R@5            & R@10           & R@50           \\ \midrule
\picword           & 23.9           & 51.7           & 65.3           & 87.8           \\ 
\searle             & \textbf{24.2}  & 52.5           & 66.3           & 88.8           \\
\compodiff          & 18.2           & \textbf{53.1}  & \textbf{70.8}  & \textbf{90.3}  \\ \midrule
\rowcolor{LightSteelBlue1}
\ours       & 21.0           & 48.7           & 61.9           & 88.1           \\\rowcolor{LightSteelBlue1}
\ours$^*$ & 23.8           & 52.3           & 65.1           & 88.9           \\ \bottomrule

\end{tabular}

%% file: tab_supp/general_circo.tex
\centering
\fontsize{8}{6}\selectfont
\setlength{\tabcolsep}{4.5pt}
\begin{tabular}{lllll} 
\toprule
   \Th{Method}                 &  mAP@5           & mAP@10         & mAP@25          & mAP@50          \\ \midrule
\picword           & 8.7             & 9.5            & 10.7            & 11.3            \\ 
\searle             & 11.7            & 12.7           & 14.3            & 15.1            \\
\compodiff          & 12.6            & 13.4           & 15.8            & 16.4            \\ \midrule
\rowcolor{LightSteelBlue1}
\ours       & \textbf{14.0}   & \textbf{14.8}  & \textbf{16.4}   & \textbf{17.2}   \\ \rowcolor{LightSteelBlue1}
\ours$^*$ &  12.0            & 12.8           & 14.4            & 15.0            \\ \bottomrule

\end{tabular}

%% file: tab_supp/general_fashion.tex
\centering
\setlength{\tabcolsep}{14pt}
\fontsize{8}{6}\selectfont
\begin{tabular}{lllllllll}
\toprule
\multirow{2}{*}{\Th{Method}}  & \multicolumn{2}{c}{Dress} & \multicolumn{2}{c}{Shirt} & \multicolumn{2}{c}{Toptee} & \multicolumn{2}{c}{Average} \\ \cmidrule{2-9}
                 & R@10           & R@50           & R@10           & R@50           & R@10            & R@50           & R@10            & R@50            \\ \midrule
\picword         & 20.0           & 40.2           & 26.2           & 43.6           & 27.9            & 47.4           & 24.7            & 43.7            \\
\searle          & 20.5           & 43.1           & 26.9           & 45.6           & 29.3            & 50.0           & 25.6            & 46.2            \\
\compodiff       & \textbf{24.8}  & \textbf{44.8}  & \textbf{29.5}  & \textbf{47.4}  & \textbf{31.4}   & \textbf{53.7}  & \textbf{28.6}   & \textbf{48.6}   \\ \midrule \rowcolor{LightSteelBlue1}
\ours     & 16.8           & 36.3           & 23.5           & 38.5           & 24.7            & 43.7           & 21.6            & 39.5            \\
\rowcolor{LightSteelBlue1}
\ours$^*$  & 17.2           & 37.8           & 24.9           & 40.8           & 24.8            & 44.7           & 22.3            & 41.1 \\  \bottomrule           
\end{tabular}

%% file: tab_supp/recall_tables.tex
\scriptsize
\setlength{\tabcolsep}{8pt}
\begin{tabular}{lcccccccccc} \toprule
\multirow{2}{*}{\Th{Method}} & \multicolumn{2}{c}{\Th{Cartoon}} & \multicolumn{2}{c}{\Th{Origami}} & \multicolumn{2}{c}{\Th{Toy}} & \multicolumn{2}{c}{\Th{Sculpture}} & \multicolumn{2}{c}{\Th{Avg}} \\ \cmidrule{2-11}
& \Th{R@10}  & \Th{R@50}  & \Th{R@10}  & \Th{R@50}  & \Th{R@10}  & \Th{R@50}  & \Th{R@10}  & \Th{R@50}  & \Th{R@10}  & \Th{R@50}  \\ \midrule
Text                           & 0.15       & 0.95       & 0.87       & 3.73       & 0.71        & 1.77  &       0.36  &       1.89  & 0.52  & 2.09  \\
Image                          & 0.31       & 4.51       & 0.21       & 1.73       & 0.54        & 5.65  &       0.33  &       4.04  & 0.35  & 3.98  \\
Text + Image                   & 1.96       & 12.91      & 2.18       & 10.68      & 1.34        & 9.89  &       1.82  &       12.15 & 1.83  & 11.41 \\  \midrule
\picword                       & 8.00       & 21.90      & 13.50      & 25.60      & 8.70        & 21.60 &       10.00 &       23.80 & 10.05 & 23.23 \\
\picword  (CC-3M)              & 7.35       & 18.53      & 12.79      & 25.54      & 10.39       & 22.96 &       10.24 &       23.76 & 10.19 & 22.70 \\
\picword  (LAION 2B-en)        & 8.17       & 20.86      & 14.08      & 25.06      & 8.73        & 22.07 &       10.43 &       23.63 & 10.35 & 22.91 \\
ARTEMIS w/ CompoDiff dataset   & 11.42      & 23.81      & 15.49      & 25.44      & 11.21       & 24.01 &       10.84 &       21.07 & 12.24 & 23.58 \\
CLIP4Cir w/ CompoDiff dataset  & 10.90      & 24.12      & 16.08      & 25.60      & 11.01       & 23.57 &       10.45 &       21.86 & 12.11 & 23.79 \\
\compodiff (T5-XL)             & 8.43       & 20.40      & 15.73      & 25.69      & 11.19       & 22.48 &       9.19  &       18.45 & 11.14 & 21.76 \\
\compodiff (CLIP+T5-XL)        & 12.91      & 24.40      & 17.22      & 26.40      & 11.57       & 26.11 &       11.53 &       22.54 & 13.31 & 24.86 \\
\compodiff (CLIP)              & 13.21      & 24.06      & 17.03      & 26.17      & 11.22       & 26.25 &       11.24 &       22.96 & 13.18 & 24.86 \\  
KEDs                           & 14.80      & 34.20      & \ts{23.50} & 34.80      & 16.50       & 36.30 &       17.40 &       36.40 & 18.00 & 35.40 \\
\magic (original prompt)        & 9.95       & 22.37      & 5.07       & 17.58      & 11.51       & 26.76      & 7.92  &       19.70 & 8.61  & 21.60 \\
\magic          & 13.65      & 31.31      & 6.59       & 19.21      & 14.80       & 31.79      & 10.33       & 24.82 & 11.34 & 26.78 \\
\midrule
\weicom                        & 11.61      & 24.36      & 15.24      & 23.72      & 8.00        & 17.89      & 13.81 &       26.18 & 12.17 & 23.04 \\
\searle (default)              & 1.49       & 12.38      & 3.78       & 13.88      & 1.99        & 15.34      & 2.18  &       15.34 & 2.36  & 14.24 \\
\searle                & 10.17      & 30.32      & 17.02      & 32.00      & 8.23        & 9.10       & 11.60 &       32.41 & 11.76 & 30.96 \\
\cirevl                        & \ts{19.20} & \ts{42.80} & 22.2      & \tb{43.10}  & \tb{30.20}  & \ts{41.30} & \ts{23.40} &  \ts{45.00 }& \ts{23.75} & \ts{43.05} \\
\midrule

\rowcolor{LightSteelBlue1}
\ours                          & \tb{23.77} & \tb{48.83} & \tb{32.84} & \ts{42.82} & \ts{25.70} & \tb{47.59} & \tb{27.86} & \tb{48.96} & \tb{27.54} & \tb{47.05} \\ \bottomrule
\end{tabular}

%% file: tab_supp/sota_imagenet_r.tex
\scriptsize
\centering
\begin{tabular}{rrrrrrr@{\xlsp}rrrrrrr@{\xlsp}rrrrrrr} \toprule
\multicolumn{7}{c}{Text} & \multicolumn{7}{c}{Image} & \multicolumn{7}{c}{Text + Image} \\ \cmidrule(lr){1-7} \cmidrule(lr){8-14} \cmidrule(lr){15-21}
     & CART & ORI  & PHO  & SCU  & TOY  & AVG  &      & CART & ORI  & PHO  & SCU  & TOY  & AVG  &      & CART & ORI  & PHO  & SCU  & TOY  & AVG  \\
CART &      & 1.3  & 0.4  & 0.8  & 0.7  & 0.8  & CART &      & 0.7  & 11.7 & 1.5  & 3.2  & 4.3  & CART &      & 5.2  & 11.3 & 4.8  & 5.1  & 6.6  \\
ORI  & 0.7  &      & 0.4  & 1.0  & 0.4  & 0.6  & ORI  & 2.5  &      & 5.3  & 2.1  & 2.6  & 3.1  & ORI  & 5.4  &      & 5.0  & 3.7  & 3.7  & 4.5  \\
PHO  & 0.6  & 0.8  &      & 0.5  & 0.8  & 0.7  & PHO  & 1.5  & 0.2  &      & 0.7  & 1.0  & 0.8  & PHO  & 3.8  & 1.3  &      & 1.8  & 1.7  & 2.2  \\
SCU  & 0.9  & 1.3  & 0.4  &      & 0.6  & 0.8  & SCU  & 3.7  & 1.9  & 13.3 &      & 4.6  & 5.9  & SCU  & 9.2  & 8.8  & 12.4 &      & 6.3  & 9.2  \\
TOY  & 0.9  & 1.0  & 0.4  & 0.8  &      & 0.8  & TOY  & 4.2  & 1.2  & 11.7 & 3.3  &      & 5.1  & TOY  & 10.3 & 6.5  & 10.9 & 6.8  &      & 8.6  \\
AVG  & 0.8  & 1.1  & 0.4  & 0.8  & 0.6  & 0.7  & AVG  & 3.0  & 1.0  & 10.5 & 1.9  & 2.8  & 3.8  & AVG  & 7.2  & 5.4  & 9.9  & 4.3  & 4.2  & 6.2  \\ \midrule
\multicolumn{7}{c}{Text  $\times$  Image}               & \multicolumn{7}{c}{WeiCom}                     & \multicolumn{7}{c}{\picword}                   \\ \cmidrule(lr){1-7} \cmidrule(lr){8-14} \cmidrule(lr){15-21}
     & CART & ORI  & PHO  & SCU  & TOY  & AVG  &      & CART & ORI  & PHO  & SCU  & TOY  & AVG  &      & CART & ORI  & PHO  & SCU  & TOY  & AVG  \\
CART &      & 12.7 & 6.4  & 8.2  & 5.5  & 8.2  & CART &      & \ts{17.2} & 2.9  & 14.3 & 6.0  & 10.1 & CART &      & 8.2  & 7.5  & 8.4  & 6.3  & 7.6  \\
ORI  & 8.6  &      & 2.9  & 7.0  & 4.0  & 5.6  & ORI  & 11.7 &           & 1.7  & \tb{12.5} & 4.6  & 7.6  & ORI  & 7.8  &      & 3.7  & 6.8  & 3.8  & 5.5  \\
PHO  & 9.1  & 9.0  &      & 5.8  & 4.0  & 7.0  & PHO  & 13.5 & \ts{10.4} &      & \ts{11.1} & 5.2  & 10.1 & PHO  & 10.3 & 6.7  &      & 7.3  & 6.3  & 7.6  \\
SCU  & 9.2  & 15.3 & 6.2  &      & 5.1  & 9.0  & SCU  & 14.7 & \ts{22.7} & 2.5  &      & 5.1  & 11.3 & SCU  & 8.8  & 13.7 & 8.1  &      & 7.0  & 9.4  \\
TOY  & 11.1 & 12.0 & 5.5  & 9.1  &      & 9.4  & TOY  & 16.0 & \ts{21.8} & 2.4  & 13.4 &      & 13.4 & TOY  & 10.6 & 10.8 & 8.1  & 7.6  &      & 9.3  \\
AVG  & 9.5  & 12.3 & 5.3  & 7.5  & 4.7  & 7.8  & AVG  & 14.0 & \ts{18.0} & 2.4  & 12.8 & 5.2  & 10.5 & AVG  & 9.4  & 9.8  & 6.9  & 7.5  & 5.9  & 7.9  \\ \midrule
\multicolumn{7}{c}{\compodiff}                  & \multicolumn{7}{c}{\searle (default)}           & \multicolumn{7}{c}{\searle}             \\ \cmidrule(lr){1-7} \cmidrule(lr){8-14} \cmidrule(lr){15-21}
     & CART & ORI  & PHO  & SCU  & TOY  & AVG  &      & CART & ORI  & PHO  & SCU  & TOY  & AVG  &      & CART & ORI  & PHO  & SCU  & TOY  & AVG  \\
CART &      & 6.0  & \ts{27.3} & 13.4 & 8.2  & 13.7 & CART &      & 7.2  & 17.4 & 6.8  & 9.3  & 10.2 & CART &      & 16.2 & 26.3 & \ts{15.0} & \ts{15.0} & \ts{18.1} \\
ORI  & \ts{13.2} &      & \ts{12.4} & \ts{9.4}  & \ts{7.5}  & \ts{10.6} & ORI  & 4.3  &      & 5.8  & 3.5  & 4.3  & 4.5  & ORI  & 11.4 &      & 11.3 & 6.4  & 7.0  & 9.0  \\
PHO  & 10.6 & 5.8  &      & 9.2  & 9.4  & 8.8  & PHO  & 4.9  & 2.2  &      & 2.6  & 3.0  & 3.2  & PHO  & 14.6 & 9.9  &      & 8.1  & 7.1  & 9.9  \\
SCU  & 11.9 & 7.2  & \ts{30.7} &      & 10.9 & 15.2 & SCU  & 8.5  & 9.4  & 14.0 &      & 8.6  & 10.1 & SCU  & \ts{18.7} & 17.5 & 20.5 &      & \ts{12.3} & \ts{17.3} \\
TOY  & 16.0 & 7.3  & \ts{27.0} & \ts{14.3} &      & \ts{16.2} & TOY  & 8.4  & 6.5  & 13.7 & 7.0  &      & 8.9  & TOY  & \ts{19.1} & 14.3 & 18.9 & 11.0 &      & 15.8 \\
AVG  & 12.9 & 6.6  & \ts{24.4} & \ts{11.6} & 9.0  & 12.9 & AVG  & 6.5  & 6.3  & 12.7 & 5.0  & 6.3  & 7.4  & AVG  & 16.0 & 14.5 & 19.3 & 10.1 & \ts{10.3} & \ts{14.0} \\ \midrule
\multicolumn{7}{c}{\magic (original prompt)}        & \multicolumn{7}{c}{\magic}          & \multicolumn{7}{c}{\ours}                    \\ \cmidrule(lr){1-7} \cmidrule(lr){8-14} \cmidrule(lr){15-21}
     & CART & ORI  & PHO  & SCU  & TOY  & AVG  &      & CART & ORI  & PHO  & SCU  & TOY       & AVG  &      & CART & ORI  & PHO  & SCU  & TOY  & AVG  \\
CART &      & 4.4  & 3.7  & 5.4  & 7.0  & 5.1  & CART &      & 5.5  & 8.6  & 7.4  & 9.7       & 7.8  & CART &      & \tb{33.1} & \tb{48.8} & \tb{29.3} & \tb{32.7} & \tb{36.0} \\
ORI  & 5.8  &      & 2.1  & 3.8  & 3.0  & 3.7  & ORI  & 12.1 &      & 5.0  & 4.1  & 4.2       & 6.3  & ORI  & \tb{16.2} &      & \tb{13.3} & 8.8  & \tb{8.9}  & \tb{11.8} \\
PHO  & 12.3 & 3.3  &      & 6.4  & 9.9  & 8.0  & PHO  & \ts{18.7} & 4.3  &      & 8.4  & \ts{12.6} & \ts{11.0} & PHO  & \tb{36.5} & \tb{26.0} &      & \tb{23.8} & \tb{25.6} & \tb{28.0} \\
SCU  & 8.2  & 4.4  & 3.1  &      & 6.3  & 5.5  & SCU  & 16.3 & 6.1  & 8.9  &      & 8.4       & 9.9  & SCU  & \tb{38.0} & \tb{35.5} & \tb{43.9} &      & \tb{28.9} & \tb{36.6} \\
TOY  & 9.6  & 4.8  & 3.8  & 6.2  &      & 6.1  & TOY  & 18.0 & 6.5  & 10.1 & 7.7  &           & 10.6 & TOY  & \tb{43.2} & \tb{36.5} & \tb{41.0} & \tb{28.1} &      & \tb{37.2} \\
AVG  & 9.0  & 4.2  & 3.2  & 5.4  & 6.6  & 5.7  & AVG  & \ts{16.3} & 5.6  & 8.1  & 6.9  & 8.7  & 9.1  & AVG  & \tb{33.5} & \tb{32.8} & \tb{36.8} & \tb{22.5} & \tb{24.0} & \tb{29.9} \\ \bottomrule
\end{tabular}

%% file: tab_supp/sota_nico.tex
\scriptsize
\centering
\begin{tabular}{rrrrrrrr@{\xlsp}rrrrrrrr@{\xlsp}rrrrrrrr} \toprule
\multicolumn{8}{c}{Text}                            & \multicolumn{8}{c}{Image}                           & \multicolumn{8}{c}{Text + Image}                       \\ 
\cmidrule(lr){1-8} \cmidrule(lr){9-16} \cmidrule(lr){17-24}
    & AUT  & DIM  & GRA  & OUT & ROC  & WAT  & AVG  &     & AUT  & DIM  & GRA  & OUT & ROC  & WAT  & AVG  &     & AUT  & DIM  & GRA  & OUT  & ROC  & WAT  & AVG  \\
AUT &      & 1.4  & 1.0  & 0.4 & 0.8  & 1.4  & 1.0  & AUT &      & 6.2  & 13.1 & 4.8 & 4.2  & 3.9  & 6.5  & AUT &      & 11.1 & 14.3 & 4.2  & 5.8  & 7.1  & 8.5  \\
DIM & 1.6  &      & 0.9  & 0.3 & 0.8  & 1.3  & 1.0  & DIM & 4.3  &      & 8.2  & 3.9 & 3.3  & 4.6  & 4.9  & DIM & 9.1  &      & 9.2  & 3.3  & 4.3  & 7.0  & 6.6  \\
GRA & 1.6  & 1.4  &      & 0.3 & 0.8  & 1.6  & 1.2  & GRA & 5.9  & 5.5  &      & 6.2 & 4.8  & 6.0  & 5.7  & GRA & 13.0 & 10.7 &      & 5.4  & 6.7  & 10.3 & 9.2  \\
OUT & 1.7  & 1.4  & 1.0  &     & 0.8  & 1.4  & 1.2  & OUT & 4.4  & 6.0  & 13.4 &     & 5.7  & 8.9  & 7.7  & OUT & 10.1 & 11.8 & 16.9 &      & 7.7  & 13.1 & 11.9 \\
ROC & 1.5  & 1.4  & 0.8  & 0.4 &      & 1.4  & 1.1  & ROC & 5.4  & 6.2  & 12.4 & 7.2 &      & 7.0  & 7.7  & ROC & 12.5 & 11.4 & 14.8 & 6.4  &      & 10.9 & 11.2 \\
WAT & 1.8  & 1.4  & 0.9  & 0.4 & 0.8  &      & 1.1  & WAT & 3.3  & 5.3  & 9.4  & 6.2 & 4.1  &      & 5.7  & WAT & 8.3  & 10.1 & 12.5 & 5.6  & 5.5  &      & 8.4  \\
AVG & 1.7  & 1.4  & 0.9  & 0.4 & 0.8  & 1.4  & 1.1  & AVG & 4.7  & 5.8  & 11.3 & 5.7 & 4.4  & 6.1  & 6.3  & AVG & 10.6 & 11.0 & 13.6 & 5.0  & 6.0  & 9.7  & 9.3  \\ \midrule
\multicolumn{8}{c}{Text  $\times$  Image}                    & \multicolumn{8}{c}{\weicom}                          & \multicolumn{8}{c}{\picword}                         \\
\cmidrule(lr){1-8} \cmidrule(lr){9-16} \cmidrule(lr){17-24}
    & AUT  & DIM  & GRA  & OUT & ROC  & WAT  & AVG  &     & AUT  & DIM  & GRA  & OUT & ROC  & WAT  & AVG  &     & AUT  & DIM  & GRA  & OUT  & ROC  & WAT  & AVG  \\
AUT &      & 13.9 & 7.2  & 1.9 & 7.9  & 10.3 & 8.2  & AUT &      & 15.1 & 6.9  & 1.2 & 7.5  & 12.2 & 8.6  & AUT &      & 12.0 & 10.1 & 3.4  & 8.2  & 15.1 & 9.8  \\
DIM & 10.9 &      & 5.0  & 1.4 & 5.8  & 8.8  & 6.4  & DIM & 14.8 &      & 5.5  & 0.9 & 5.6  & 10.2 & 7.4  & DIM & 12.7 &      & 6.2  & 2.7  & 5.7  & 13.2 & 8.1  \\
GRA & 15.7 & 17.3 &      & 2.4 & 10.3 & 14.9 & 12.1 & GRA & 19.6 & 18.5 &      & 1.2 & 9.4  & 16.5 & 13.0 & GRA & 16.0 & 10.5 &      & 4.2  & 8.2  & 17.2 & 11.2 \\
OUT & 13.4 & 17.0 & 9.5  &     & 9.6  & 14.1 & 12.7 & OUT & 16.7 & 17.5 & 9.1  &     & 8.1  & 14.5 & 13.2 & OUT & 12.4 & 9.8  & 10.1 &      & 8.1  & 16.0 & 11.3 \\
ROC & 14.7 & 15.6 & 7.6  & 2.7 &      & 11.6 & 10.5 & ROC & 18.0 & 16.8 & 7.5  & 1.5 &      & 12.8 & 11.3 & ROC & 15.0 & 10.7 & 9.4  & 4.9  &      & 15.1 & 11.0 \\
WAT & 12.2 & 14.6 & 7.7  & 2.3 & 7.5  &      & 8.8  & WAT & 15.9 & 15.9 & 8.6  & 1.4 & 7.0  &      & 9.7  & WAT & 9.6  & 8.5  & 7.8  & 4.5  & 5.5  &      & 7.2  \\
AVG & 13.4 & 15.7 & 7.4  & 2.1 & 8.2  & 12.0 & 9.8  & AVG & 17.0 & 16.8 & 7.5  & 1.2 & 7.5  & 13.2 & 10.5 & AVG & 13.1 & 10.3 & 8.7  & 3.9  & 7.2  & 15.3 & 9.8  \\ \midrule
\multicolumn{8}{c}{CompoDiff}                       & \multicolumn{8}{c}{\searle (default)}                & \multicolumn{8}{c}{\searle}                   \\
\cmidrule(lr){1-8} \cmidrule(lr){9-16} \cmidrule(lr){17-24}
    & AUT  & DIM  & GRA  & OUT & ROC  & WAT  & AVG  &     & AUT  & DIM  & GRA  & OUT & ROC  & WAT  & AVG  &     & AUT  & DIM  & GRA  & OUT  & ROC  & WAT  & AVG  \\
AUT &      & 13.9 & 14.2 & 4.4 & 7.6  & 10.4 & 10.1 & AUT &      & 9.3  & 14.0 & 5.6      & 7.7  & 10.0 & 9.3  & AUT &      & 16.3 & 15.0 & 5.3  & 12.3 & 18.6 & 13.5 \\
DIM & 13.7 &      & 7.9  & 4.5 & 5.6  & 7.5  & 7.8  & DIM & 11.7 &      & 10.5 & 5.1      & 6.3  & 10.6 & 8.8  & DIM & 19.9 &      & 12.8 & 5.1  & \ts{11.1} & \ts{19.7} & 13.7 \\
GRA & 15.0 & 12.9 &      & 4.2 & 7.9  & 12.7 & 10.5 & GRA & 14.6 & 9.3  &      & \ts{7.6} & 9.0  & 14.2 & 10.9 & GRA & 25.9 & 18.3 &      & 6.7  & \ts{14.8} & 23.9 & 17.9 \\
OUT & 12.5 & 12.8 & 12.0 &     & 7.9  & 12.0 & 11.4 & OUT & 11.4 & 9.0  & 16.8 &          & 9.7  & 16.4 & 12.6 & OUT & 19.6 & 16.5 & 17.6 &      & \ts{14.0} & 22.4 & 18.0 \\
ROC & 15.8 & 14.7 & 12.3 & 5.5 &      & 11.4 & 11.9 & ROC & 13.0 & 8.8  & 14.2 & \ts{8.0} &      & 13.3 & 11.4 & ROC & 21.7 & 15.3 & 15.9 & 6.9  &      & 19.3 & 15.8 \\
WAT & 13.0 & 13.3 & 12.2 & 5.1 & 7.2  &      & 10.1 & WAT & 9.3  & 7.9  & 12.3 & \ts{7.5} & 7.1  &      & 8.8  & WAT & 15.2 & 14.3 & 13.8 & 6.0  & \ts{10.0} &      & 11.8 \\
AVG & 14.0 & 13.5 & 11.7 & 4.7 & 7.2  & 10.8 & 10.3 & AVG & 12.0 & 8.9  & 13.6 & \ts{6.8} & 7.9  & 12.8 & 10.3 & AVG & 20.5 & 16.1 & 15.0 & 6.0  & \ts{12.4} & 20.8 & 15.1 \\ \midrule
\multicolumn{8}{c}{\magic (original prompt)}             & \multicolumn{8}{c}{\magic}               & \multicolumn{8}{c}{\ours}                          \\
\cmidrule(lr){1-8} \cmidrule(lr){9-16} \cmidrule(lr){17-24}
    & AUT  & DIM  & GRA  & OUT & ROC  & WAT  & AVG  &     & AUT       & DIM       & GRA       & OUT      & ROC  & WAT  & AVG  &     & AUT  & DIM  & GRA  & OUT  & ROC  & WAT  & AVG  \\
AUT &      & 19.7 & 16.8 & 4.8 & 7.3  & 15.4 & 12.8 & AUT &           & \ts{28.3} & \tb{25.0} & \ts{5.7} & \ts{12.4} & \ts{22.5} & \ts{18.8} & AUT &      & \tb{33.5} & \ts{22.1} & \tb{9.0}  & \tb{24.0} & \tb{33.2} & \tb{24.4} \\
DIM & 17.3 &      & 12.0 & 4.6 & 5.8  & 12.8 & 10.5 & DIM & \ts{22.1} &           & \ts{18.4} & \ts{6.2} & 9.7       & 19.5      & \ts{15.2} & DIM & \tb{37.0} &      & \tb{20.4} & \tb{9.2}  & \tb{21.8} & \tb{33.6} & \tb{24.4} \\
GRA & 23.9 & 21.1 &      & 5.9 & 8.7  & 19.3 & 15.8 & GRA & \ts{30.2} & \ts{31.5} &           & 7.1      & 14.5 & \ts{27.4} & \ts{22.1}      & GRA & \tb{41.1} & \tb{35.8} &      & \tb{10.1} & \tb{26.3} & \tb{37.0} & \tb{30.1} \\
OUT & 20.8 & 18.1 & 17.2 &     & 8.4  & 17.9 & 16.5 & OUT & \ts{25.9} & \ts{28.1} & \tb{25.0} &          & 13.5 & \ts{25.6} & \ts{23.6}      & OUT & \tb{36.1} & \tb{33.5} & \ts{24.2} &      & \tb{24.2} & \tb{34.6} & \tb{30.5} \\
ROC & 22.9 & 15.5 & 15.8 & 6.1 &      & 15.6 & 15.2 & ROC & \ts{29.4} & \ts{25.5} & \tb{24.0} & 7.6      &      & \ts{23.5} & \ts{22.0}      & ROC & \tb{37.5} & \tb{31.2} & \ts{22.9} & \tb{10.2} &      & \tb{32.8} & \tb{26.9} \\
WAT & 16.9 & 14.3 & 13.7 & 5.4 & 6.2  &      & 11.3 & WAT & \ts{21.5} & \ts{22.6} & \tb{21.3} & 6.6      & 9.6  &      & \ts{16.3}           & WAT & \tb{28.7} & \tb{27.6} & \ts{18.9} & \tb{8.4}  & \tb{18.3} &      & \tb{20.4} \\
AVG & 20.4 & 17.7 & 15.1 & 5.4 & 7.3  & 16.2 & 13.7 & AVG & \ts{25.8} & \ts{27.2} & \tb{22.7} & 6.6      & 11.9 & \ts{23.7} & \ts{19.7}      & AVG & \tb{36.1} & \tb{32.3} & \ts{21.7} & \tb{9.4}  & \tb{22.9} & \tb{34.2} & \tb{26.1} \\ \bottomrule
\end{tabular}

%% file: tab_supp/sota_minidn.tex
\scriptsize
\centering
\begin{tabular}{rrrrrr@{\xlsp}rrrrrr@{\xlsp}rrrrrr} \toprule
\multicolumn{6}{c}{Text}                & \multicolumn{6}{c}{Image}              & \multicolumn{6}{c}{Text + Image}         \\
 \cmidrule(lr){1-6} \cmidrule(lr){7-12} \cmidrule(lr){13-18}
     & CLI  & PAI  & PHO  & SKE  & AVG  &     & CLI  & PAI  & PHO  & SKE  & AVG  &     & CLI  & PAI  & PHO  & SKE  & AVG  \\
CLI  &      & 0.7  & 0.4  & 0.8  & 0.6  & CLI &      & 2.4  & 10.9 & 8.2  & 7.2  & CLI &      & 4.4  & 9.9  & 14.5 & 9.6  \\
PAI  & 0.4  &      & 0.4  & 0.8  & 0.5  & PAI & 5.1  &      & 11.7 & 5.1  & 7.3  & PAI & 9.2  &      & 10.9 & 9.8  & 10.0 \\
PHO  & 0.4  & 0.7  &      & 0.8  & 0.6  & PHO & 6.1  & 3.2  &      & 3.8  & 4.4  & PHO & 11.3 & 7.4  &      & 9.0  & 9.2  \\
SKE  & 0.4  & 0.7  & 0.4  &      & 0.5  & SKE & 10.1 & 4.0  & 9.3  &      & 7.8  & SKE & 11.3 & 6.3  & 7.9  &      & 8.5  \\
AVG  & 0.4  & 0.7  & 0.4  & 0.8  & 0.6  & AVG & 7.1  & 3.2  & 10.6 & 5.7  & 6.7  & AVG & 10.6 & 6.1  & 9.6  & 11.1 & 9.3  \\ \midrule
\multicolumn{6}{c}{Text  $\times$  Image}        & \multicolumn{6}{c}{\weicom}             & \multicolumn{6}{c}{\picword}           \\ \cmidrule(lr){1-6} \cmidrule(lr){7-12} \cmidrule(lr){13-18}
     & CLI  & PAI  & PHO  & SKE  & AVG  &     & CLI  & PAI  & PHO  & SKE  & AVG  &     & CLI  & PAI  & PHO  & SKE  & AVG  \\
CLI  &      & 7.1  & 4.5  & 15.4 & 9.0  & CLI &      & 8.3  & 1.6  & 12.8 & 7.5  & CLI &      & 11.7 & 9.5  & 19.0 & 13.4 \\
PAI  & 9.6  &      & 5.3  & 11.1 & 8.7  & PAI & 8.3  &      & 1.8  & 11.1 & 7.0  & PAI & 7.8  &      & 6.8  & 11.3 & 8.6  \\
PHO  & 15.4 & 13.4 &      & 18.9 & 15.9 & PHO & 10.7 & 14.1 &      & 20.6 & 15.1 & PHO & 15.1 & 18.4 &      & 20.4 & 18.0 \\
SKE  & 6.8  & 7.7  & 3.2  &      & 5.9  & SKE & 4.2  & 7.7  & 1.3  &      & 4.4  & SKE & 7.7  & 9.8  & 6.6  &      & 8.0  \\
AVG  & 10.6 & 9.4  & 4.4  & 15.1 & 9.9  & AVG & 7.7  & 10.0 & 1.5  & 14.8 & 8.5  & AVG & 10.2 & 13.3 & 7.6  & 16.9 & 12.0 \\ \midrule
\multicolumn{6}{c}{\compodiff}           & \multicolumn{6}{c}{\searle (default)}   & \multicolumn{6}{c}{\searle}     \\ 
\cmidrule(lr){1-6} \cmidrule(lr){7-12} \cmidrule(lr){13-18}
     & CLI  & PAI  & PHO  & SKE  & AVG  &     & CLI  & PAI  & PHO  & SKE  & AVG  &     & CLI  & PAI  & PHO  & SKE  & AVG  \\
CLI  &             & 13.9 & \ts{23.8} & 19.5         & 19.1      & CLI &      & 11.2 & 17.4 & 16.8 & 15.1 & CLI &      & \ts{20.9} & 21.9 & 32.3 & \ts{25.0} \\
PAI  & \ts{24.5}   &           & \ts{30.1}    & 18.2 & \ts{24.3} & PAI & 9.0  &      & 13.9 & 8.6  & 10.5 & PAI & 16.7 &      & 18.2 & 21.3 & 18.7 \\
PHO  & \ts{27.9}   & 21.0      &              & 21.4 & 23.4      & PHO & 10.1 & 11.6 &      & 8.0  & 9.9  & PHO & 20.1 & 23.3 &      & 27.9 & 23.7 \\
SKE  & \ts{25.1}   & \ts{21.6} & \ts{28.5}    &      & \ts{25.1} & SKE & 12.3 & 11.9 & 13.3 &      & 12.5 & SKE & 18.2 & 20.9 & 19.8 &      & 19.6 \\
AVG  & \ts{25.8}   & 18.8      & \ts{27.5}    & 19.7 & \ts{23.0} & AVG & 10.4 & 11.5 & 14.9 & 11.2 & 12.0 & AVG & 18.3 & \ts{21.7} & 20.0 & 27.2 & 21.8 \\ \midrule
\multicolumn{6}{c}{\magic (original prompt)} & \multicolumn{6}{c}{\magic}  & \multicolumn{6}{c}{\ours}            \\
 \cmidrule(lr){1-6} \cmidrule(lr){7-12} \cmidrule(lr){13-18}
     & CLI  & PAI  & PHO  & SKE  & AVG  &     & CLI  & PAI  & PHO  & SKE  & AVG  &     & CLI  & PAI  & PHO  & SKE  & AVG  \\
CLI  &      & 9.0  & 7.7  & 17.5 & 11.4 & CLI &      & 18.6      & 17.7 & \ts{36.9} & 24.4      & CLI &               & \tb{45.4} &\tb{ 34.2} & \tb{46.3} & \tb{42.0} \\
PAI  & 6.7  &      & 6.2  & 15.7 & 9.5  & PAI & 12.2 &           & 12.4 & \ts{28.0} & 17.5      & PAI & \tb{25.8} &               & \tb{30.4} & \tb{38.8} & \tb{31.7} \\
PHO  & 11.3 & 12.9 &      & 24.3 & 16.2 & PHO & 19.2 & \ts{24.7} &      & \ts{41.9} & \ts{28.6} & PHO & \tb{30.3} & \tb{46.4} &               & \tb{46.7} & \tb{41.1} \\
SKE  & 3.2  & 6.8  & 3.3  &      & 4.4  & SKE & 7.6  & 13.6      & 8.0  &           & 9.7       & SKE & \tb{26.7} & \tb{43.2} & \tb{33.2} &               & \tb{34.4} \\
AVG  & 7.1  & 9.6  & 5.7  & 19.1 & 10.4 & AVG & 13.0 & 19.0      & 12.7 & \ts{35.6} & 20.1      & AVG & \tb{27.6} & \tb{45.0} & \tb{32.6} & \tb{43.9} & \tb{37.3} \\ \bottomrule
\end{tabular}

%% file: fig_supp/retrievals_freedom_solo_imagenet_r.tex
\begin{figure*}
\scriptsize
\centering
\setlength{\tabcolsep}{0.8pt}
\newcommand{\sz}{.08}
\begin{tabular}{cccccccccc}

\makecell{ \\ (a) \\ \hspace{20pt} \\ [-10pt] \phantom{}} &
\bfigsupq[\sz]{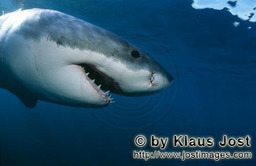} &
\bfigsupc[\sz]{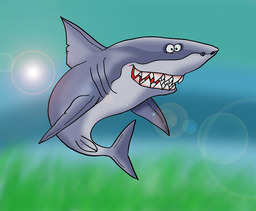} &
\bfigsupc[\sz]{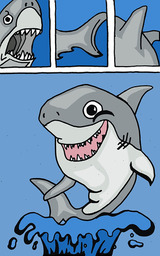} &
\bfigsupc[\sz]{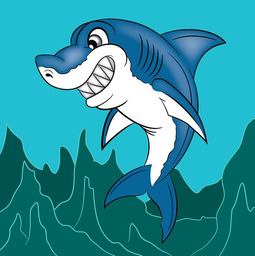} &
\bfigsupc[\sz]{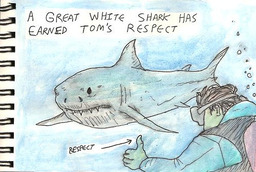} &
\bfigsupc[\sz]{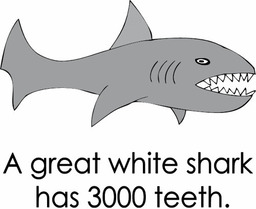} &
\bfigsupc[\sz]{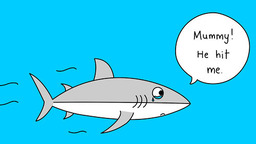} &
\bfigsupc[\sz]{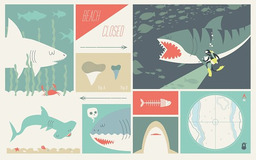} &
\bfigsupc[\sz]{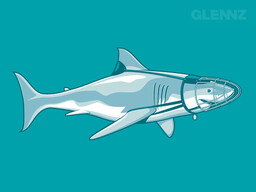} 
\\ \vspace{15pt}

&
\bfigsupq[\sz]{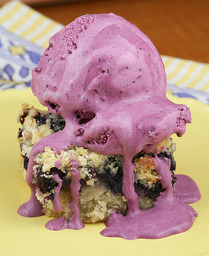} &
\bfigsupc[\sz]{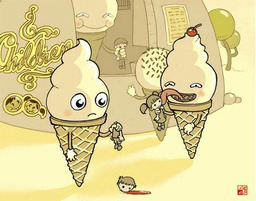} &
\bfigsupc[\sz]{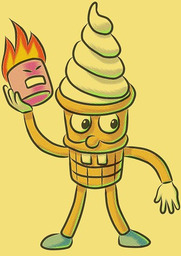} &
\bfigsupc[\sz]{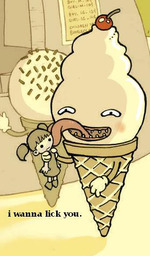} &
\bfigsupc[\sz]{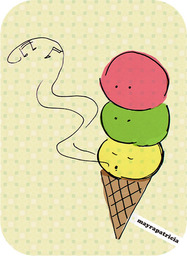} &
\bfigsupc[\sz]{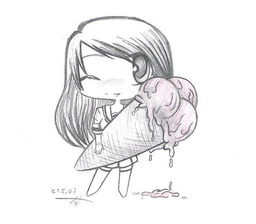} &
\bfigsupc[\sz]{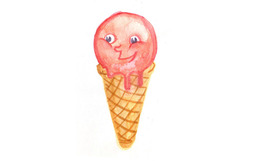} &
\bfigsupw[\sz]{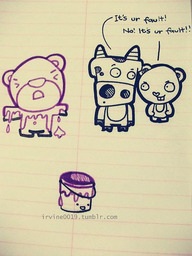} &
\bfigsupc[\sz]{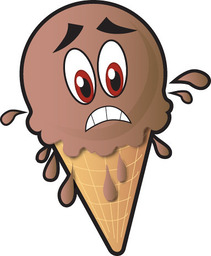} 
\\

\makecell{ \\ (b) \\ \hspace{20pt} \\ [-10pt] \phantom{}} &
\bfigsupq[\sz]{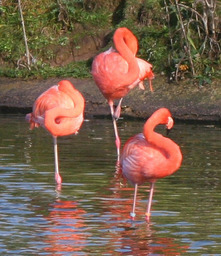} &
\bfigsupc[\sz]{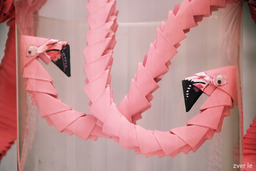} &
\bfigsupc[\sz]{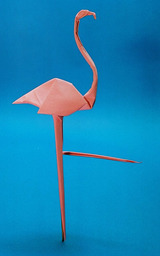} &
\bfigsupc[\sz]{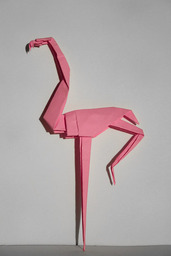} &
\bfigsupc[\sz]{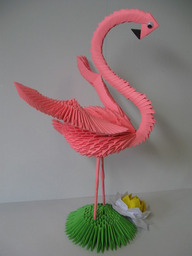} &
\bfigsupc[\sz]{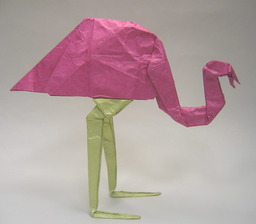} &
\bfigsupc[\sz]{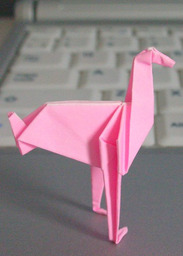} &
\bfigsupc[\sz]{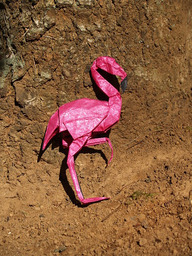} &
\bfigsupc[\sz]{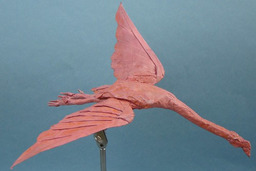} 
\\ \vspace{15pt}

&
\bfigsupq[\sz]{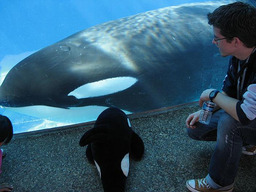} &
\bfigsupc[\sz]{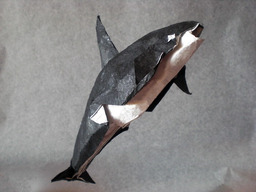} &
\bfigsupc[\sz]{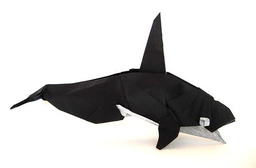} &
\bfigsupc[\sz]{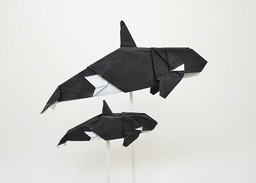} &
\bfigsupc[\sz]{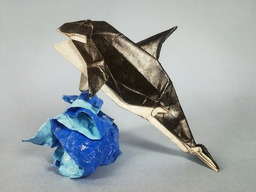} &
\bfigsupc[\sz]{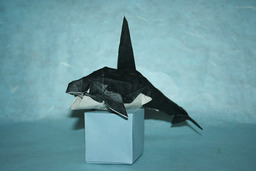} &
\bfigsupw[\sz]{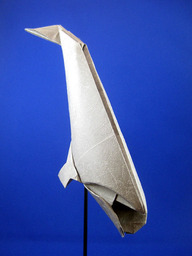} &
\bfigsupc[\sz]{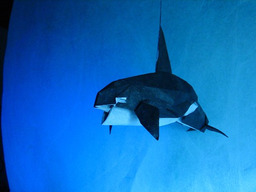} &
\bfigsupc[\sz]{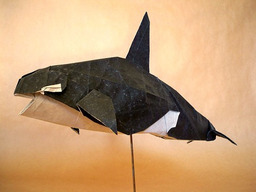}  
\\

\makecell{ \\ (c) \\ \hspace{20pt} \\ [-10pt] \phantom{}} &
\bfigsupq[\sz]{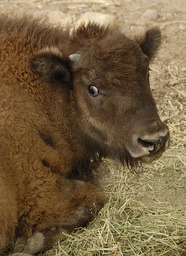} &
\bfigsupc[\sz]{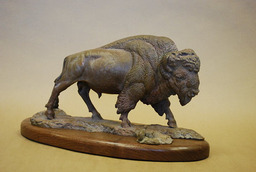} &
\bfigsupc[\sz]{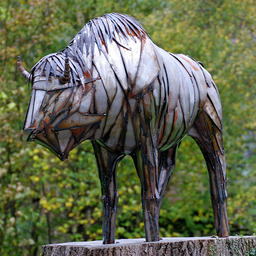} &
\bfigsupc[\sz]{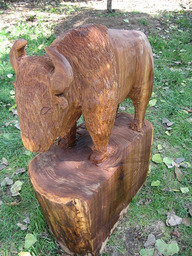} &
\bfigsupc[\sz]{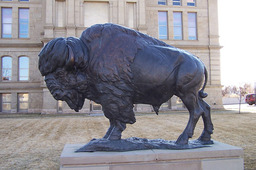} &
\bfigsupc[\sz]{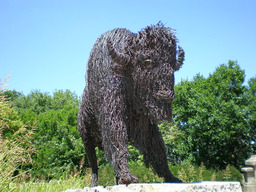} &
\bfigsupw[\sz]{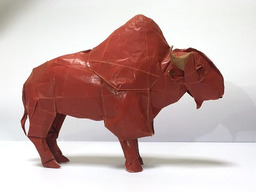} &
\bfigsupc[\sz]{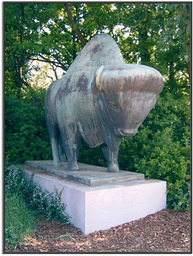} &
\bfigsupc[\sz]{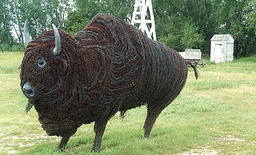}  
\\ \vspace{15pt}

&
\bfigsupq[\sz]{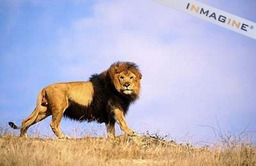} &
\bfigsupc[\sz]{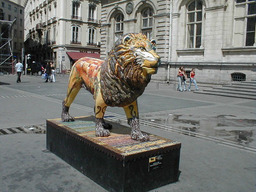} &
\bfigsupc[\sz]{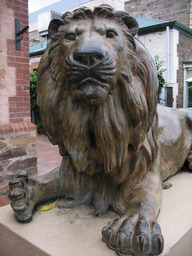} &
\bfigsupw[\sz]{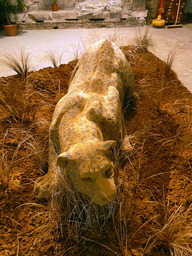} &
\bfigsupw[\sz]{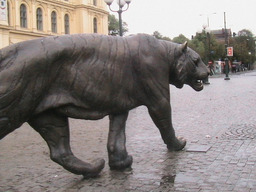} &
\bfigsupc[\sz]{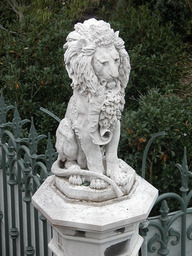} &
\bfigsupc[\sz]{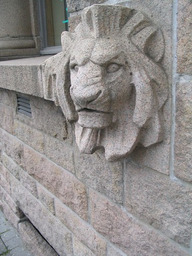} &
\bfigsupc[\sz]{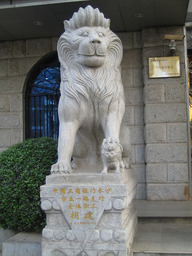} &
\bfigsupw[\sz]{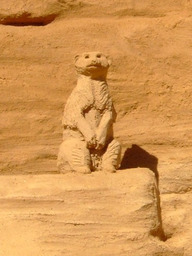}  
\\

\makecell{ \\ (d) \\ \hspace{20pt} \\ [-10pt] \phantom{}} &
\bfigsupq[\sz]{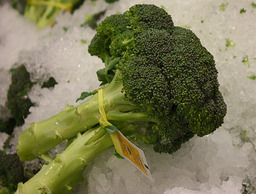} &
\bfigsupc[\sz]{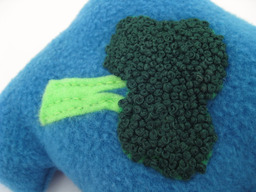} &
\bfigsupc[\sz]{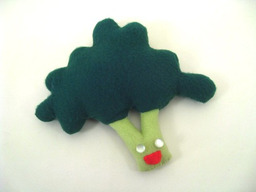} &
\bfigsupc[\sz]{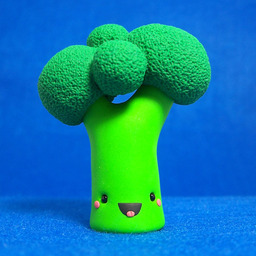} &
\bfigsupc[\sz]{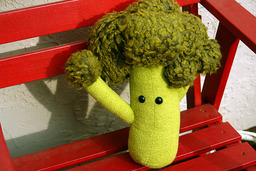} &
\bfigsupc[\sz]{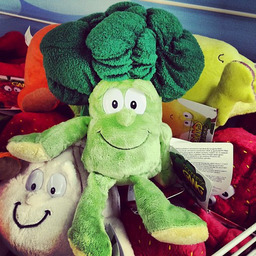} &
\bfigsupc[\sz]{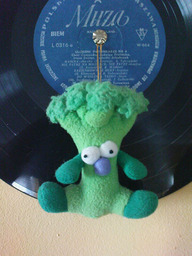} &
\bfigsupc[\sz]{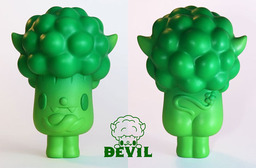} &
\bfigsupc[\sz]{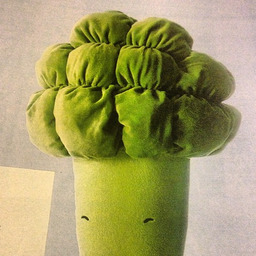} 
\\
\vspace{10pt}
&
\bfigsupq[\sz]{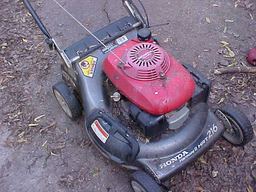} &
\bfigsupc[\sz]{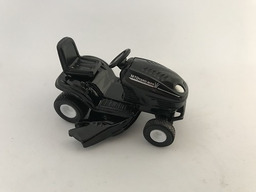} &
\bfigsupc[\sz]{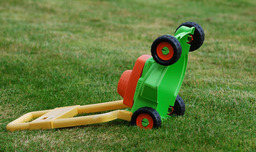} &
\bfigsupc[\sz]{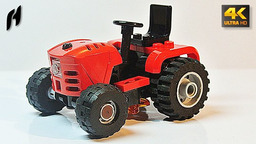} &
\bfigsupc[\sz]{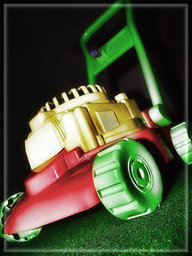} &
\bfigsupw[\sz]{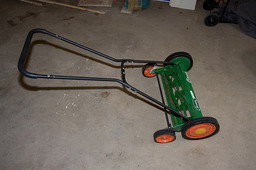} &
\bfigsupc[\sz]{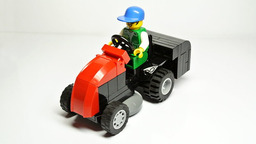} &
\bfigsupc[\sz]{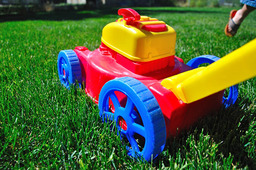} &
\bfigsupc[\sz]{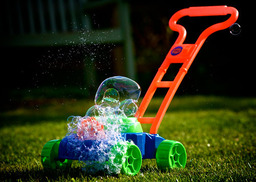} \\

&
image query &
1 & 
2 &
3 &
4 &
5 &
6 &
7 &
8 

\end{tabular}
\caption{\emph{Top retrieval results} of \ours. Domain conversion on \imagenetr: (a) \Th{Photo} $\rightarrow$ \Th{Cartoon}; (b) \Th{Photo} $\rightarrow$ \Th{Origami}; (c) \Th{Photo} $\rightarrow$ \Th{Sculpture}; (d) \Th{Photo} $\rightarrow$ \Th{Toy}. \textcolor{OrangeFrame}{Orange}: image query; \textcolor{OliveGreen}{green}: correctly retrieved; \textcolor{BrickRed}{red}: incorrectly retrieved.}
\label{fig:freedom_solo_imagenet_r}
\end{figure*}

%% file: fig_supp/retrievals_freedom_solo_sbir.tex
\begin{figure*}
\scriptsize
\centering
\setlength{\tabcolsep}{1.2pt}
\newcommand{\sz}{.08}
\begin{tabular}{ccccccccccc}
\vspace{7pt}
\bfigsupq[\sz]{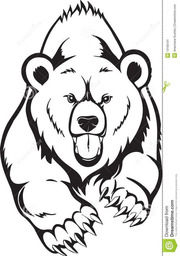} &
\bfigsupc[\sz]{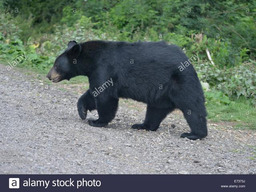} &
\bfigsupc[\sz]{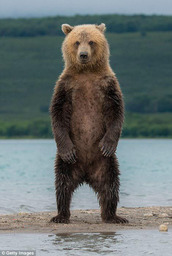} &
\bfigsupc[\sz]{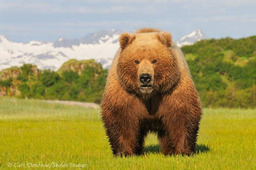} &
\bfigsupc[\sz]{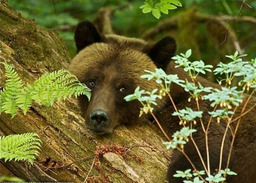} &
\bfigsupc[\sz]{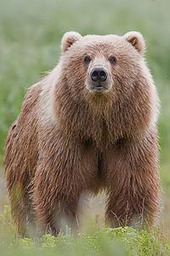} &
\bfigsupc[\sz]{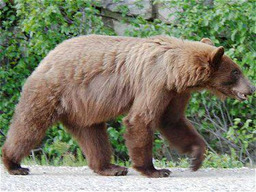} &
\bfigsupc[\sz]{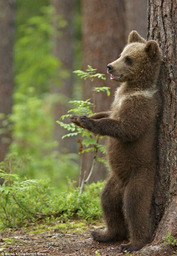} &
\bfigsupc[\sz]{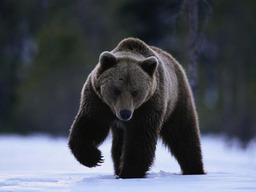}  
\\ \vspace{7pt}

\bfigsupq[\sz]{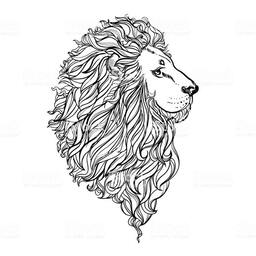} &
\bfigsupc[\sz]{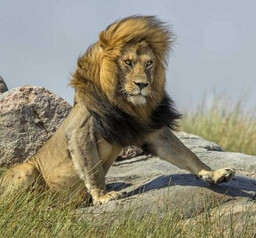} &
\bfigsupc[\sz]{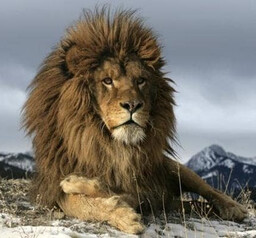} &
\bfigsupc[\sz]{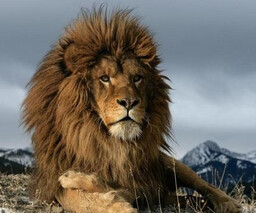} &
\bfigsupc[\sz]{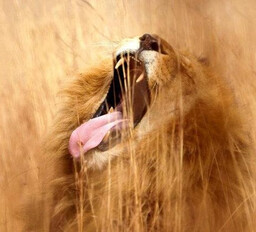} &
\bfigsupc[\sz]{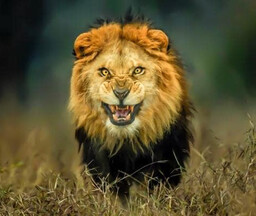} &
\bfigsupc[\sz]{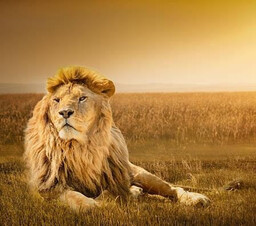} &
\bfigsupc[\sz]{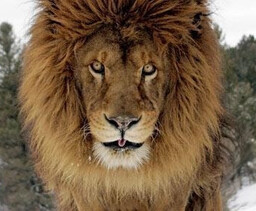} &
\bfigsupc[\sz]{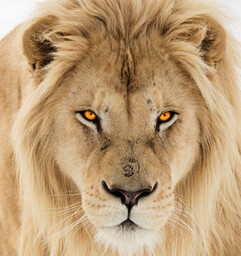}  
\\ \vspace{7pt}

\bfigsupq[\sz]{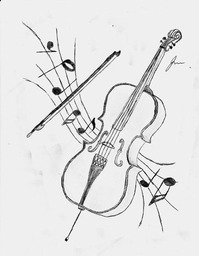} &
\bfigsupc[\sz]{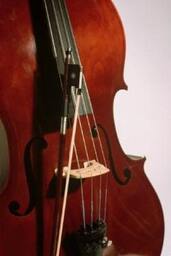} &
\bfigsupc[\sz]{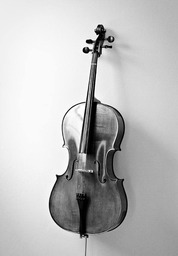} &
\bfigsupc[\sz]{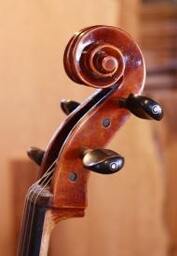} &
\bfigsupw[\sz]{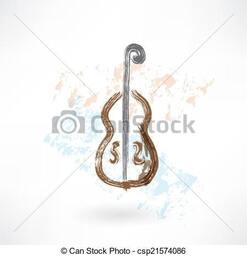} &
\bfigsupc[\sz]{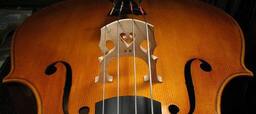} &
\bfigsupc[\sz]{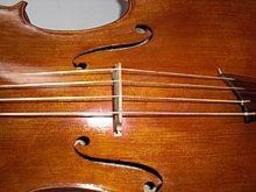} &
\bfigsupc[\sz]{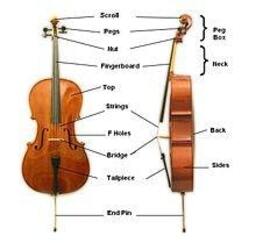} &
\bfigsupc[\sz]{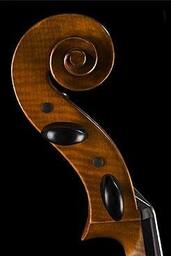}  
\\ \vspace{7pt}

\bfigsupq[\sz]{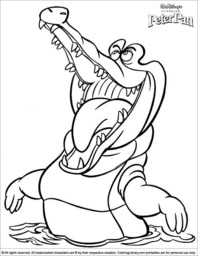} &
\bfigsupc[\sz]{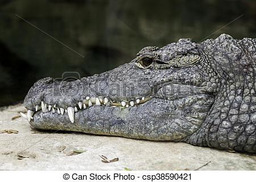} &
\bfigsupc[\sz]{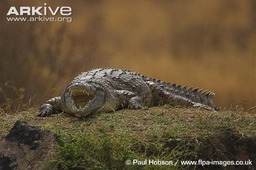} &
\bfigsupc[\sz]{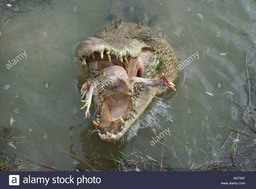} &
\bfigsupc[\sz]{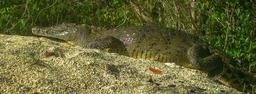} &
\bfigsupc[\sz]{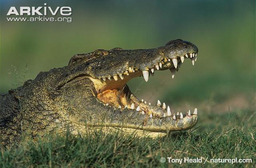} &
\bfigsupc[\sz]{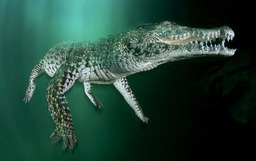} &
\bfigsupc[\sz]{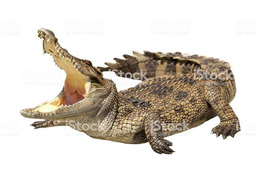} &
\bfigsupc[\sz]{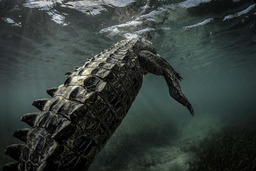} 
\\ \vspace{5pt}

\bfigsupq[\sz]{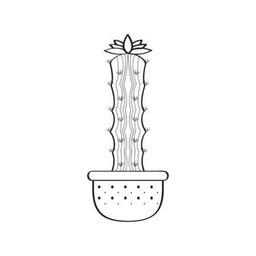} &
\bfigsupc[\sz]{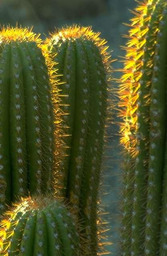} &
\bfigsupc[\sz]{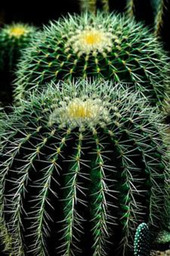} &
\bfigsupc[\sz]{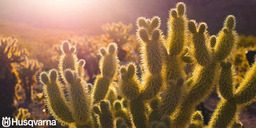} &
\bfigsupc[\sz]{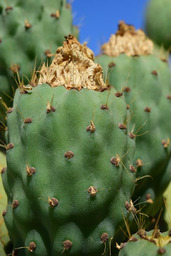} &
\bfigsupc[\sz]{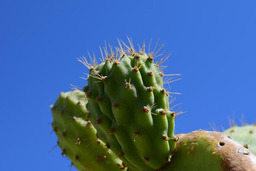} &
\bfigsupc[\sz]{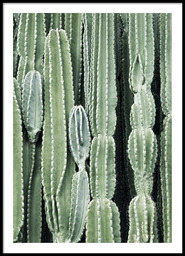} &
\bfigsupc[\sz]{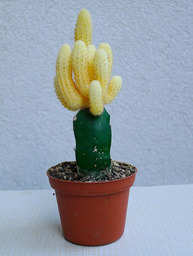} &
\bfigsupc[\sz]{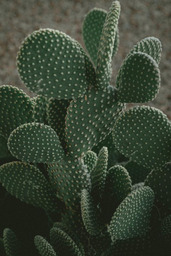} 
\\ \vspace{5pt}

\bfigsupq[\sz]{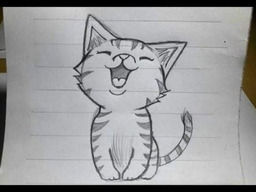} &
\bfigsupc[\sz]{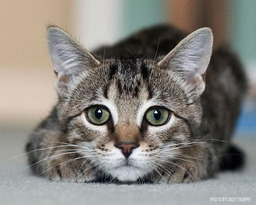} &
\bfigsupc[\sz]{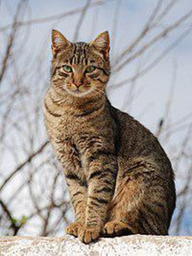} &
\bfigsupc[\sz]{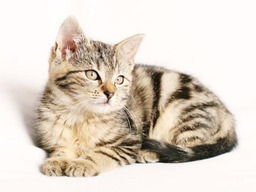} &
\bfigsupc[\sz]{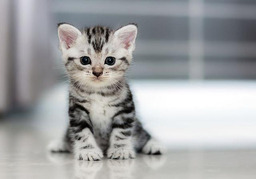} &
\bfigsupc[\sz]{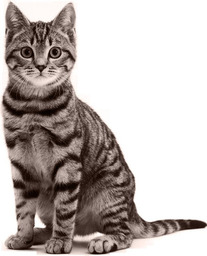} &
\bfigsupc[\sz]{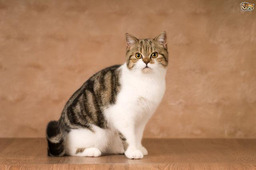} &
\bfigsupw[\sz]{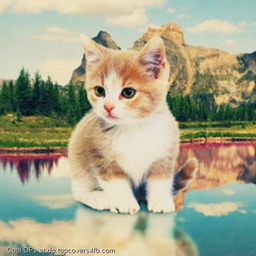} &
\bfigsupc[\sz]{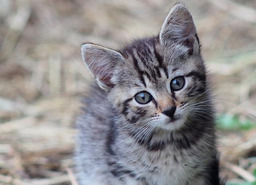} 
\\ \vspace{5pt}

\bfigsupq[\sz]{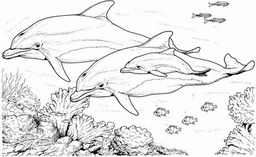} &
\bfigsupc[\sz]{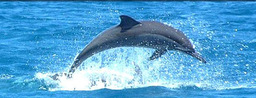} &
\bfigsupc[\sz]{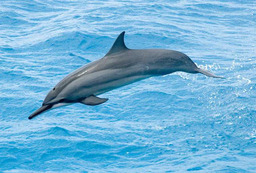} &
\bfigsupc[\sz]{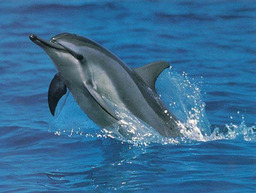} &
\bfigsupc[\sz]{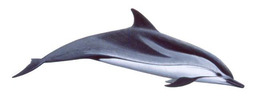} &
\bfigsupc[\sz]{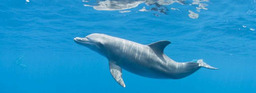} &
\bfigsupc[\sz]{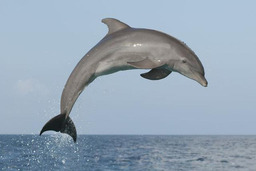} &
\bfigsupc[\sz]{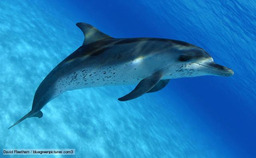} &
\bfigsupc[\sz]{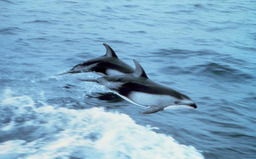} 
\\ \vspace{5pt}

\bfigsupq[\sz]{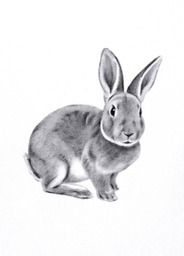} &
\bfigsupc[\sz]{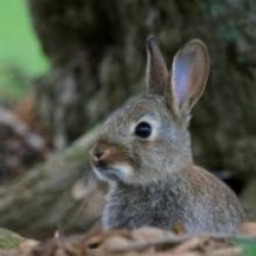} &
\bfigsupc[\sz]{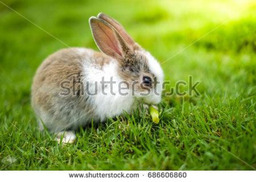} &
\bfigsupc[\sz]{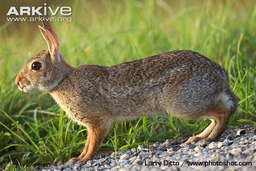} &
\bfigsupc[\sz]{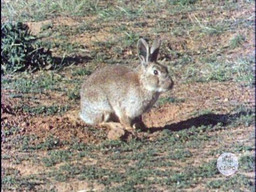} &
\bfigsupc[\sz]{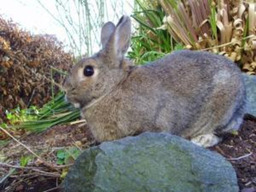} &
\bfigsupc[\sz]{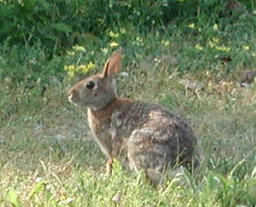} &
\bfigsupc[\sz]{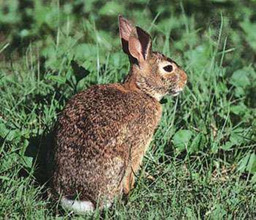} &
\bfigsupc[\sz]{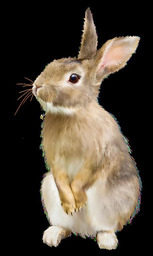} 
\\ \vspace{5pt}

\bfigsupq[\sz]{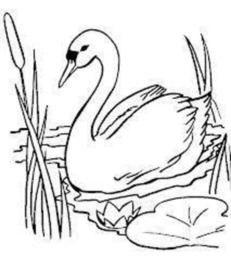} &
\bfigsupc[\sz]{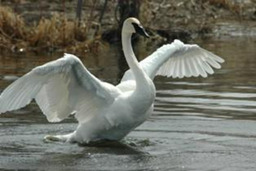} &
\bfigsupc[\sz]{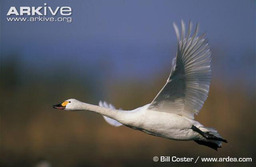} &
\bfigsupc[\sz]{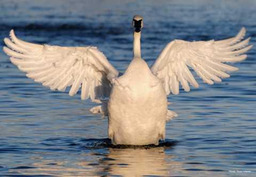} &
\bfigsupc[\sz]{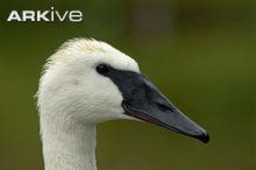} &
\bfigsupc[\sz]{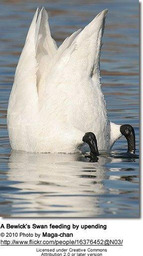} &
\bfigsupc[\sz]{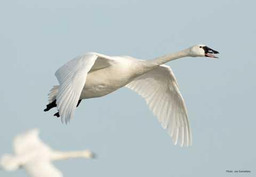} &
\bfigsupc[\sz]{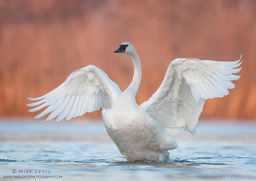} &
\bfigsupc[\sz]{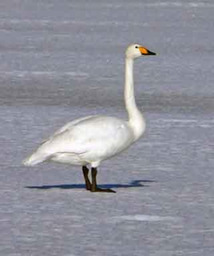} 
\\ \vspace{5pt}

\bfigsupq[\sz]{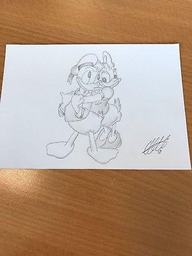} &
\bfigsupc[\sz]{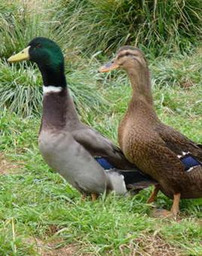} &
\bfigsupw[\sz]{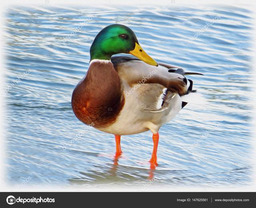} &
\bfigsupw[\sz]{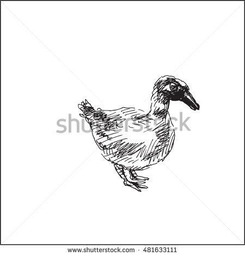} &
\bfigsupw[\sz]{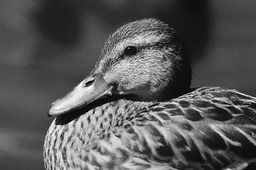} &
\bfigsupc[\sz]{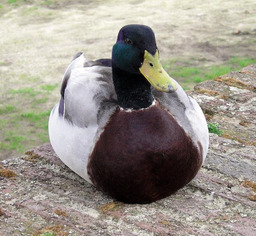} &
\bfigsupw[\sz]{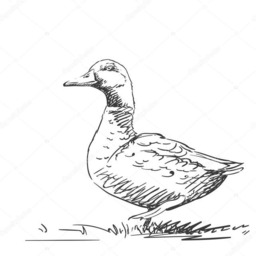} &
\bfigsupc[\sz]{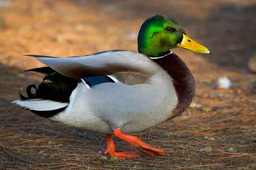} &
\bfigsupw[\sz]{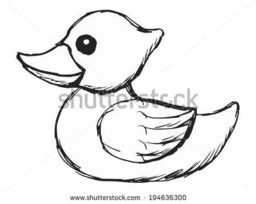}
\\

image query &
1 &
2 &
3 &
4 &
5 &
6 &
7 &
8 

\end{tabular}
\caption{\emph{Top retrieval results} of \ours. Sketch-based image retrieval (\Th{Sketch} $\rightarrow$ \Th{Photo}) on \minidn. \textcolor{OrangeFrame}{Orange}: image query; \textcolor{OliveGreen}{green}: correctly retrieved; \textcolor{BrickRed}{red}: incorrectly retrieved.}
\label{fig:freedom_solo_sbir}
\end{figure*}

%% file: fig_supp/retrievals_competitors_ltll.tex
\begin{figure*}
\scriptsize
\centering
\setlength{\tabcolsep}{1.2pt}
\newcommand{\sz}{.08}
\begin{tabular}{ccccccccccc}

\bfigsupq[\sz]{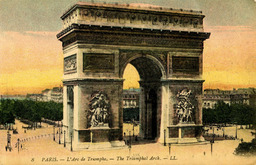} &
\multirow{1}{*}[10ex]{\rotatebox[origin=c]{90}{\picword}} &
\bfigsupw[\sz]{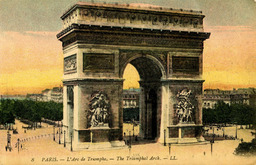} &
\bfigsupw[\sz]{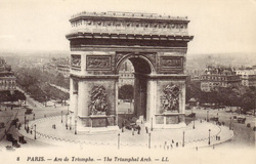} &
\bfigsupw[\sz]{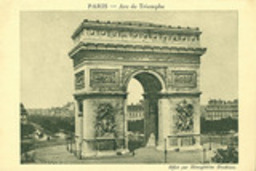} &
\bfigsupw[\sz]{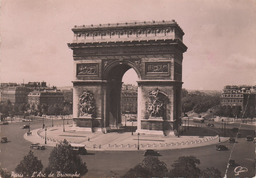} &
\bfigsupw[\sz]{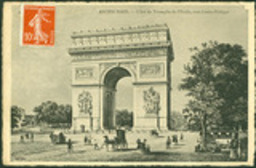} &
\bfigsupw[\sz]{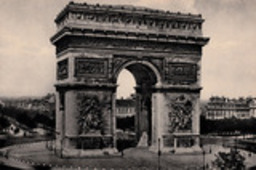} &
\bfigsupw[\sz]{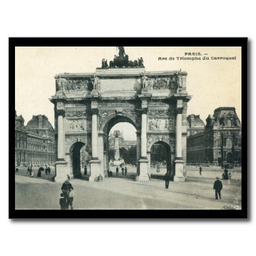} &
\bfigsupc[\sz]{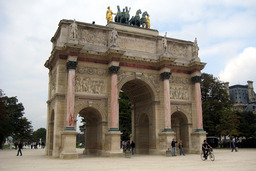} \vspace{5pt} \\

& 
\multirow{1}{*}[10ex]{\rotatebox[origin=c]{90}{\compodiff}} &
\bfigsupw[\sz]{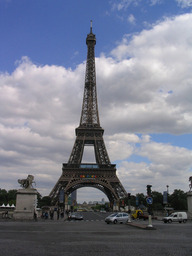} &
\bfigsupw[\sz]{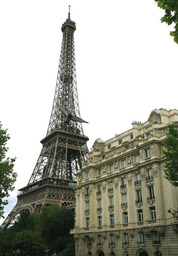} &
\bfigsupc[\sz]{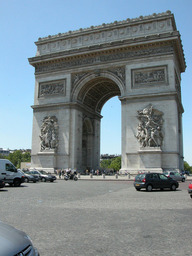} &
\bfigsupw[\sz]{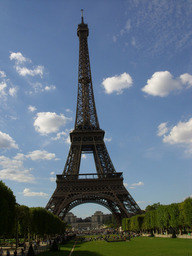} &
\bfigsupw[\sz]{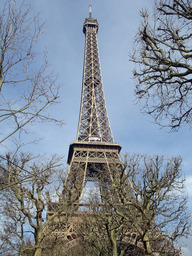} &
\bfigsupc[\sz]{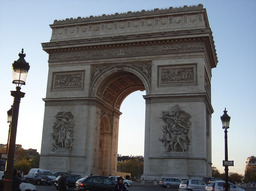} &
\bfigsupc[\sz]{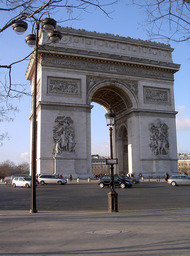} &
\bfigsupw[\sz]{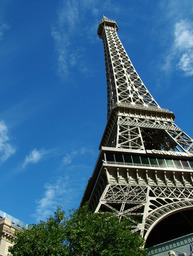} \vspace{5pt} \\

& 
\multirow{1}{*}[10ex]{\rotatebox[origin=c]{90}{\searle}} &
\bfigsupw[\sz]{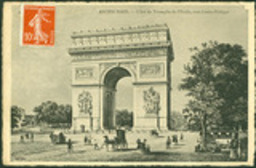} &
\bfigsupw[\sz]{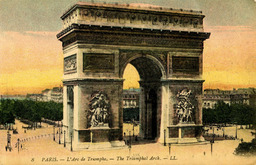} &
\bfigsupw[\sz]{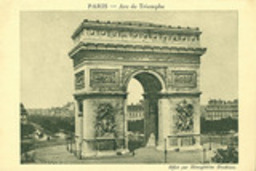} &
\bfigsupw[\sz]{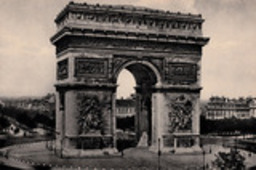} &
\bfigsupw[\sz]{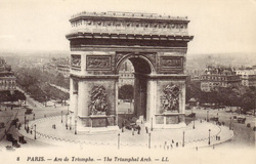} &
\bfigsupc[\sz]{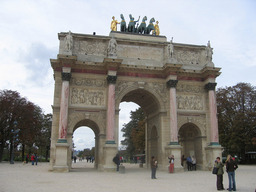} &
\bfigsupc[\sz]{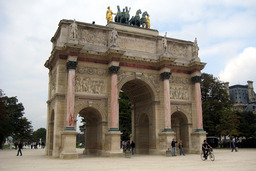} &
\bfigsupw[\sz]{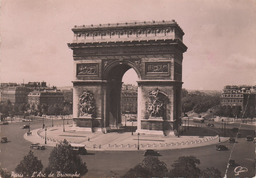} \vspace{5pt} \\

& 
\multirow{1}{*}[10ex]{\rotatebox[origin=c]{90}{\magic}} &
\bfigsupw[\sz]{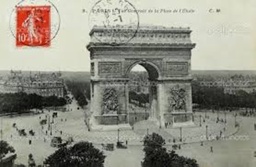} &
\bfigsupw[\sz]{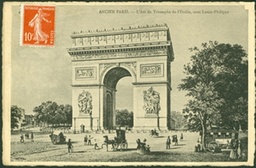} &
\bfigsupw[\sz]{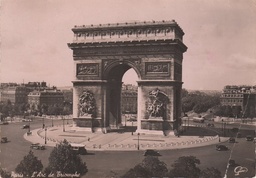} &
\bfigsupw[\sz]{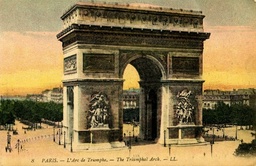} &
\bfigsupw[\sz]{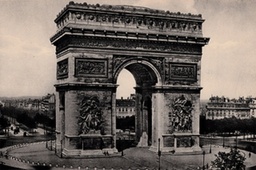} &
\bfigsupw[\sz]{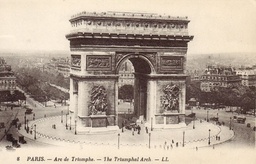} &
\bfigsupw[\sz]{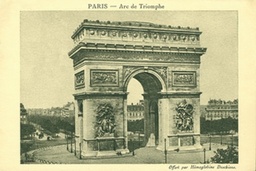} &
\bfigsupc[\sz]{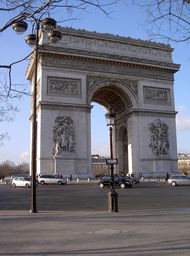} \vspace{5pt} \\

& 
\multirow{1}{*}[10ex]{\rotatebox[origin=c]{90}{\ours}} &
\bfigsupc[\sz]{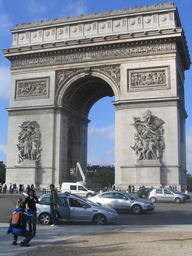} &
\bfigsupc[\sz]{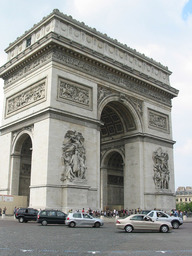} &
\bfigsupc[\sz]{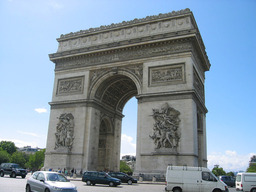} &
\bfigsupc[\sz]{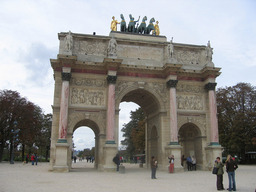} &
\bfigsupc[\sz]{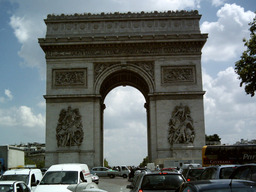} &
\bfigsupc[\sz]{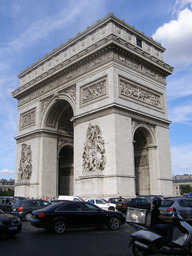} &
\bfigsupc[\sz]{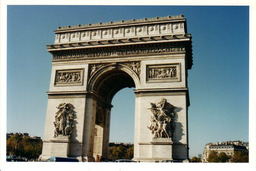} &
\bfigsupc[\sz]{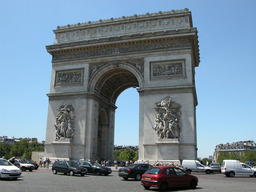} \vspace{15pt} \\

\bfigsupq[\sz]{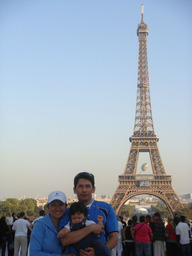} &
\multirow{1}{*}[10ex]{\rotatebox[origin=c]{90}{\picword}} &
\bfigsupw[\sz]{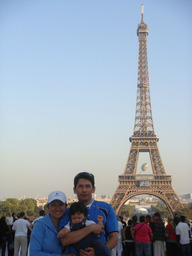} &
\bfigsupw[\sz]{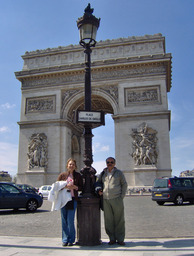} &
\bfigsupw[\sz]{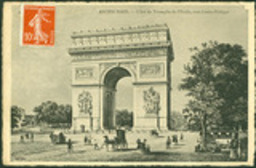} &
\bfigsupw[\sz]{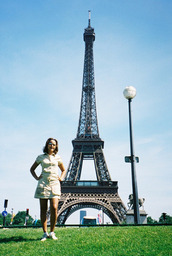} &
\bfigsupw[\sz]{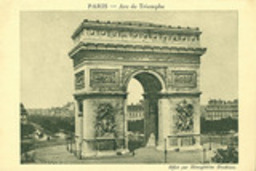} &
\bfigsupw[\sz]{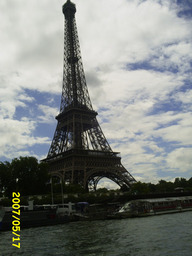} &
\bfigsupw[\sz]{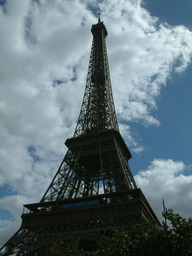} &
\bfigsupw[\sz]{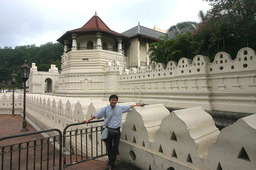} \vspace{5pt} \\

& 
\multirow{1}{*}[10ex]{\rotatebox[origin=c]{90}{\compodiff}} &
\bfigsupc[\sz]{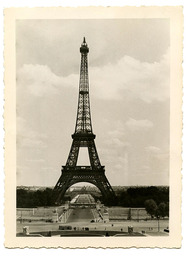} &
\bfigsupw[\sz]{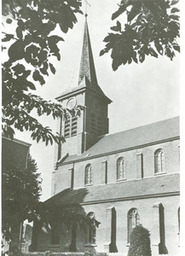} &
\bfigsupw[\sz]{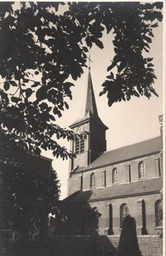} &
\bfigsupw[\sz]{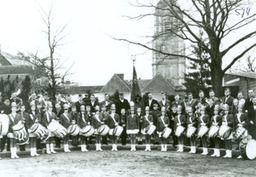} &
\bfigsupw[\sz]{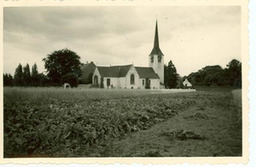} &
\bfigsupw[\sz]{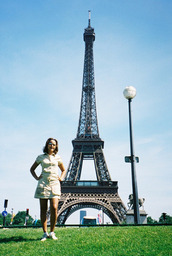} &
\bfigsupw[\sz]{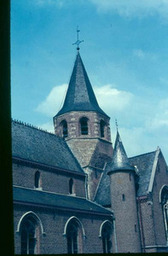} &
\bfigsupw[\sz]{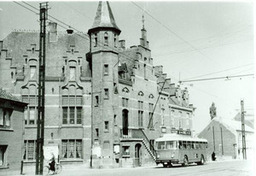} \vspace{5pt} \\

& 
\multirow{1}{*}[10ex]{\rotatebox[origin=c]{90}{\searle}} &
\bfigsupw[\sz]{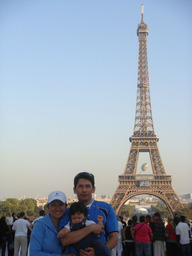} &
\bfigsupw[\sz]{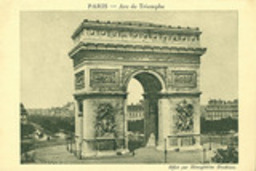} &
\bfigsupw[\sz]{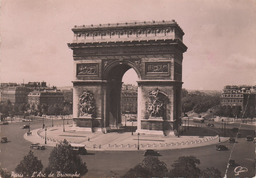} &
\bfigsupw[\sz]{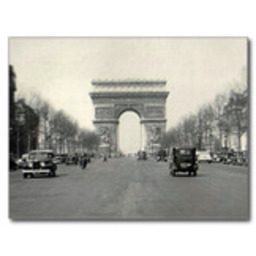} &
\bfigsupw[\sz]{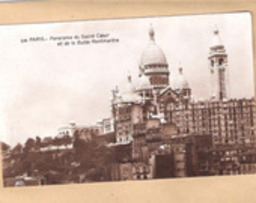} &
\bfigsupw[\sz]{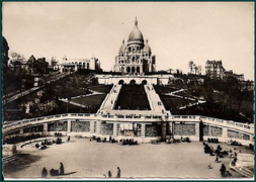} &
\bfigsupw[\sz]{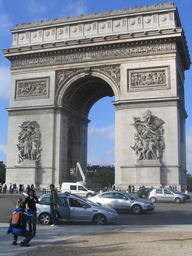} &
\bfigsupw[\sz]{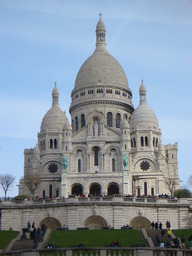} \vspace{5pt} \\

& 
\multirow{1}{*}[10ex]{\rotatebox[origin=c]{90}{\magic}} &
\bfigsupw[\sz]{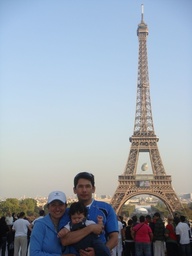} &
\bfigsupw[\sz]{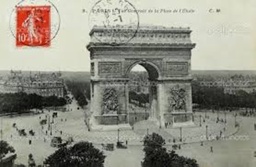} &
\bfigsupw[\sz]{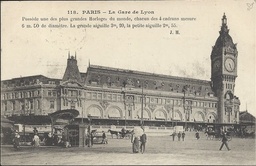} &
\bfigsupw[\sz]{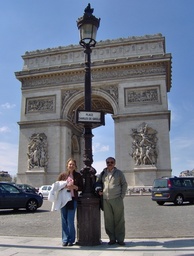} &
\bfigsupc[\sz]{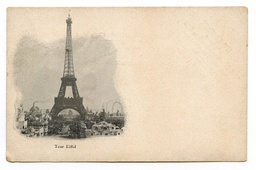} &
\bfigsupw[\sz]{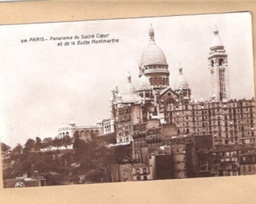} &
\bfigsupc[\sz]{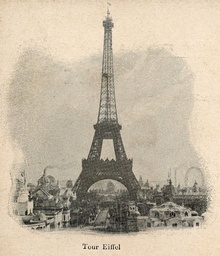} &
\bfigsupw[\sz]{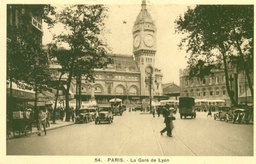} \vspace{5pt} \\

& 
\multirow{1}{*}[10ex]{\rotatebox[origin=c]{90}{\ours}} &
\bfigsupc[\sz]{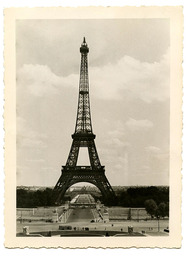} &
\bfigsupw[\sz]{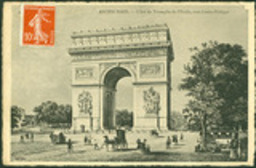} &
\bfigsupc[\sz]{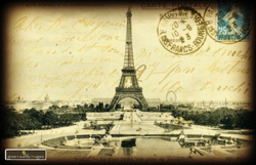} &
\bfigsupc[\sz]{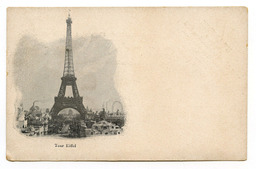} &
\bfigsupw[\sz]{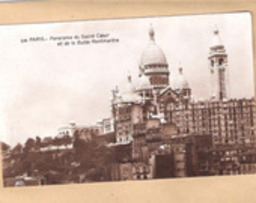} &
\bfigsupc[\sz]{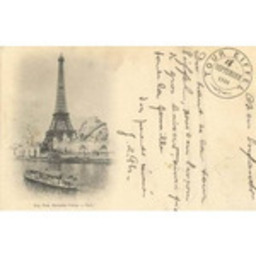} &
\bfigsupc[\sz]{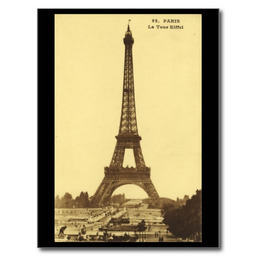} &
\bfigsupc[\sz]{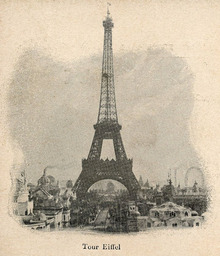} \vspace{5pt} \\

&
&
1 &
2 &
3 &
4 &
5 &
6 &
7 &
8 \\

\end{tabular}
\caption{\emph{Top retrieval results}. Competitors vs. \ours. Domain conversion (\Th{Archive} $\rightarrow$ \Th{Today}, \Th{Today} $\rightarrow$ \Th{Archive}) on \ltll. \textcolor{OrangeFrame}{Orange}: image query; \textcolor{OliveGreen}{green}: correctly retrieved; \textcolor{BrickRed}{red}: incorrectly retrieved.}
\label{fig:freedom_competitors_ltll}
\end{figure*}

%% file: fig_supp/retrievals_competitors_nico.tex
\begin{figure*}
\scriptsize
\centering
\setlength{\tabcolsep}{1.2pt}
\newcommand{\sz}{.08}
\begin{tabular}{ccccccccccc}

\bfigsupq[\sz]{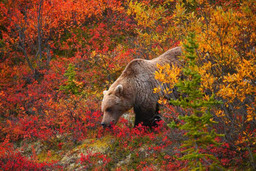} &
\multirow{1}{*}[10ex]{\rotatebox[origin=c]{90}{\picword}} &
\bfigsupw[\sz]{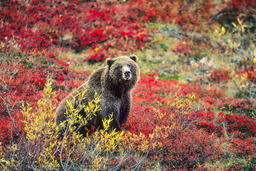} &
\bfigsupw[\sz]{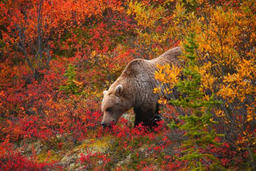} &
\bfigsupw[\sz]{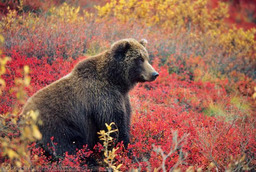} &
\bfigsupw[\sz]{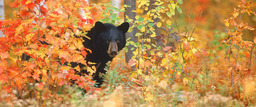} &
\bfigsupw[\sz]{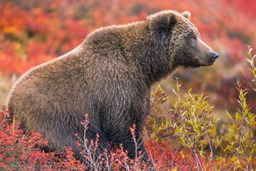} &
\bfigsupw[\sz]{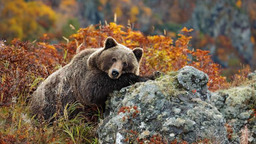} &
\bfigsupw[\sz]{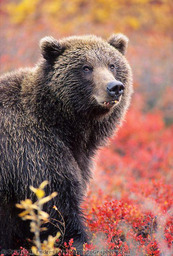} &
\bfigsupc[\sz]{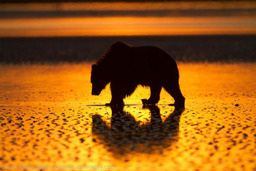} \vspace{5pt} \\

& 
\multirow{1}{*}[10ex]{\rotatebox[origin=c]{90}{\compodiff}} &
\bfigsupw[\sz]{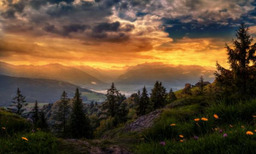} &
\bfigsupw[\sz]{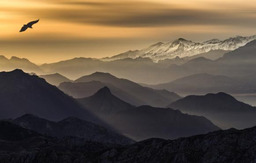} &
\bfigsupw[\sz]{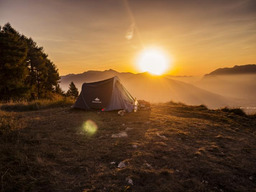} &
\bfigsupw[\sz]{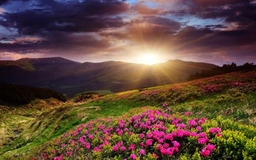} &
\bfigsupw[\sz]{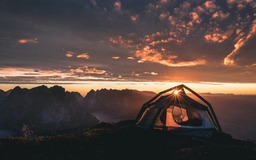} &
\bfigsupw[\sz]{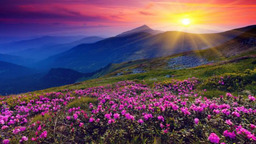} &
\bfigsupw[\sz]{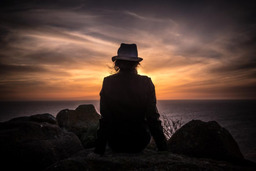} &
\bfigsupw[\sz]{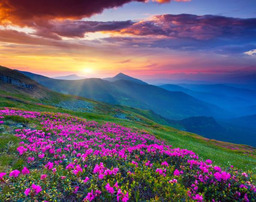} \vspace{5pt} \\

& 
\multirow{1}{*}[10ex]{\rotatebox[origin=c]{90}{\searle}} &
\bfigsupw[\sz]{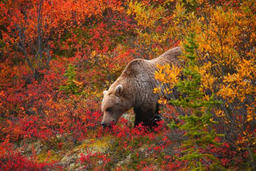} &
\bfigsupw[\sz]{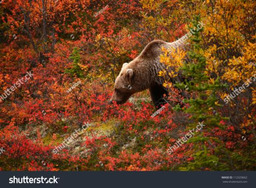} &
\bfigsupw[\sz]{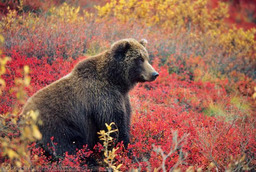} &
\bfigsupw[\sz]{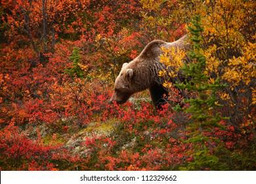} &
\bfigsupw[\sz]{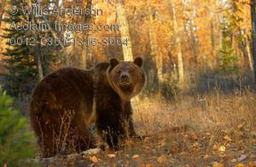} &
\bfigsupw[\sz]{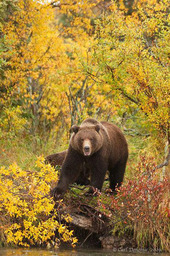} &
\bfigsupw[\sz]{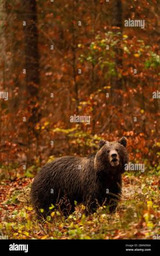} &
\bfigsupw[\sz]{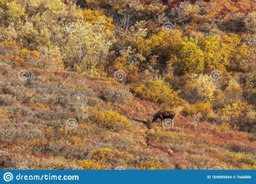} \vspace{5pt} \\

& 
\multirow{1}{*}[10ex]{\rotatebox[origin=c]{90}{\magic}} &
\bfigsupc[\sz]{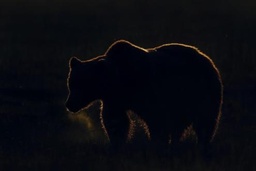} &
\bfigsupc[\sz]{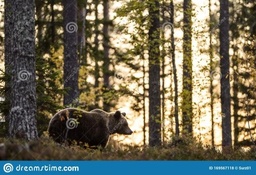} &
\bfigsupc[\sz]{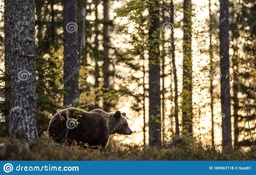} &
\bfigsupw[\sz]{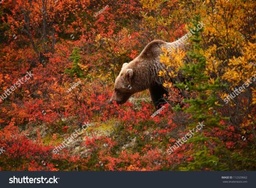} &
\bfigsupc[\sz]{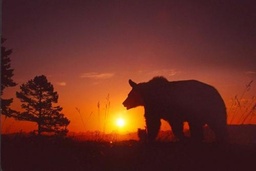} &
\bfigsupw[\sz]{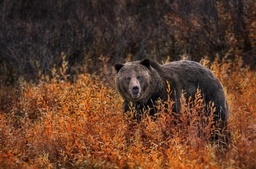} &
\bfigsupc[\sz]{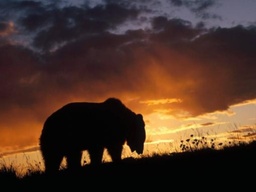} &
\bfigsupw[\sz]{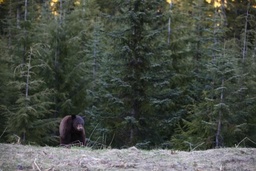} \vspace{5pt} \\

& 
\multirow{1}{*}[10ex]{\rotatebox[origin=c]{90}{\ours}} &
\bfigsupw[\sz]{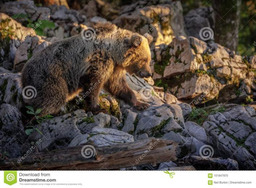} &
\bfigsupc[\sz]{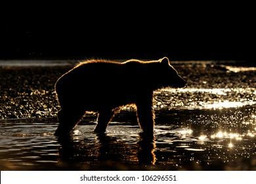} &
\bfigsupc[\sz]{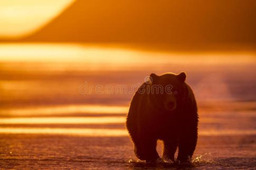} &
\bfigsupc[\sz]{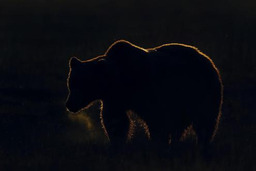} &
\bfigsupc[\sz]{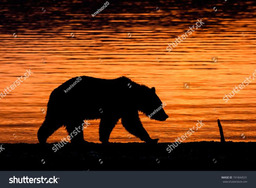} &
\bfigsupc[\sz]{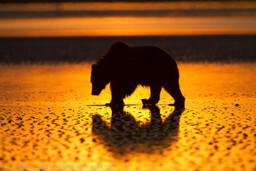} &
\bfigsupw[\sz]{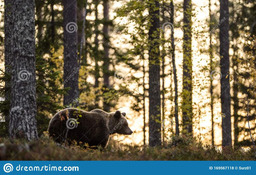} &
\bfigsupw[\sz]{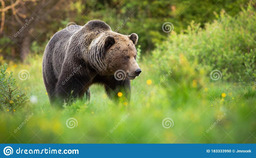} \vspace{15pt} \\

\bfigsupq[\sz]{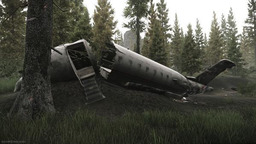} &
\multirow{1}{*}[10ex]{\rotatebox[origin=c]{90}{\picword}} &
\bfigsupw[\sz]{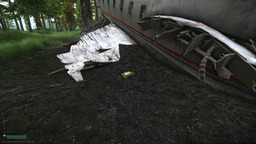} &
\bfigsupw[\sz]{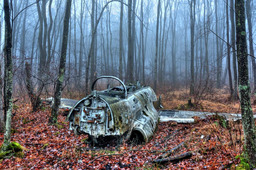} &
\bfigsupw[\sz]{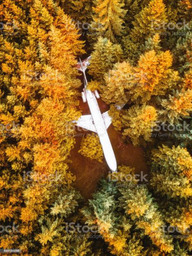} &
\bfigsupw[\sz]{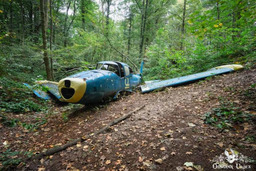} &
\bfigsupw[\sz]{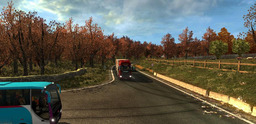} &
\bfigsupw[\sz]{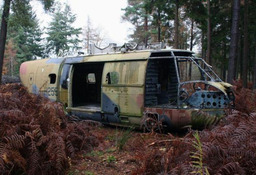} &
\bfigsupw[\sz]{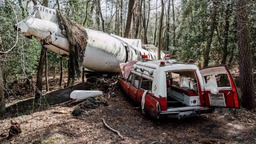} &
\bfigsupc[\sz]{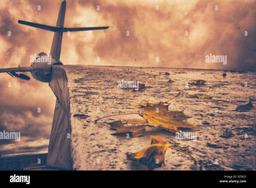} \vspace{5pt} \\

& 
\multirow{1}{*}[10ex]{\rotatebox[origin=c]{90}{\compodiff}} &
\bfigsupw[\sz]{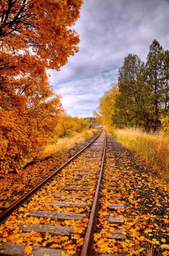} &
\bfigsupw[\sz]{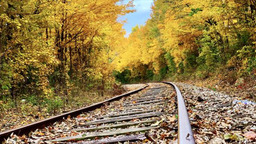} &
\bfigsupw[\sz]{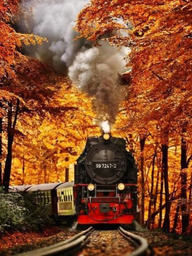} &
\bfigsupw[\sz]{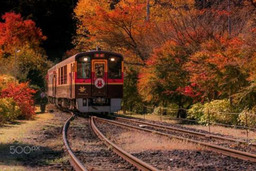} &
\bfigsupw[\sz]{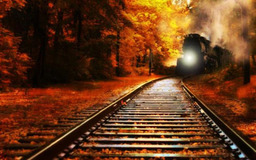} &
\bfigsupw[\sz]{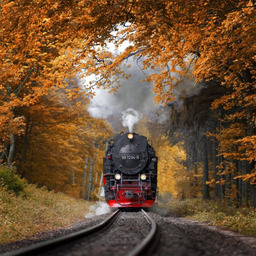} &
\bfigsupw[\sz]{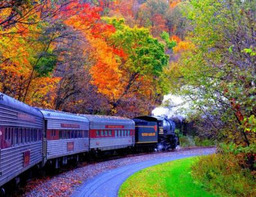} &
\bfigsupw[\sz]{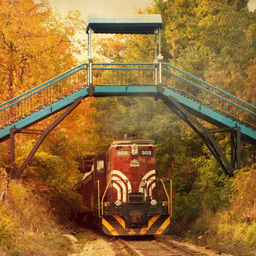} \vspace{5pt} \\

& 
\multirow{1}{*}[10ex]{\rotatebox[origin=c]{90}{\searle}} &
\bfigsupw[\sz]{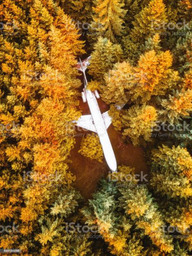} &
\bfigsupw[\sz]{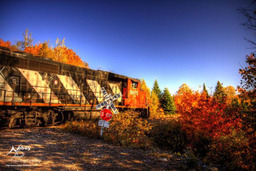} &
\bfigsupw[\sz]{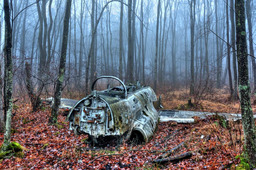} &
\bfigsupw[\sz]{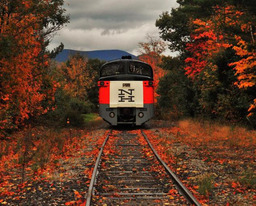} &
\bfigsupw[\sz]{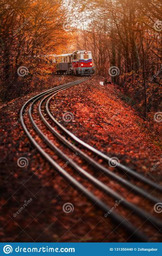} &
\bfigsupw[\sz]{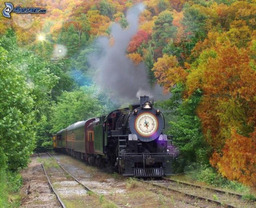} &
\bfigsupw[\sz]{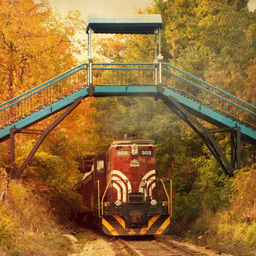} &
\bfigsupw[\sz]{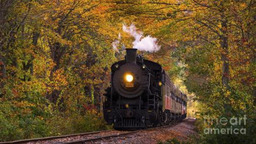} \vspace{5pt} \\

& 
\multirow{1}{*}[10ex]{\rotatebox[origin=c]{90}{\magic}} &
\bfigsupw[\sz]{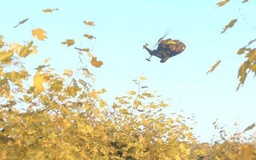} &
\bfigsupw[\sz]{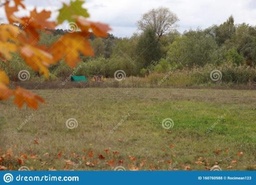} &
\bfigsupw[\sz]{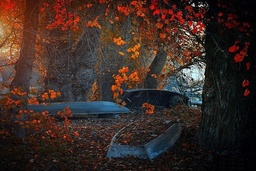} &
\bfigsupc[\sz]{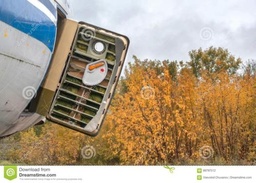} &
\bfigsupw[\sz]{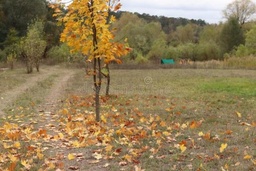} &
\bfigsupw[\sz]{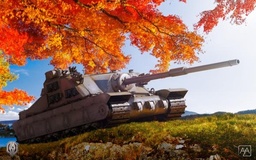} &
\bfigsupw[\sz]{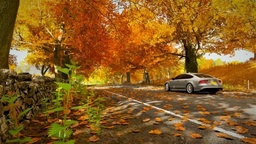} &
\bfigsupw[\sz]{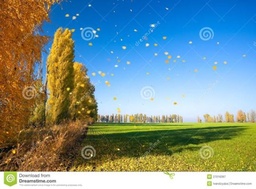} \vspace{5pt} \\

& 
\multirow{1}{*}[10ex]{\rotatebox[origin=c]{90}{\ours}} &
\bfigsupc[\sz]{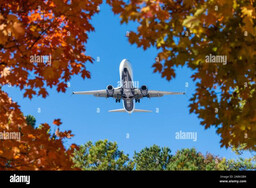} &
\bfigsupc[\sz]{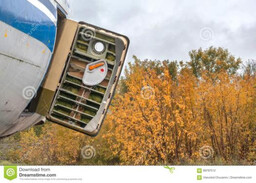} &
\bfigsupc[\sz]{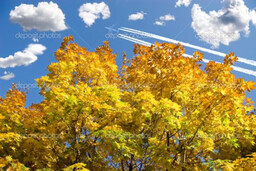} &
\bfigsupc[\sz]{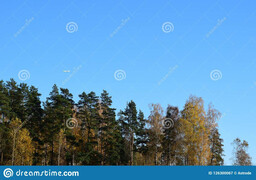} &
\bfigsupw[\sz]{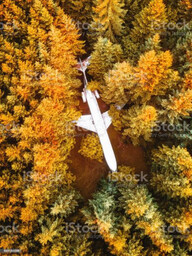} &
\bfigsupc[\sz]{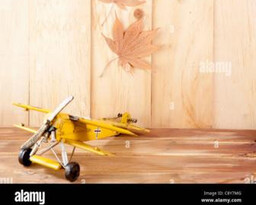} &
\bfigsupc[\sz]{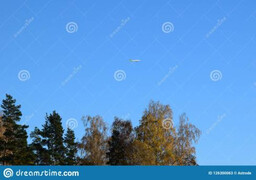} &
\bfigsupw[\sz]{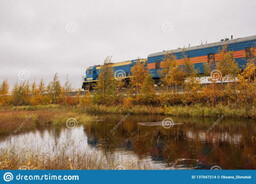} \vspace{5pt} \\

&
&
1 &
2 &
3 &
4 &
5 &
6 &
7 &
8 \\

\end{tabular}
\caption{\emph{Top retrieval results}. Competitors vs. \ours. Domain conversion (\Th{Autumn} $\rightarrow$ \Th{Dimlight}, \Th{Grass} $\rightarrow$ \Th{Autumn}) on \nico. \textcolor{OrangeFrame}{Orange}: image query; \textcolor{OliveGreen}{green}: correctly retrieved; \textcolor{BrickRed}{red}: incorrectly retrieved.}
\label{fig:freedom_competitors_nico}
\end{figure*}